\documentclass[letterpaper, 10 pt, conference]{ieeeconf}  

\IEEEoverridecommandlockouts                              

\usepackage{amssymb}
\usepackage{amsmath}
\usepackage[ruled,lined,linesnumbered]{algorithm2e}
\usepackage[most]{tcolorbox}
\usepackage{xcolor}
\usepackage{graphicx}
\usepackage{subfigure}
\usepackage{url}
\usepackage[inkscapelatex=false]{svg}
\newcommand{\PL}[1]{{\color{black}{#1}}}

\title{\LARGE \bf LLM-Flock: LLM-Driven Decentralized Multi-Robot Flocking via Influence-Based Consensus with Collision Avoidance}

\newif\ifanonymous
\anonymousfalse

\ifanonymous
    \author{Anonymous Authors}
\else

\author{Peihan Li and Lifeng Zhou
\thanks{*This work was supported by the NSF FRR program (No. 2542649).}
\thanks{Peihan Li and Lifeng Zhou are with the Department of Electrical and Computer Engineering, Drexel University, Philadelphia, PA 19104, USA
        {\tt\small \{pl525, lz457\}@drexel.edu}}}%
\fi

\begin{document}
\maketitle

\begin{abstract}
\PL{Large Language Models (LLMs) have shown strong capabilities in reasoning and planning, prompting researchers to explore their use as decentralized decision-makers for multi-robot formation control. However, directly applying LLMs to such tasks often leads to unstable behavior, in which robots may collapse to the centroid of their positions or diverge entirely, due to hallucinated reasoning and limited coordination awareness. We propose LLM-Flock, a decentralized multi-robot flocking framework that addresses these challenges through two key mechanisms: (1) an influence-based consensus protocol, where robots iteratively refine individually generated formation plans by adopting those of more connected neighbors, driving the team toward a unified plan without centralized control, and (2) a two-layer collision avoidance strategy that pairs LLM-based urgency reasoning with a programmatic collision checker and re-prompting mechanism. We evaluate LLM-Flock across five state-of-the-art LLMs, including both closed-source (GPT-5, Claude 4.5) and open-source (Llama3.1-405b, Qwen-Max, DeepSeek-R1) models, demonstrating notable improvements in formation stability and convergence over prior LLM-based methods. Real-world experiments with Crazyflie drones further validate the framework's applicability to physical multi-robot systems.}

\end{abstract}

\section{Introduction}
\label{sec:introduction}

Decentralized formation control in multi-robot systems has been a popular topic over the decades~\cite{liu2018survey, peng2018decentralized}. Enabling a team of robots to coordinate and maintain desired formations without centralized control is critical for applications such as environmental monitoring, search-and-rescue missions, and distributed exploration~\cite{konda2020decentralized, bouraqadi2009flocking, notomista2022multi, ma2018multi, burgard2005coordinated}. Classical decentralized control strategies, including rule-based flocking algorithms and graph-based consensus methods, achieve global coordination through local interactions~\cite{olfati2006flocking, shao2023dynamic}. However, these approaches often rely on simple heuristics or static interaction models, limiting adaptability in dynamic environments and complex mission objectives~\cite{de2010heuristic, antonelli2014decentralized}. More importantly, deploying such methods in practice typically requires specialized expertise in robotics and control theory. End-users, such as first responders, biologists, or inspectors, must either undergo extensive training or rely on pre-programmed behaviors, creating a significant barrier to the widespread use of multi-robot systems in the field~\cite{chen2024scalable}.

Recent advancements in large language models (LLMs) have introduced new opportunities for flexible and generalized reasoning in robotics~\cite{tao2024survey, li2025large, mandi2024roco, wang2024dart, wu_hierarchical_2024}. Trained on massive text corpora, LLMs demonstrate structured planning, goal inference, and task adaptation across diverse domains without explicit retraining~\cite{achiam2023gpt, grattafiori2024llama}. Unlike traditional controllers, LLMs can interpret high-level natural-language instructions and translate them into concrete actions, thereby lowering the barrier for non-expert users to operate complex multi-robot systems~\cite{li2024challenges, li2025large}. In the context of decentralized formation control, this suggests a new paradigm in which end-users describe mission objectives in plain language and robots collaboratively generate and refine formation strategies through LLM-driven reasoning~\cite{guo_large_2024}.

However, directly applying LLMs to decentralized multi-robot coordination presents significant challenges~\cite{hunt_survey_2024, kim_survey_2024, guo_large_2024}. Independent LLM-driven robots often exhibit hallucinated reasoning, logical inconsistencies, and a lack of collective awareness~\cite{huang2023survey, mcdonald2024reducing}. Without structured mechanisms to align individual plans, such systems can collapse into degenerate behaviors, such as converging to the centroid or diverging chaotically, undermining effective decentralized control~\cite{li2024challenges, chen2024solving}.

\begin{figure}
    \centering
    \includegraphics[width=1\linewidth]{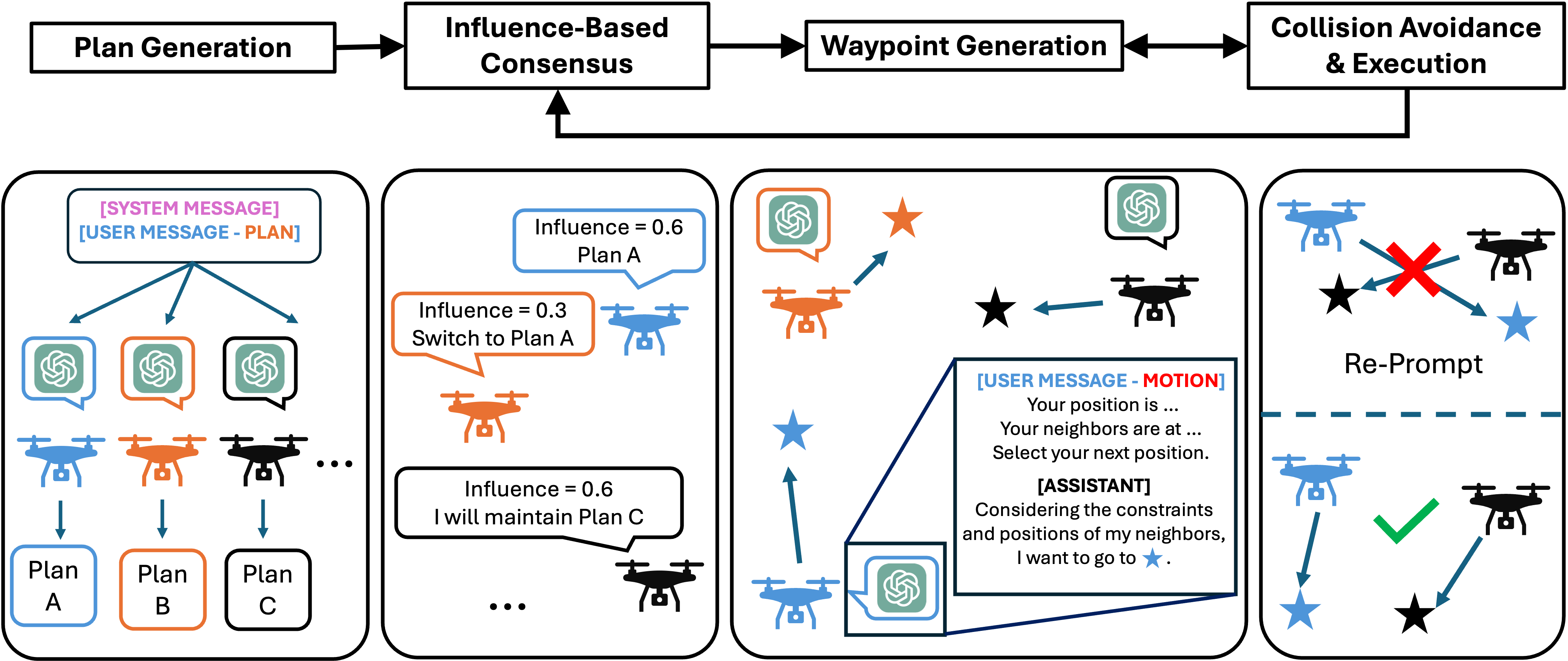}

    \caption{LLM-Flock framework for decentralized multi-robot flocking. Each robot first uses an LLM to generate a candidate plan from a shared prompt. Then robots exchange plans locally and iteratively refine them through an influence-based consensus protocol. Each robot’s LLM determines the waypoint location at each step, which is verified by the two-layer collision-avoidance mechanism, enabling coordination without centralized control.}
    \label{fig:llm-flock}
    \vspace{-0.4cm}
\end{figure}

To address these challenges, we propose \textit{LLM-Flock} (Fig.~\ref{fig:llm-flock}), a decentralized multi-robot flocking framework that integrates LLM-driven local planning with an influence-based consensus protocol\PL{ and collision-avoidance checking and re-prompting}. Each robot independently queries an LLM to generate a candidate formation plan based on local objectives and environmental constraints. Robots then iteratively negotiate plan adoption based on the number of neighbors, dynamically driving the team toward coherent and stable formations without centralized coordination or retraining. 

The main contributions of this work are: 
\begin{itemize} 
\item We motivate the use of LLMs for decentralized flocking not only as a technical alternative but as a way to lower the barrier for non-experts to operate multi-robot systems, improving the accessibility and usability of multi-robot coordination in real-world settings with natural language commands.
\PL{\item We propose a generalizable decentralized framework for multi-robot flocking that combines LLM-based local planning with influence-based consensus and a two-layer collision-avoidance mechanism.}
\item We design a structured interaction protocol that unifies decentralized reasoning through lightweight local negotiation. 
\item We validate the framework across a diverse set of closed-source and open-source LLMs, demonstrating its effectiveness and generality. 
\item We demonstrate practical feasibility through comprehensive simulations and real-world experiments with physical multi-drone systems. 
\end{itemize}

\noindent The remainder of this paper is organized as follows: Section~\ref{sec:related_works} reviews prior work on multi-robot formation control and multi-robot coordination with LLMs. Section~\ref{sec:methods} details the proposed LLM-Flock framework. Section~\ref{sec:results} presents simulation and real-world experiments validating the framework. Section~\ref{sec:conclusion} concludes the paper.

\section{Related Works}
\label{sec:related_works}
\textbf{Multi-Robot Formation Control.}  
Flocking and formation control are well-established topics in multi-robot systems. Classical methods, such as Reynolds' boid model and Olfati-Saber’s decentralized algorithms, achieve coordination through local rules like separation, alignment, and cohesion~\cite{reynolds1987flocks, olfati2006flocking, blumenkamp2022framework}. Later studies introduced graph-based consensus protocols to improve stability and scalability~\cite{goarin2024graph}. Reinforcement learning (RL) approaches, including MADDPG and QMIX, enable decentralized strategies for multi-robot coordination~\cite{lowe2017multi, rashid2020monotonic}, but typically require extensive training and are sensitive to environment dynamics. Imitation learning such as graph neural networks~\cite{chen2023spatial, chen2024learning, tolstaya2020learning, khan2020graph} offers an alternative, but it faces challenges such as dataset dependence and covariate shift.

\textbf{LLMs for Multi-Robot Coordination.}  
LLMs offer new opportunities for flexible reasoning in multi-robot coordination. Recent frameworks such as RoCo~\cite{mandi2024roco} and SMART-LLM~\cite{kannan2024smart} demonstrate the potential of LLMs for task planning and coordination in multi-robot systems. Other studies, such as ZeroCAP~\cite{venkatesh_zerocap_2024}, have explored LLM-based formation control, showing promising results. However, these approaches rely on centralized architectures with global environment knowledge, limiting their scalability and real-world applicability. 

\textbf{Challenges of LLM-Based Decentralized Reasoning.}  
Despite their promise, LLM-driven robots exhibit hallucinated outputs, logical inconsistencies, and limited awareness of coordination, as noted in~\cite{chen2024solving}. Early attempts of using LLMs for multi-robot flocking show that independently operating LLM-based robots frequently collapse toward a centroid or diverge due to misaligned individual plans, highlighting the need for structured negotiation mechanisms to align decentralized reasoning~\cite{li2024challenges}. 

\textbf{Positioning of This Work.}  
To address these challenges, we propose LLM-Flock, a framework that integrates LLM-based local planning with an influence-based decentralized consensus protocol. By coupling prompt-driven reasoning with lightweight local negotiation, our approach enables multi-robot teams to dynamically converge toward coherent flocking formations without centralized control or extensive retraining. In contrast to prior work that relies solely on unstructured LLM outputs~\cite{li2024challenges}, LLM-Flock explicitly grounds decentralized coordination by assigning each robot an LLM for plan generation and decision-making through structured, influence-based plan refinement, enabling scalable and robust flocking behaviors.

\section{Methods}
\label{sec:methods}
We propose LLM-Flock, a framework for decentralized multi-robot flocking that integrates LLM-driven planning with an influence-based consensus mechanism \PL{validated by a two-layer collision avoidance mechanism}, as shown in Fig.~\ref{fig:llm-flock}. The framework enables a team of robots to autonomously achieve and maintain a coherent formation without centralized control. Each robot is individually equipped with an LLM that acts as its local planner and decision-maker. The interaction between the robot and its LLM is structured through two types of messages: (i) a one-time plan generation message at the beginning of the mission, and (ii) periodic position update messages during motion execution. The communication of these two message types guides the LLM to reason about both the initial formation objective and the dynamic local adjustments throughout the mission.

LLM-Flock operates through \PL{four} key stages: (1) individual plan generation, (2) influence-based, decentralized plan consensus, 
\PL{(3) individual waypoints generation based on the local information, }and (4) coordinated motion execution with two-layer collision avoidance. The plan generation stage is executed once at mission initialization, after which the framework alternates periodically between plan consensus and motion execution. Each stage is detailed below.

\subsection{Individual Plan Generation}
We consider a team of \textit{N} mobile robots, indexed as $i \in \{1, \cdots, N\}$. Each robot is initialized randomly in the environment at $\mathbf{x}_i(0) \in \mathbb{R}^2$. To guide the LLMs during the planning phase, each robot is initialized with a system message that defines the task context, environmental constraints, and expected behavior. The system message explicitly describes the objective of forming a specified geometric formation, the inter-robot distance requirements, the safe distance constraint, the maximum allowed movement per round, and the communication range. An example system message template is shown below, 

\begin{tcolorbox}[colback=cyan!5!white, colframe=cyan!70!white, breakable, enhanced, title=\textbf{[SYSTEM]}]
\small
\texttt{You are a drone navigating in a 2D space. Your objective is to determine your next position to contribute to forming a shape with your neighbors while maintaining specific distance constraints. Your neighbors are also moving.}

\texttt{Key Requirements:}
\begin{itemize}
    \item \texttt{Formation: Form a/an \{shape\} centered at \{$\mathbf{x}_c$\}.}
    \item \texttt{Desired Distance: Maintain a desired distance \{$d_d$\} between each drone.}
    \item \texttt{Safe Distance: Keep a minimum safe distance \{$d_s$\} from other drones.}
    \item \texttt{Maximum Speed: Your movement per step cannot exceed \{$v_m$\} units.}
    \item \texttt{Communication Range: Your communication range is \{$r_{comm}$\}.}
\end{itemize}

\end{tcolorbox}

where mission-specific parameters such as desired formation shape, center location $\mathbf{x}_c\in\mathbb{R}^2$, desired inter-robot distance $d_d$, minimum safe distance between robots $d_s$, maximum velocity the robot can move in each round $v_m$, and the communication radius for the robot $r_{comm}$ are dynamically filled based on the task specification.

After receiving the system message, each robot initiates the mission by sending an initial \textit{plan generation user message} to its LLM. This message instructs the LLM to propose a full set of target positions for all robots that satisfy the specified formation constraints. In addition to the overall plan, the LLM is also required to assign the robot’s own role by indicating which specific position it should take within the formation. The plan generation user message template is detailed below.
\begin{tcolorbox}[colback=cyan!5!white, colframe=cyan!70!white, title=\textbf{[USER - Plan Generation]}]
\small
\texttt{Please make a plan of the locations for the team of \{$N$\} robots forming the shape. Remember the given requirements about the shape and desired distance. Consider "\textbf{urgency}" to resolve the collision avoidance. Lower urgency gives the way. Please give the final answer in the form of JSON format: \{"plan": [[x1, y1], [x2, y2], $\ldots$, [xn, yn]], "my\_plan": a, "reasoning": "..."\}, where a is the index of the location you are taking.}
\end{tcolorbox}
This message provides the robot with the team size of $N$ and reiterates the formation requirements to guide the LLM's reasoning. The LLM is instructed to output a full global plan, denoted as $\mathcal{P}_i = \{ \mathbf{p}_1^i, \mathbf{p}_2^i, \dots, \mathbf{p}_N^i \}$, where each $\mathbf{p}_j^i \in \mathbb{R}^2$ specifies the desired position for robot $j$ according to robot $i$'s local planning. In addition to generating the global plan, the LLM must assign the robot's own role within the formation by indicating the index of the position it will take, i.e., $\mathbf{g}_i = \mathcal{P}_i[a]$. The output format is explicitly defined to ensure accurate and consistent information extraction across all robots. A sample response from the LLM generating the plan is shown below.





\begin{tcolorbox}[colback=brown!10!white, colframe=brown!70!white, enhanced, breakable, title=\textbf{[ASSISTANT]}]
\small
\texttt{To achieve an equilateral triangle formation centered at [50, 50] with a side length of 10 units, we can determine the triangle's vertices $\cdots$}



\vspace{0.5em}
\texttt{\{"plan": [[55.77, 50], [47.11, 55], [47.11, 45]], "my\_plan": 0\}}
\end{tcolorbox}

After individual plan generation, the system proceeds to the decentralized consensus phase, where robots negotiate and align on a common formation plan.

\subsection{Influence-based Decentralized Plan Consensus}
After the individual plan generation, each robot possesses a locally generated formation plan. However, coherent global formation cannot be achieved if robots pursue conflicting plans. Thus, a decentralized consensus process is necessary to synchronize the team toward a unified plan. To facilitate decentralized negotiation, we introduce the notion of influence, which quantifies a robot’s communication centrality within the network. Intuitively, a robot with more immediate neighbors is considered more influential, as it can propagate its plan more rapidly across the team. The influence from robot $i$ at time $t$ is defined as 
\begin{equation}
        \textit{I}_i(t)=\frac{|\mathcal{N}_i(t)|}{N},
\end{equation}
where $\mathcal{N}_i(t) = \{ j \in \{1, \dots, N\} \setminus \{i\} \mid \|\mathbf{x}_i(t) - \mathbf{x}_j(t)\| \leq r_{comm} \}$ denotes the set of neighboring robots within communication range $r_{comm}$, and $N$ is the total number of robots.

At each consensus round, every robot broadcasts its current formation plan $\mathcal{P}_i$ and computed influence score $I_i(t)$ to its neighboring robots $\mathcal{N}_i(t)$. Upon receiving plans and influence scores from its neighbors, robot $i$ identifies the neighbor $k$ with the highest influence score $I_k(t)$. If the neighbor’s influence is greater than its own, i.e., $I_k(t) > I_i(t)$, and the neighbor’s plan differs from its current plan ($\mathcal{P}_k \neq \mathcal{P}_i$), robot $i$ adopts the neighbor’s plan $\mathcal{P}_i \leftarrow \mathcal{P}_k$.

After adopting a new plan, each robot dynamically assigns itself a goal position from the adopted plan based on proximity. Goal assignment is conducted in a decentralized greedy fashion: each robot selects the closest available goal position among the unassigned waypoints in the plan. Once a goal position is assigned to a robot, it becomes unavailable for others. Specifically, robot $i$ selects its goal $\mathbf{g}_i$ according to:
\begin{equation}
    \mathbf{g}_i = \arg\min_{\mathbf{p} \in \mathcal{P}_i^{\text{avail}}} \|\mathbf{x}_i(t) - \mathbf{p}\|,
\end{equation}
where $\mathcal{P}_i^{\text{avail}}$ denotes the set of unassigned goal positions remaining in the plan $\mathcal{P}_i$, and $\mathbf{x}_i(t)$ is the robot’s current position. This nearest-available-goal assignment ensures a smooth transition during plan updates, minimizes unnecessary movement, and promotes stability throughout the formation process. If no neighbor possesses a higher influence score, the robot retains its current plan.

This decentralized negotiation process is repeated periodically, with robots dynamically alternating between consensus negotiation and motion execution. Over successive rounds, high-influence plans naturally propagate through the network, enabling the team to progressively refine and converge toward a shared formation plan without requiring global communication or centralized synchronization. The formal influence-based consensus procedure is summarized in Algorithm~\ref{alg:consensus}.

\begin{algorithm}[h]
\caption{Influence-Based Plan Consensus}
\label{alg:consensus}
\KwIn{Initial plans $\mathcal{P}_i$, neighbor sets $\mathcal{N}_i(t)$, influence scores $I_i(t)$ for all $i \in \{1, \dots, N\}$}
\KwOut{Updated plans $\mathcal{P}_i$ and assigned goals $\mathbf{g}_i$ for all robots}
\ForAll{robot $i$ \textbf{in parallel}}{
    Receive $\mathcal{P}_j$ and $I_j(t)$ from all $j \in \mathcal{N}_i(t)$\;
    $k \leftarrow \arg\max_{j \in \mathcal{N}_i(t)} I_j(t)$\;
    \If{$I_k(t) > I_i(t)$ \textbf{and} $\mathcal{P}_k \neq \mathcal{P}_i$}{
        $\mathcal{P}_i \leftarrow \mathcal{P}_k$\;
        $\mathcal{P}_i^{\text{avail}} \leftarrow \mathcal{P}_i$\;
        $\mathbf{g}_i \leftarrow \arg\min_{\mathbf{p} \in \mathcal{P}_i^{\text{avail}}} \|\mathbf{x}_i(t) - \mathbf{p}\|$\;
        $\mathcal{P}_i^{\text{avail}} \leftarrow \mathcal{P}_i^{\text{avail}} \setminus \{\mathbf{g}_i\}$\;
    }
}
\end{algorithm}
Having established the decentralized plan-consensus mechanism, we now describe how individual robots \PL{generate candidate waypoints toward their assigned goals.}

\subsection{Waypoint Generation} 

\PL{Following the influence-based consensus process, each robot generates a candidate waypoint for the current round. The robot queries its LLM with a \textit{user message} for waypoint generation, which encodes its current state and nearby environmental information. Specifically, the message includes the robot's current position $\mathbf{x}_i \in \mathbb{R}^2$, the set of neighboring robot positions $\mathcal{X}_i = \{\mathbf{x}_j \mid j \in \mathcal{N}_i(t)\}$, and its assigned global formation plan $\mathcal{P}_i$. To ensure robust parsing and minimal ambiguity, the output format is strictly specified as JSON schema. The template of the user message for waypoint generation is shown below.}
 
\begin{tcolorbox}[colback=cyan!5!white, colframe=cyan!70!white, enhanced, breakable, title=\textbf{[USER - Waypoint Generation]}]
\small
\texttt{Current Position: \{$\mathbf{x}_i$\}. Moving Neighbor Positions: \{$\mathbf{X}_i$\}. Plan: \{$\mathcal{P}_i$\}. You will need to go to \{$\mathbf{g}_i$\} as your final destination.}

\texttt{Task: Decide your next position considering the constraints and the neighbor locations. Briefly explain your decision and provide new position in JSON format: \{"position": [x, y], "reasoning": "..."\}}
\end{tcolorbox}
\PL{Upon receiving the query of position update, the LLM reasons about the robot's next waypoint $\mathbf{w}_i(t) \in \mathbb{R}^2$ toward its assigned goal $\mathbf{g}_i$, accounting for the updated neighbor configuration and formation constraints.} A sample response of an LLM choosing the next-step position based on the updated locations of the neighbors is shown below.

    
    


\begin{tcolorbox}[colback=brown!10!white, colframe=brown!70!white, enhanced, breakable, title=\textbf{[ASSISTANT RESPONSE]}]
\small
\texttt{I'm currently at [12.63, 70.48] with my target being [55.77, 50]. The straight-line$\cdots$}


    
    
    
    
    


\vspace{0.5em}
\texttt{\{"position": [18.05, 67.91]\}}
\end{tcolorbox}
\noindent \PL{These candidate waypoints are then verified by the collision avoidance procedure described next.}

\PL{\subsection{Collision Avoidance and Execution}}

Once all robots have generated candidate waypoints, potential path conflicts must be resolved before positions are updated. We introduce a two-layer collision avoidance strategy that combines LLM-based reasoning with programmatic safety verification, summarized in Algorithm~\ref{alg:collision}.

\begin{algorithm}[h]
\caption{Collision Checking and Re-Prompting}
\label{alg:collision}
\KwIn{Current positions $\mathbf{x}_i$, neighbor sets $\mathcal{N}_i(t)$, base prompts $m_i$, safe distance $d_s$, max retries $K$, urgencies $U_i$ for all $i \in \{1, \dots, N\}$}
\KwOut{Safe next waypoints $\mathbf{w}_i$ for robot $i$}
\ForAll{robot $i$ \textbf{in parallel}}{
    $\mathbf{w}_i \leftarrow \textsc{LLM}(m_i)$\;
    \lIf{$\mathbf{w}_i$ is invalid}{$\mathbf{w}_i \leftarrow \mathbf{x}_i$}
}
\For{$k = 1$ \KwTo $K$}{
    $C \leftarrow \textsc{FindCollisionPairs}(\mathcal{N}, \mathbf{x}, \mathbf{w}, d_s)$\;
    \lIf{$C = \emptyset$}{\textbf{break}}
    $Y \leftarrow \textsc{LowerUrgency}(C)$\;
    \ForAll{robot $i \in Y$ \textbf{in parallel}}{
        $j^* \leftarrow \arg\max_{j:(i,j)\in C} U_j$ \tcp*{highest-urgency conflict}
        $\mathbf{w}_i \leftarrow \textsc{LLM}(m_i + \text{collision re-prompt w.r.t. } j^*)$\;
        \lIf{$\mathbf{w}_i$ is invalid}{$\mathbf{w}_i \leftarrow \mathbf{x}_i$}
    }
}
$C \leftarrow \textsc{FindCollisionPairs}(\mathcal{N}, \mathbf{x}, \mathbf{w}, d_s)$\;
\ForAll{$i \in \textsc{LowerUrgency}(C)$}{
    $\mathbf{w}_i \leftarrow \mathbf{x}_i$ \tcp*{fallback: stay stationary}
}
\end{algorithm}

We define the \textit{urgency} of robot $i$ as the Euclidean distance between its current position $\mathbf{x}_i \in \mathbb{R}^2$ and its assigned goal $\mathbf{g}_i \in \mathbb{R}^2$:
\begin{equation}
U_i = \|\mathbf{x}_i - \mathbf{g}_i\|.
\end{equation}
Robots with higher urgency are given priority in conflicts, reflecting their greater need to make progress. Ties are broken deterministically by robot index.

\textbf{Layer 1: Urgency-Based Priority via Prompting.}
The system prompt is extended with instructions requiring each LLM to compute urgency values, compare priorities with neighbors, and reason about potential collisions. User prompts include neighboring robots' goals in addition to their current positions, enabling the LLM to proactively avoid conflicts during waypoint generation.

\textbf{Layer 2: Collision Checking and Re-Prompting.}
Since LLM spatial reasoning is not always reliable, a programmatic collision checker is employed to verify all planned trajectories after each robot generates its next waypoint. A conflict between robots $i$ and $j$ is detected if (i) their proposed next positions violate $\|\mathbf{w}_i - \mathbf{w}_j\| \leq d_s$, or (ii) their motion segments from current to proposed positions intersect within distance $d_s$. When a violation is detected, the lower-urgency robot is re-prompted using the identity and intended position of its highest-urgency conflicting neighbor. Re-prompting is attempted up to $K$ times (we use $K=3$). If a collision-free waypoint is still not produced, the lower-urgency robot remains stationary for the current round, ensuring collision avoidance through a deterministic fallback. The re-prompting template is shown below.

\begin{tcolorbox}[colback=cyan!5!white, colframe=cyan!70!white, enhanced, breakable, title=\textbf{[USER - Collision Re-Prompt]}]
\small
\texttt{Safety Checker Feedback:}\\
\texttt{Your proposed trajectory [\{$x_i$\}, \{$y_i$\}] -> [\{$w^x_i$\}, \{$w^y_i$\}] is not collision-free.}\\
\texttt{Required minimum separation: \{$d_s$\} units.}\\[4pt]
\texttt{Conflicts detected:}
\begin{itemize}
    \item \texttt{With Robot \{$j$\}: min predicted distance = \{$d_{ij}$\}}
\end{itemize}
\texttt{Fixed trajectories to avoid:}
\begin{itemize}
    \item \texttt{Robot \{$j$\}: [\{$x_j$\}, \{$y_j$\}] -> [\{$w^x_j$\}, \{$w^y_j$\}]}
\end{itemize}
\texttt{Return a NEW next position that is collision-free while keeping the same JSON schema.}
\end{tcolorbox}

In Algorithm~\ref{alg:collision}, \textsc{FindCollisionPairs} returns all pairs $(i,j)$ with $j \in \mathcal{N}_i(t)$ whose planned trajectories violate the safe distance, and \textsc{LowerUrgency} selects the lower-urgency robot from each pair.
\PL{After the collision avoidance procedure completes, all robots
execute their verified waypoints as the next positions,
and the framework returns to the consensus stage for the
next round.}

\PL{In the following section, we evaluate the framework through simulations and real-world experiments.}

\section{Results}
\label{sec:results}
\begin{figure}[h]
    \centering
    \subfigure[Step 0]{\includegraphics[width=0.22\linewidth]{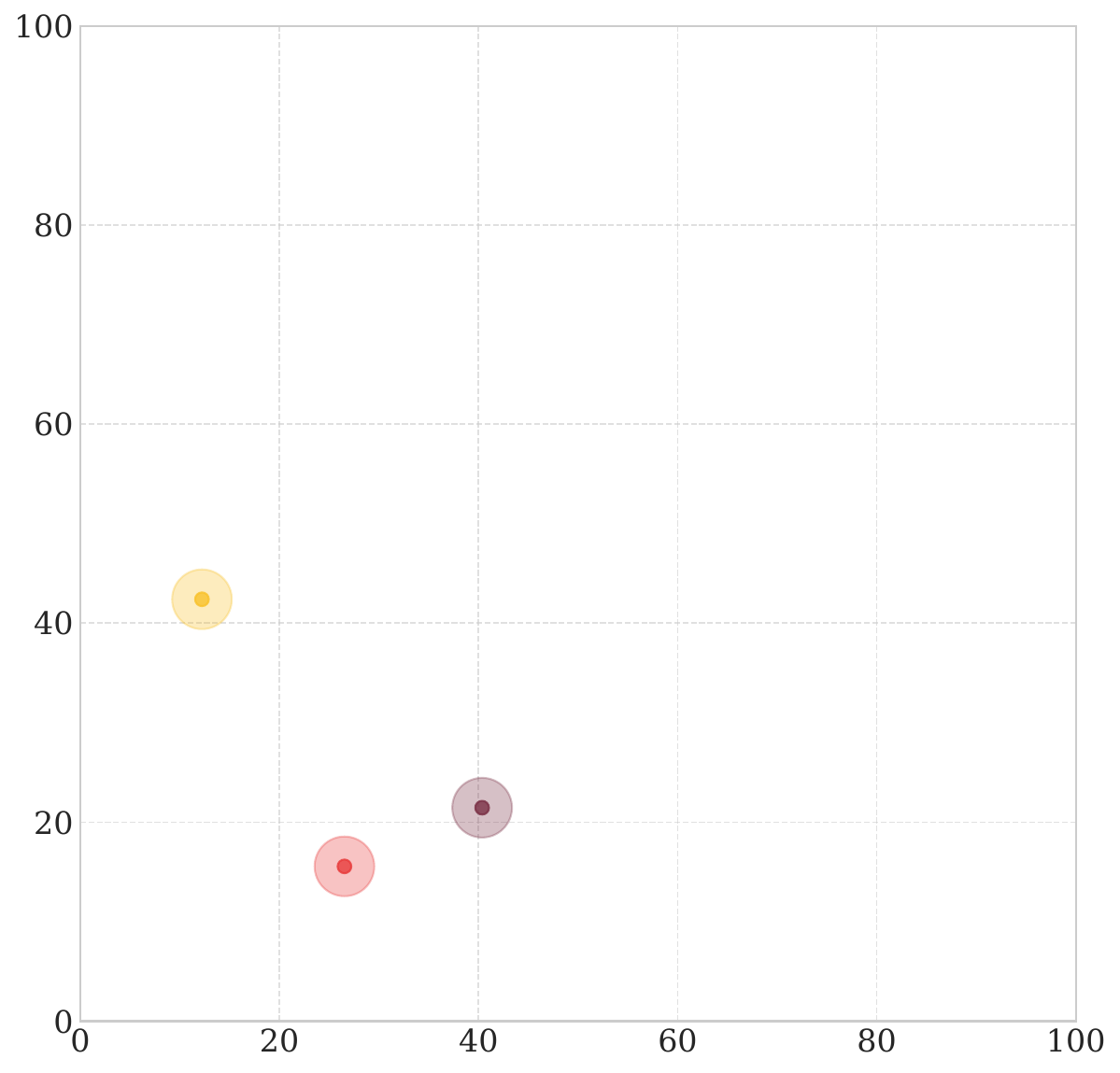}\label{fig:quali-tri3-a}}
    \subfigure[Step 4]{\includegraphics[width=0.22\linewidth]{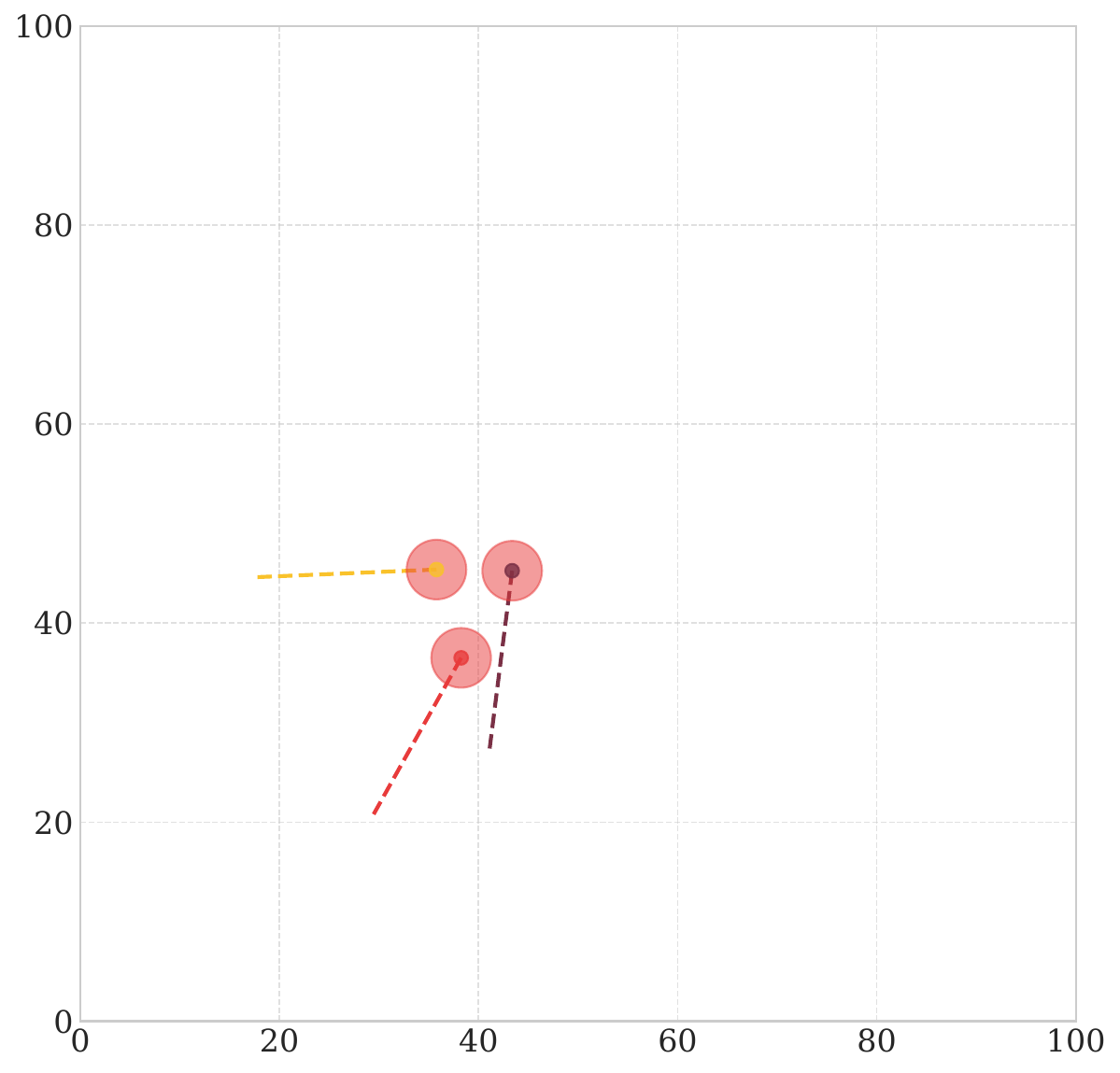}\label{fig:quali-tri3-b}}
    \subfigure[Step 7]{\includegraphics[width=0.22\linewidth]{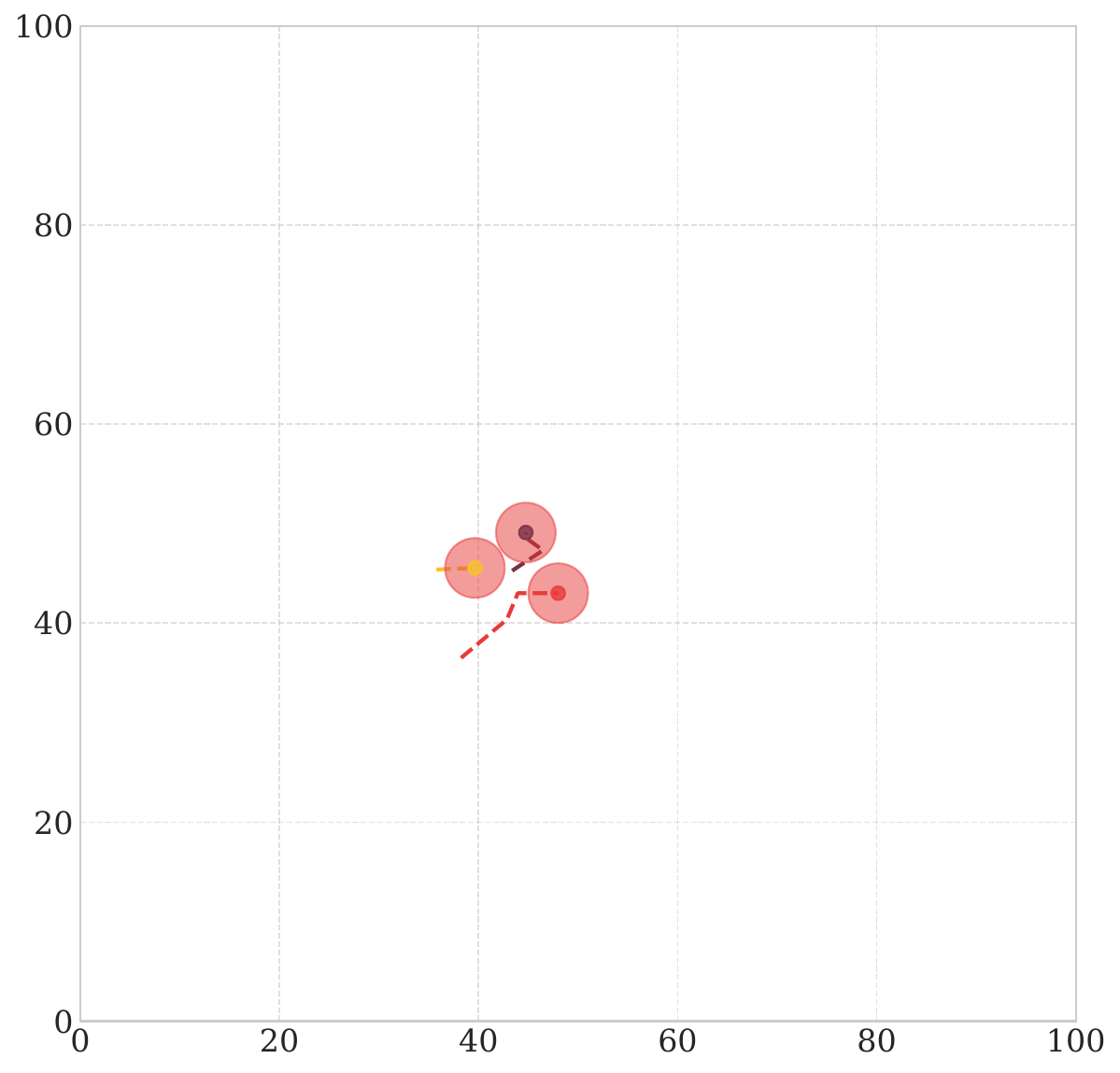}\label{fig:quali-tri3-c}}
    \subfigure[Step 11]{\includegraphics[width=0.22\linewidth]{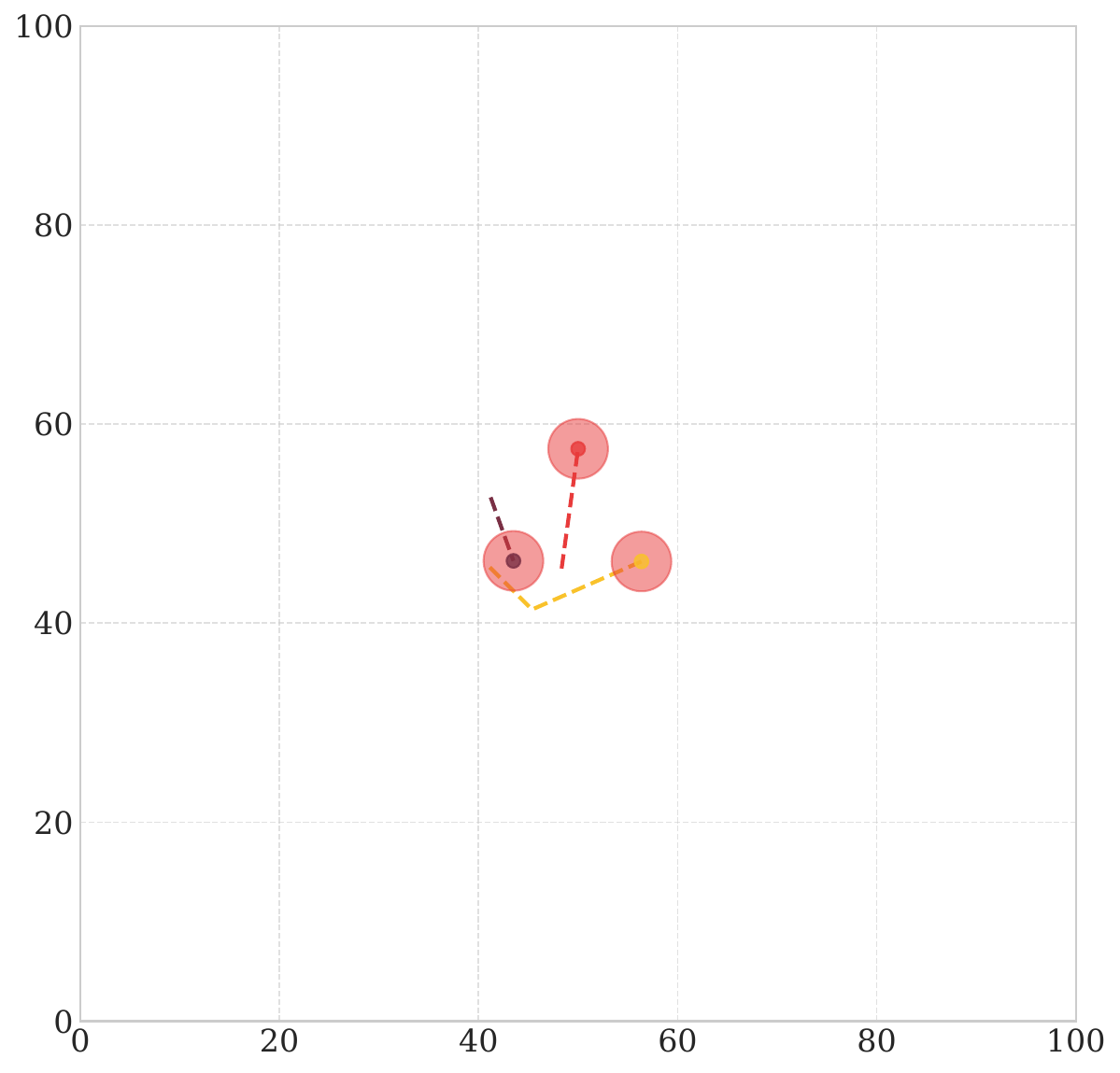}\label{fig:quali-tri3-d}}
    \subfigure[Step 0]{\includegraphics[width=0.22\linewidth]{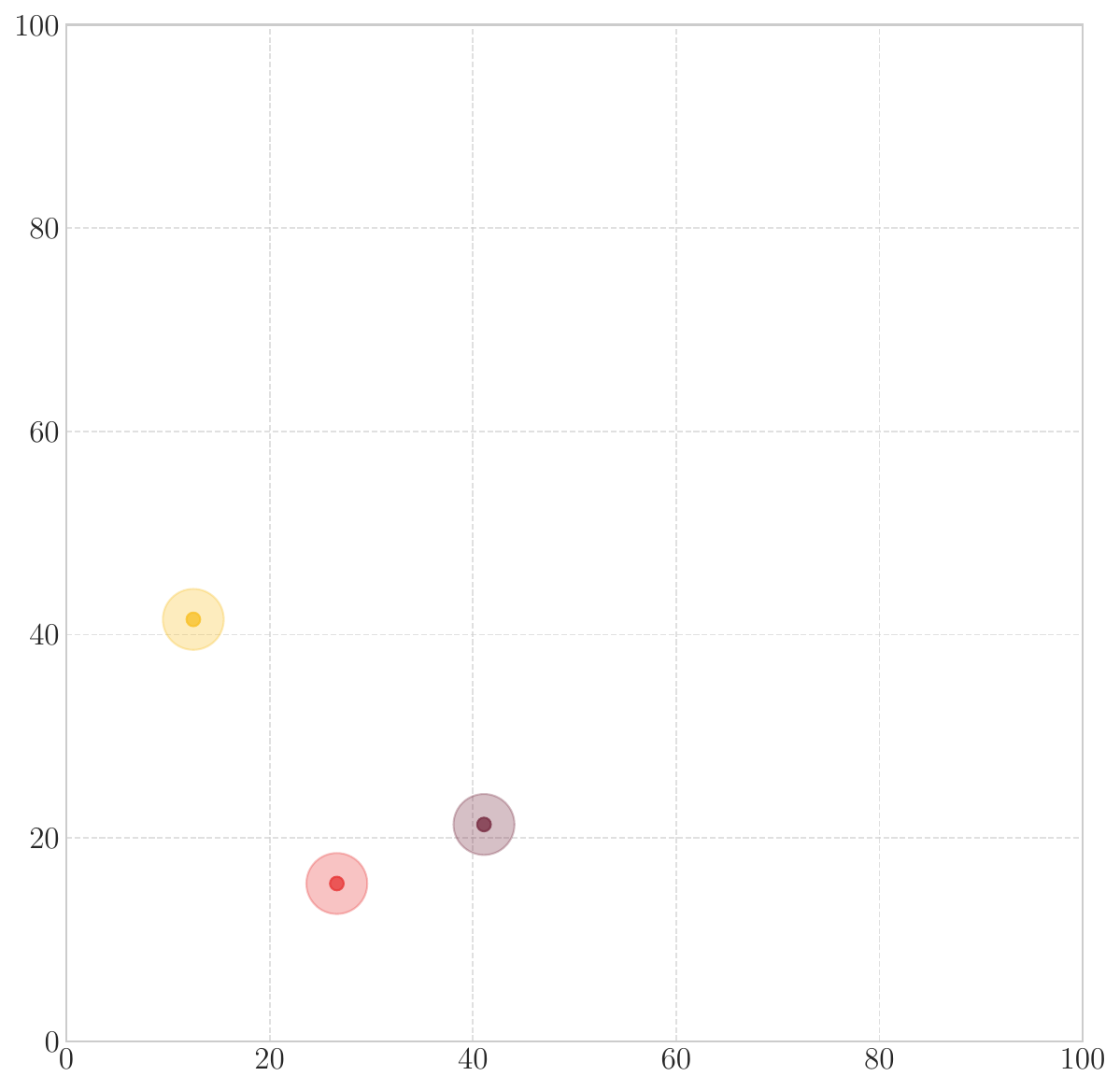}\label{fig:quali-tri3-e}}
    \subfigure[Step 4]{\includegraphics[width=0.22\linewidth]{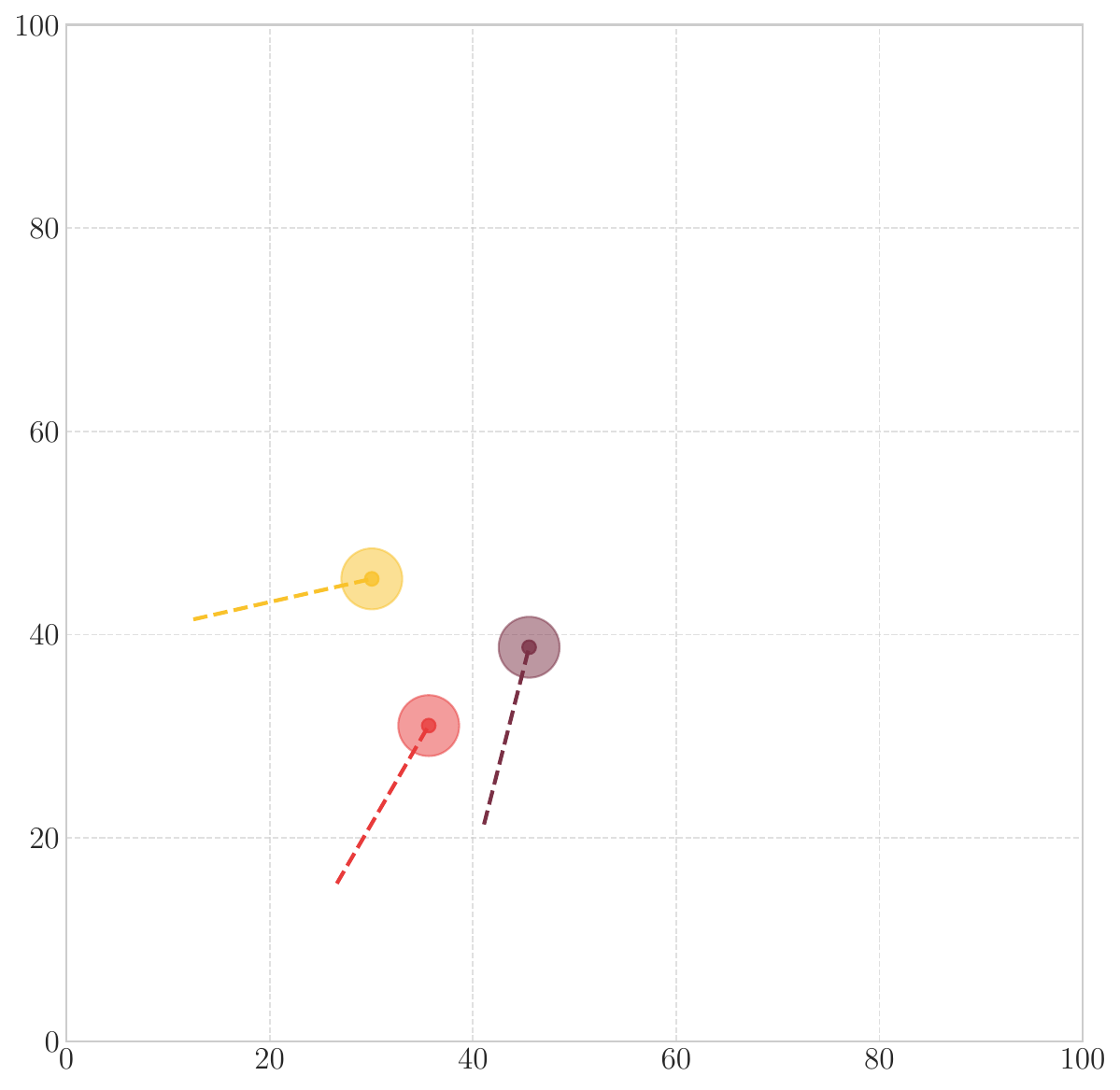}\label{fig:quali-tri3-f}}
    \subfigure[Step 7]{\includegraphics[width=0.22\linewidth]{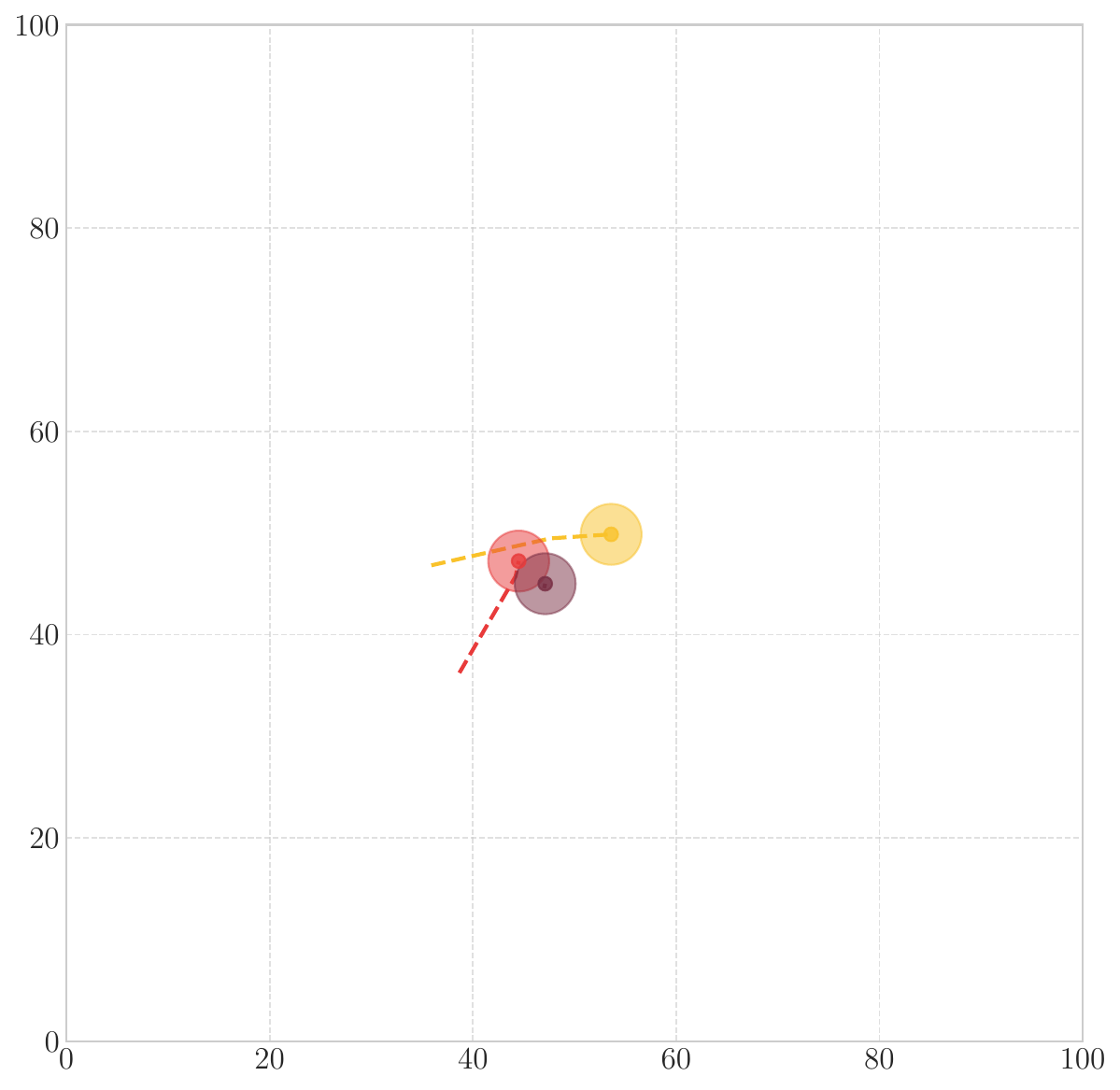}\label{fig:quali-tri3-g}}
    \subfigure[Step 11]{\includegraphics[width=0.22\linewidth]{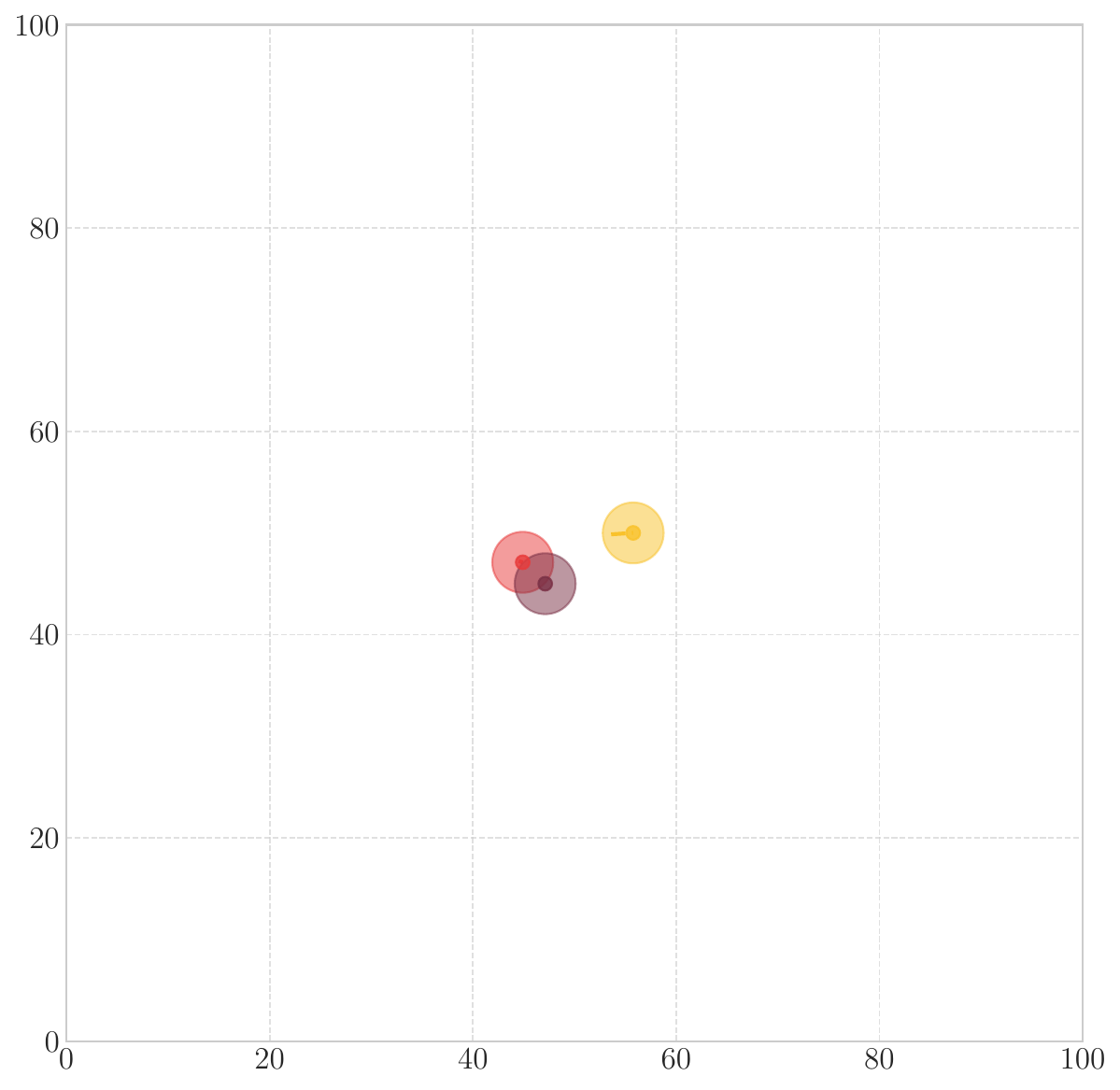}\label{fig:quali-tri3-h}}
    \caption{Comparison of formation results with and without influence-based consensus. (a)-(d) show robot trajectories using influence-based plan consensus with three robots forming triangle formations, and (e)-(h) show the trajectories without consensus.}
    \label{fig:quali-baseline}
\end{figure}

\begin{figure}[!h]
    \centering
    
    \subfigure[Step 0]{\includegraphics[width=0.22\linewidth]{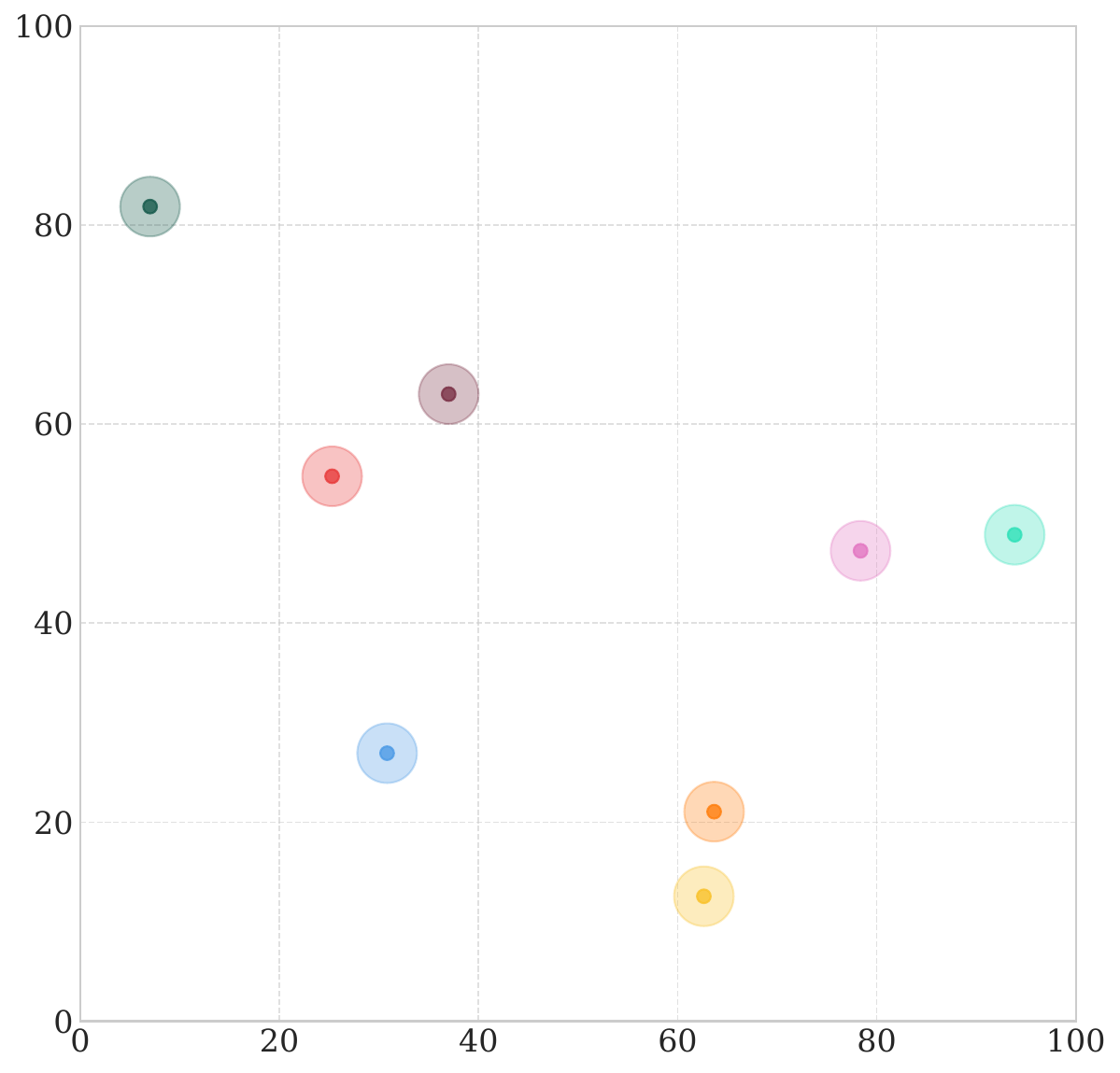}\label{fig-append:quali-square-a}}
    \subfigure[Step 4]{\includegraphics[width=0.22\linewidth]{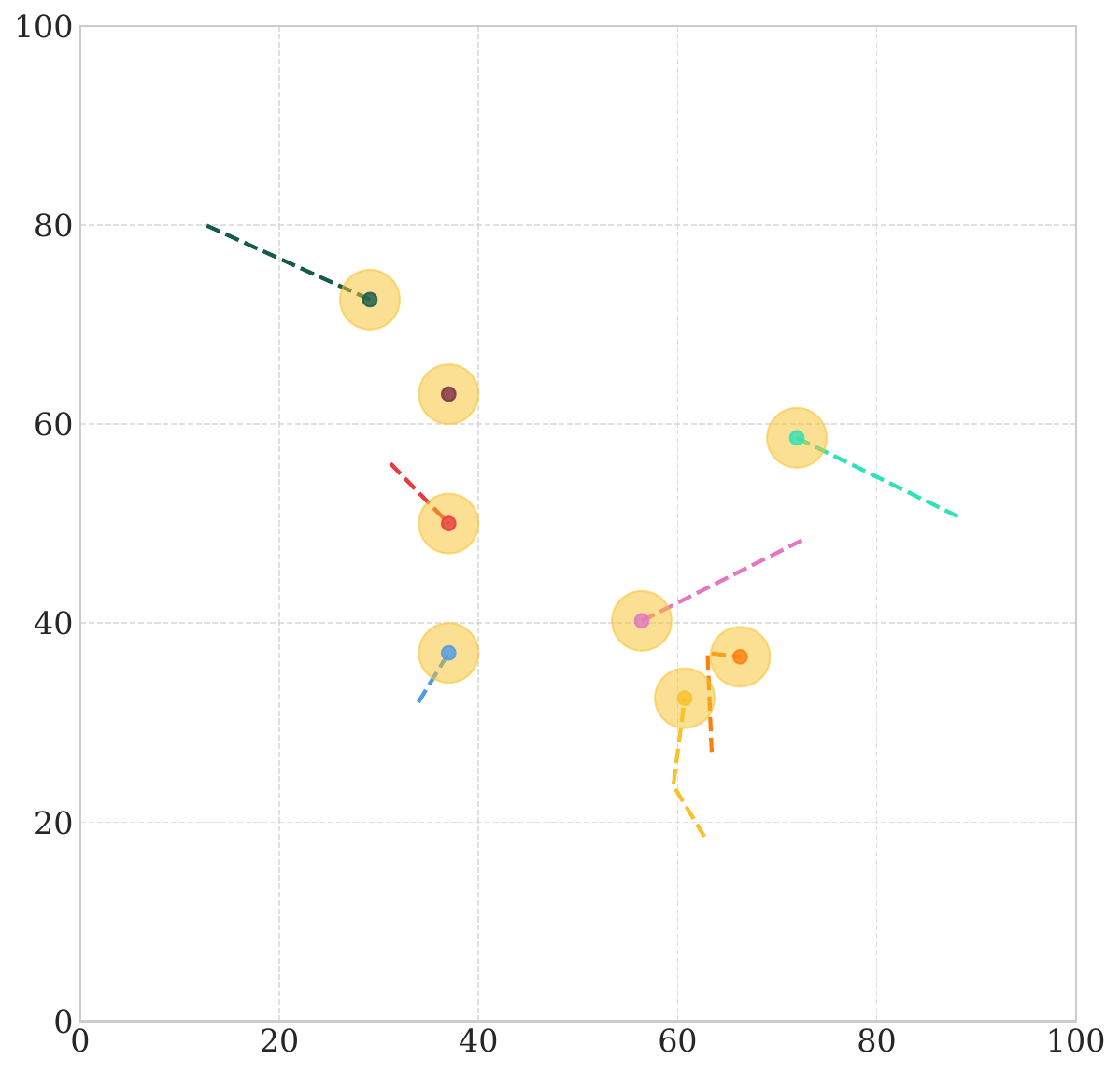}\label{fig-append:quali-square-b}}
    \subfigure[Step 7]{\includegraphics[width=0.22\linewidth]{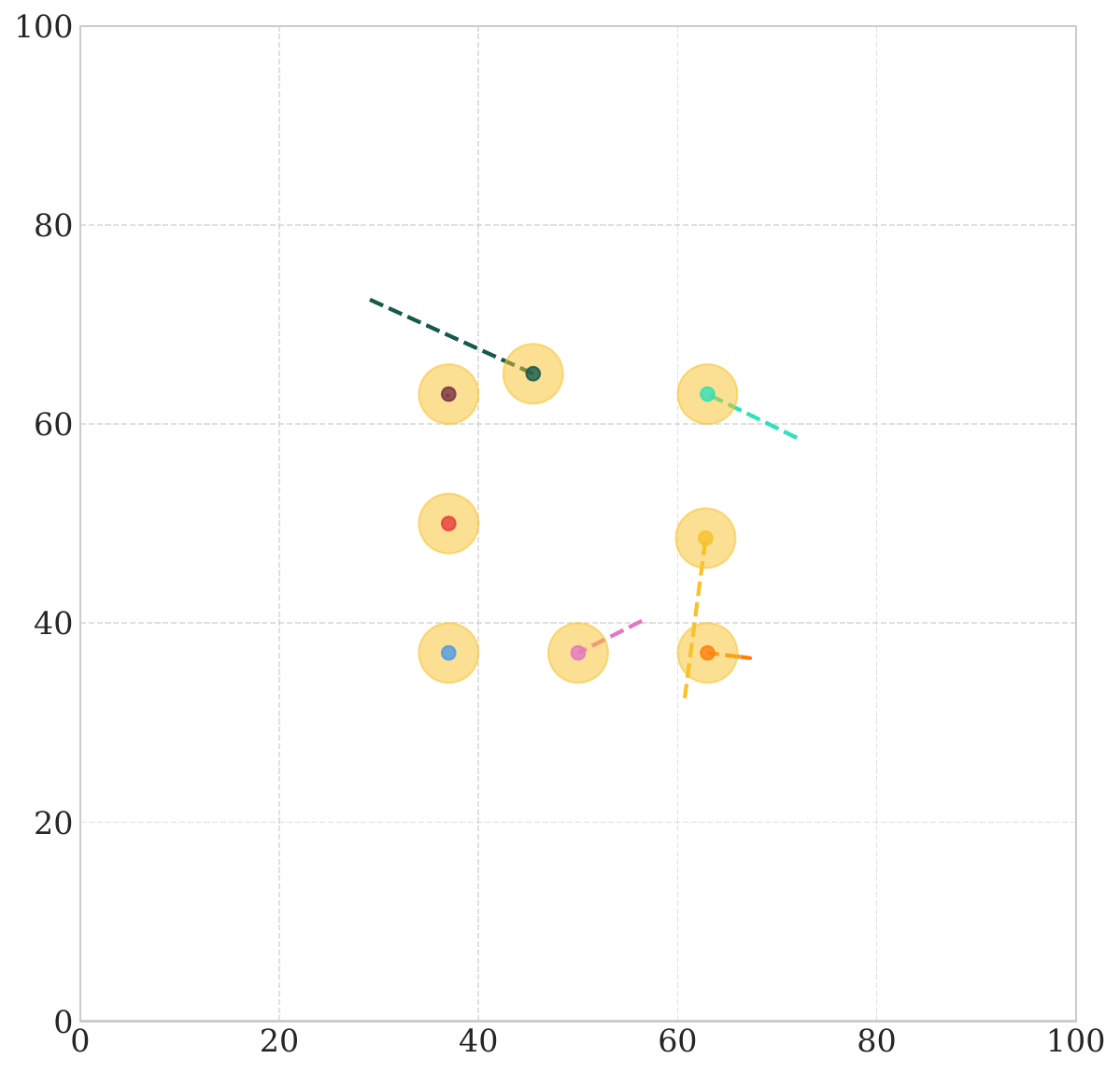}\label{fig-append:quali-square-c}}
    \subfigure[Step 10]{\includegraphics[width=0.22\linewidth]{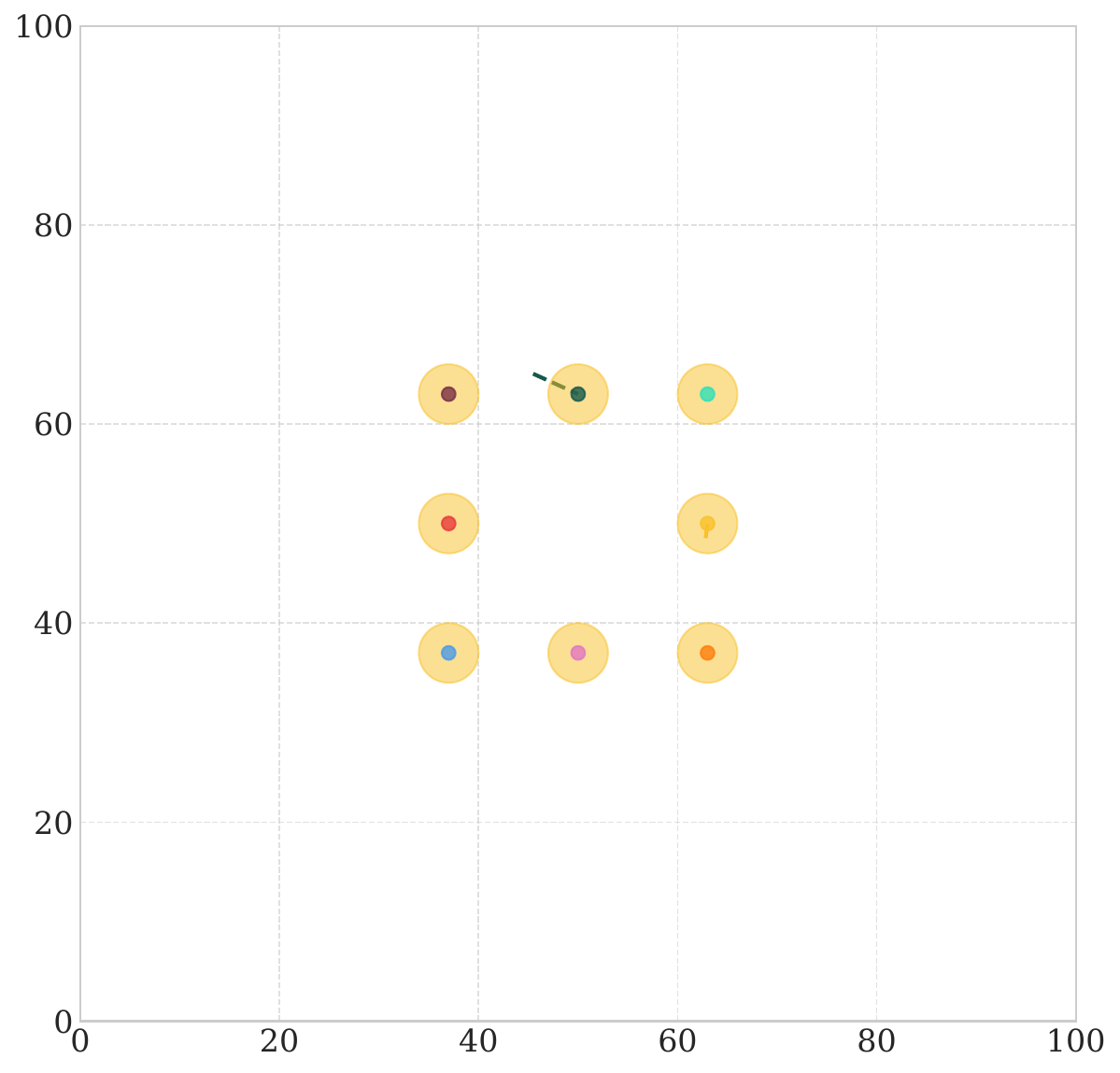}\label{fig-append:quali-square-d}}

    \subfigure[Step 0]{\includegraphics[width=0.22\linewidth]{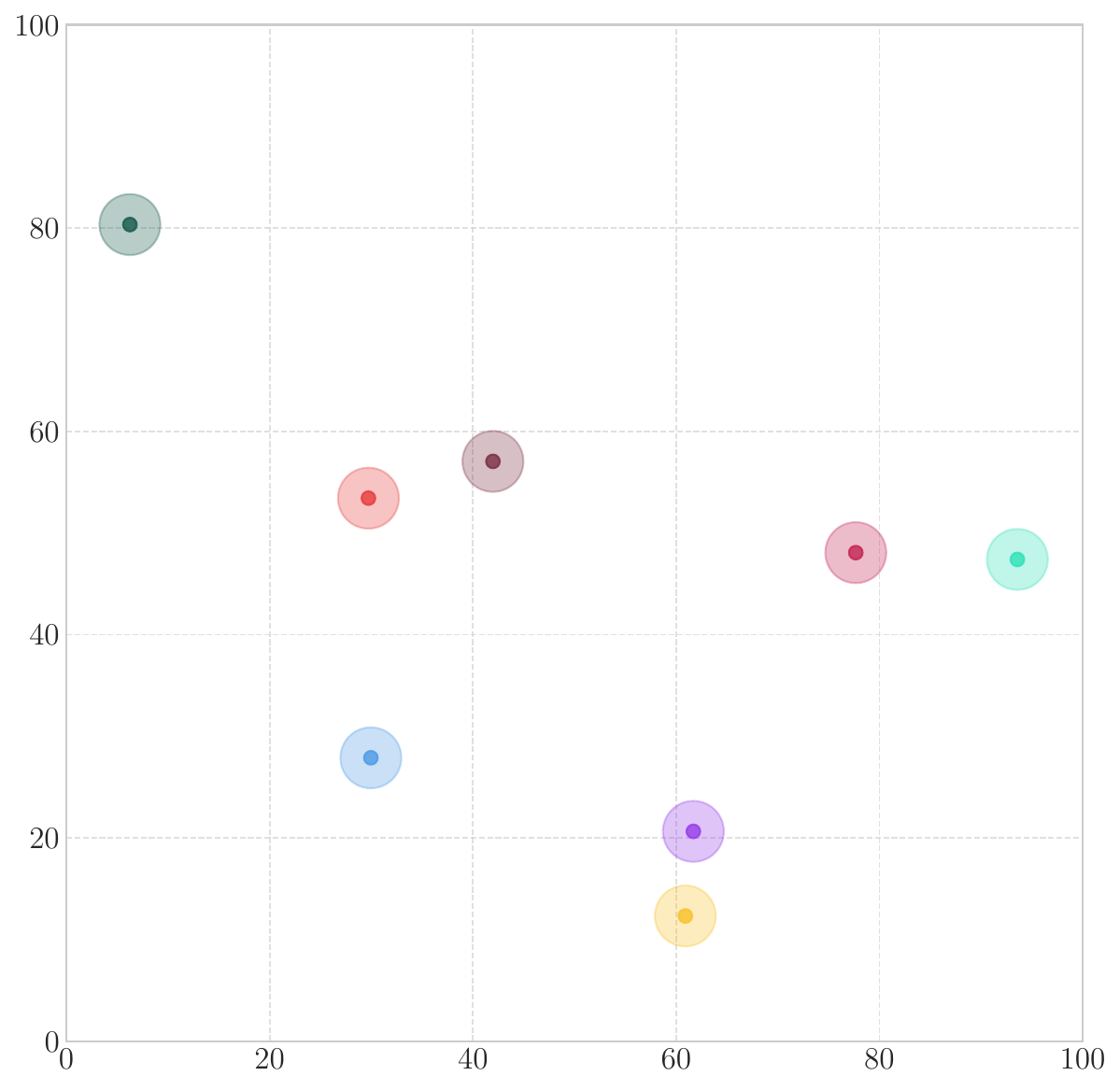}\label{fig-append:quali-square-e}}
    \subfigure[Step 4]{\includegraphics[width=0.22\linewidth]{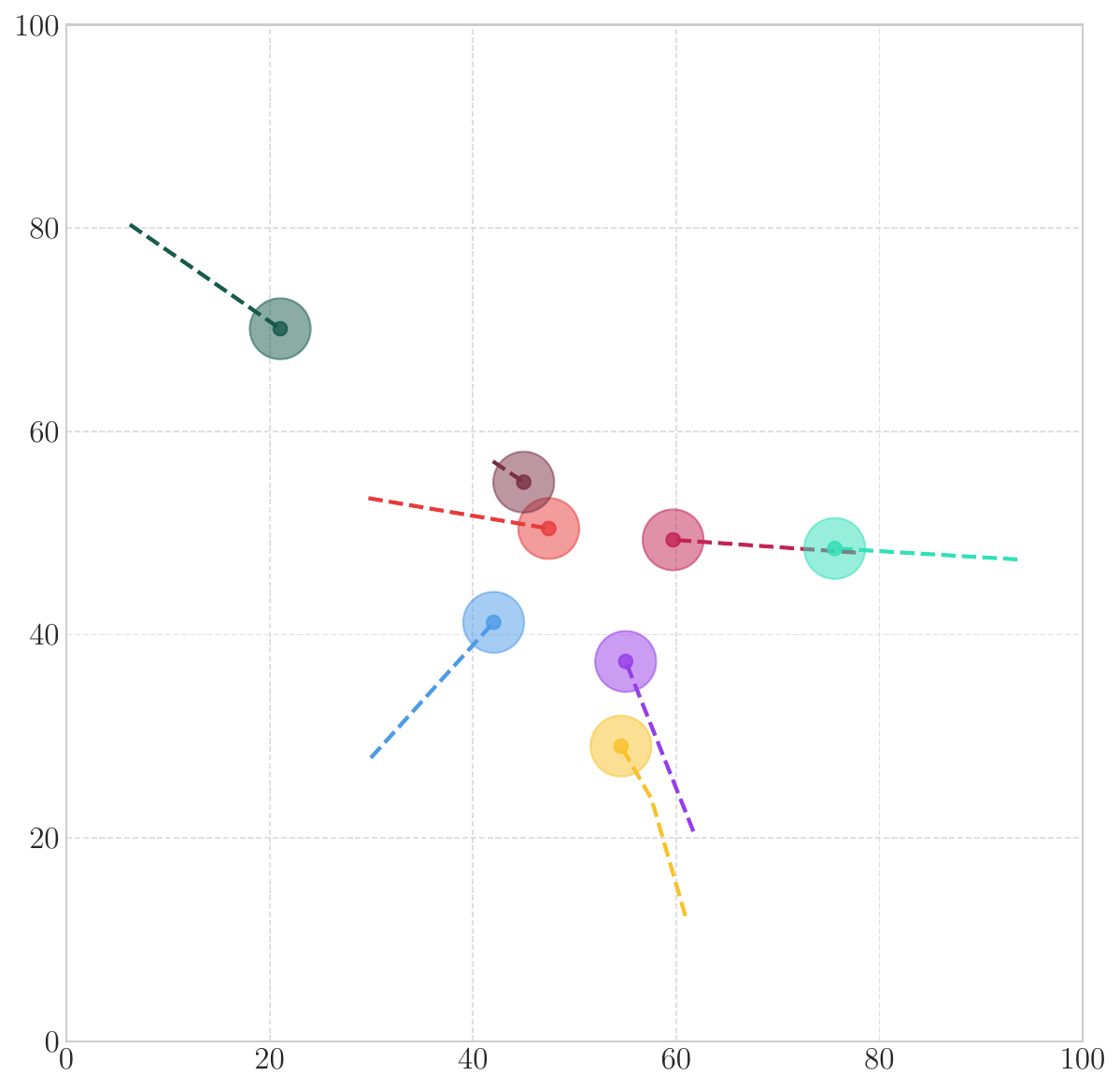}\label{fig-append:quali-square-f}}
    \subfigure[Step 7]{\includegraphics[width=0.22\linewidth]{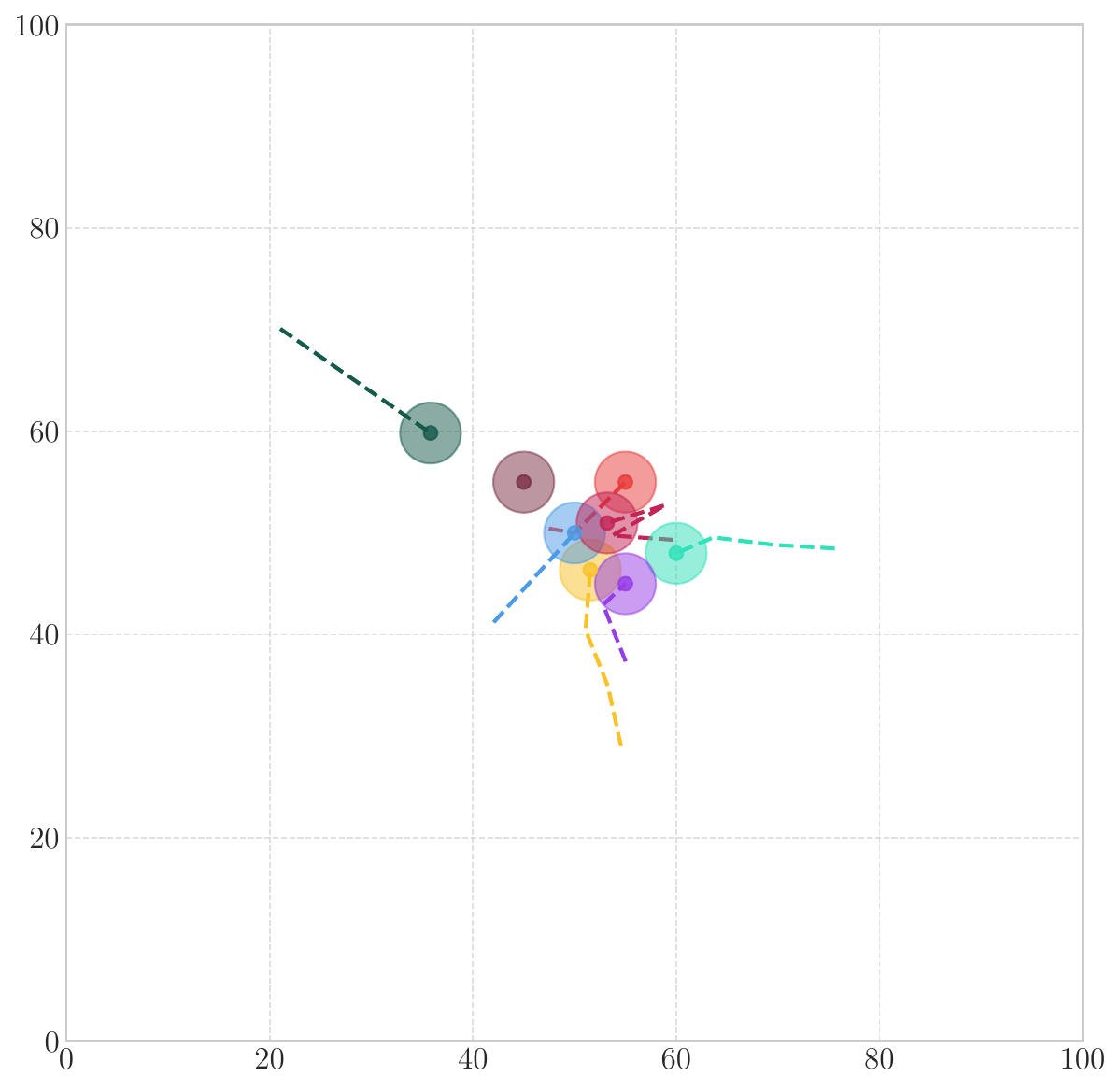}\label{fig-append:quali-square-g}}
    \subfigure[Step 10]{\includegraphics[width=0.22\linewidth]{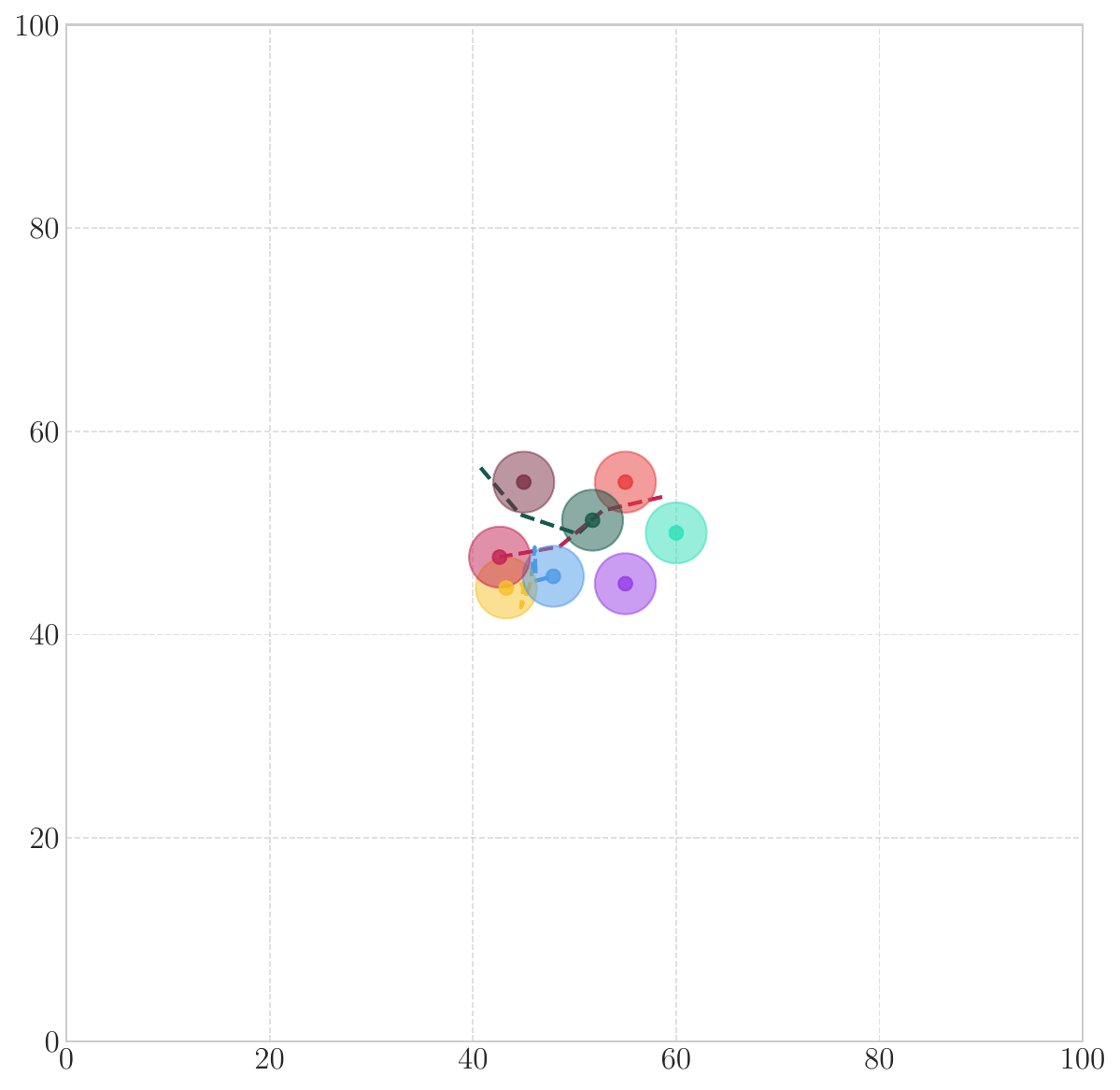}\label{fig-append:quali-square-h}}
    \caption{Eight robots forming a square. (a)–(d): Influence-based plan consensus leads to a uniformly distributed square. (e)–(h): Baseline planning without consensus yields scattered and misaligned placements.}

    \label{fig-append:quali-square}
\end{figure}

\begin{figure}[!h]
    \centering
    \subfigure[Step 0]{\includegraphics[width=0.22\linewidth]{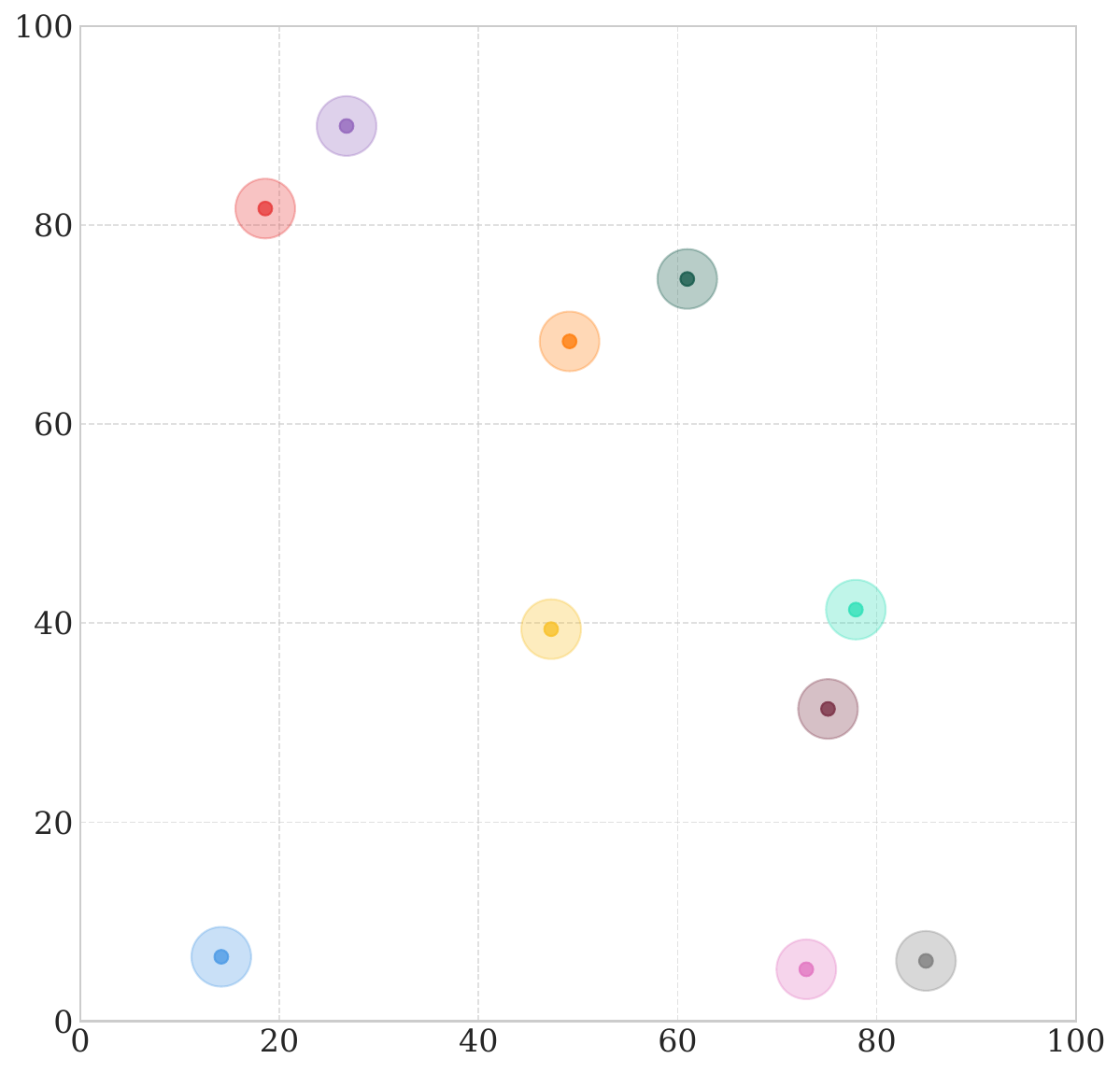}\label{fig-append:quali-circle-a}}
    \subfigure[Step 4]{\includegraphics[width=0.22\linewidth]{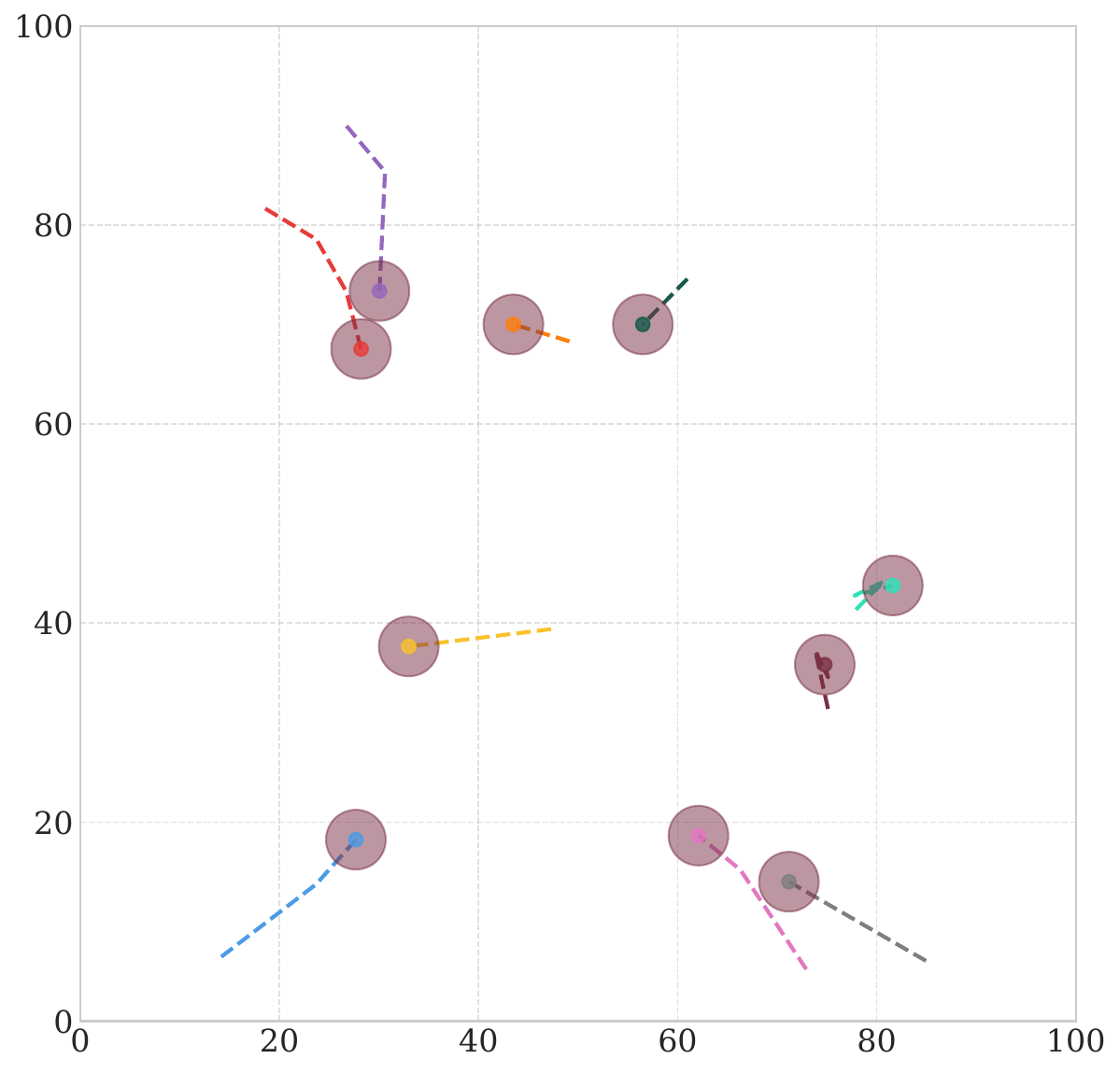}\label{fig-append:quali-circle-b}}
    \subfigure[Step 7]{\includegraphics[width=0.22\linewidth]{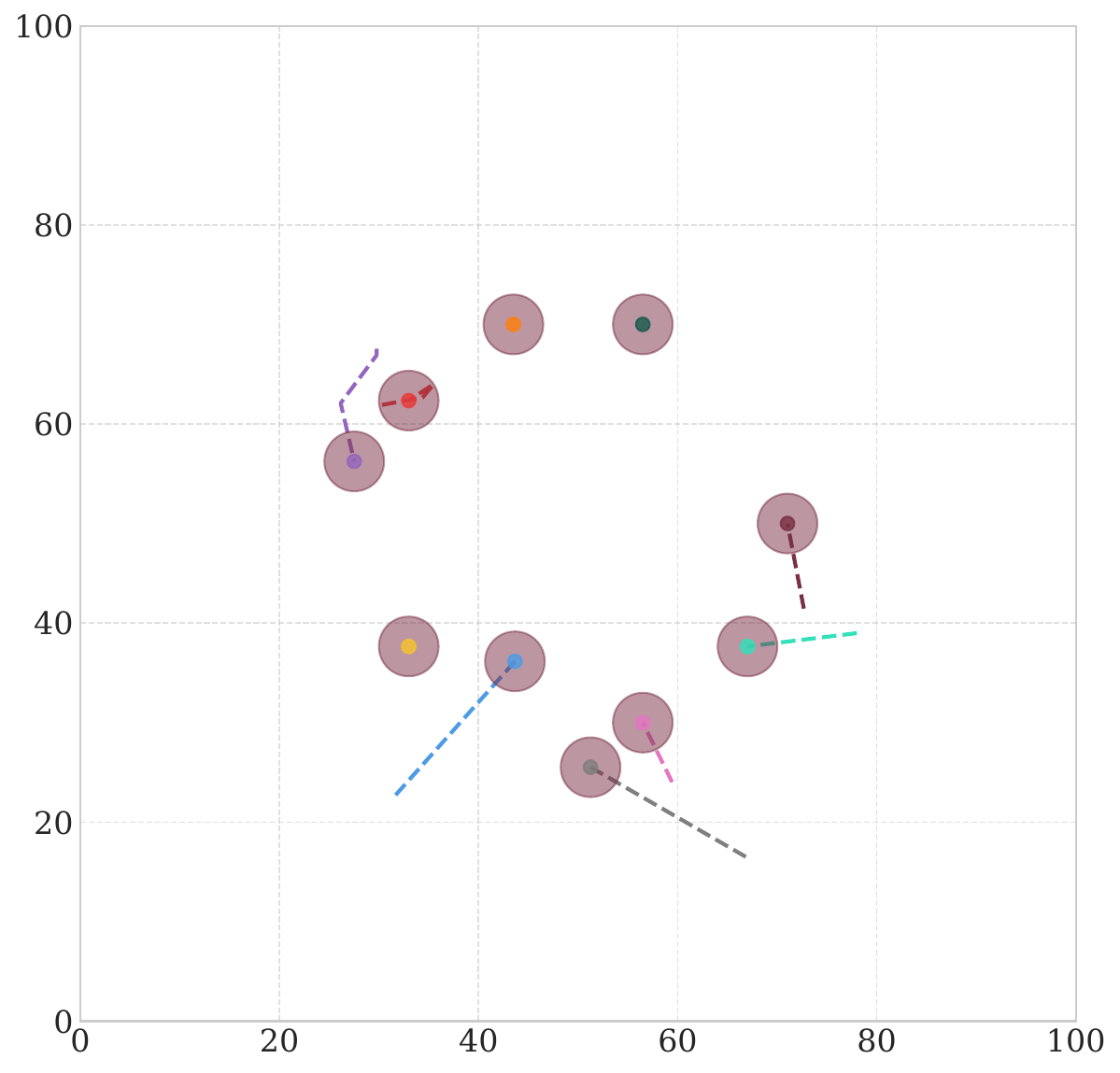}\label{fig-append:quali-circle-c}}
    \subfigure[Step 10]{\includegraphics[width=0.22\linewidth]{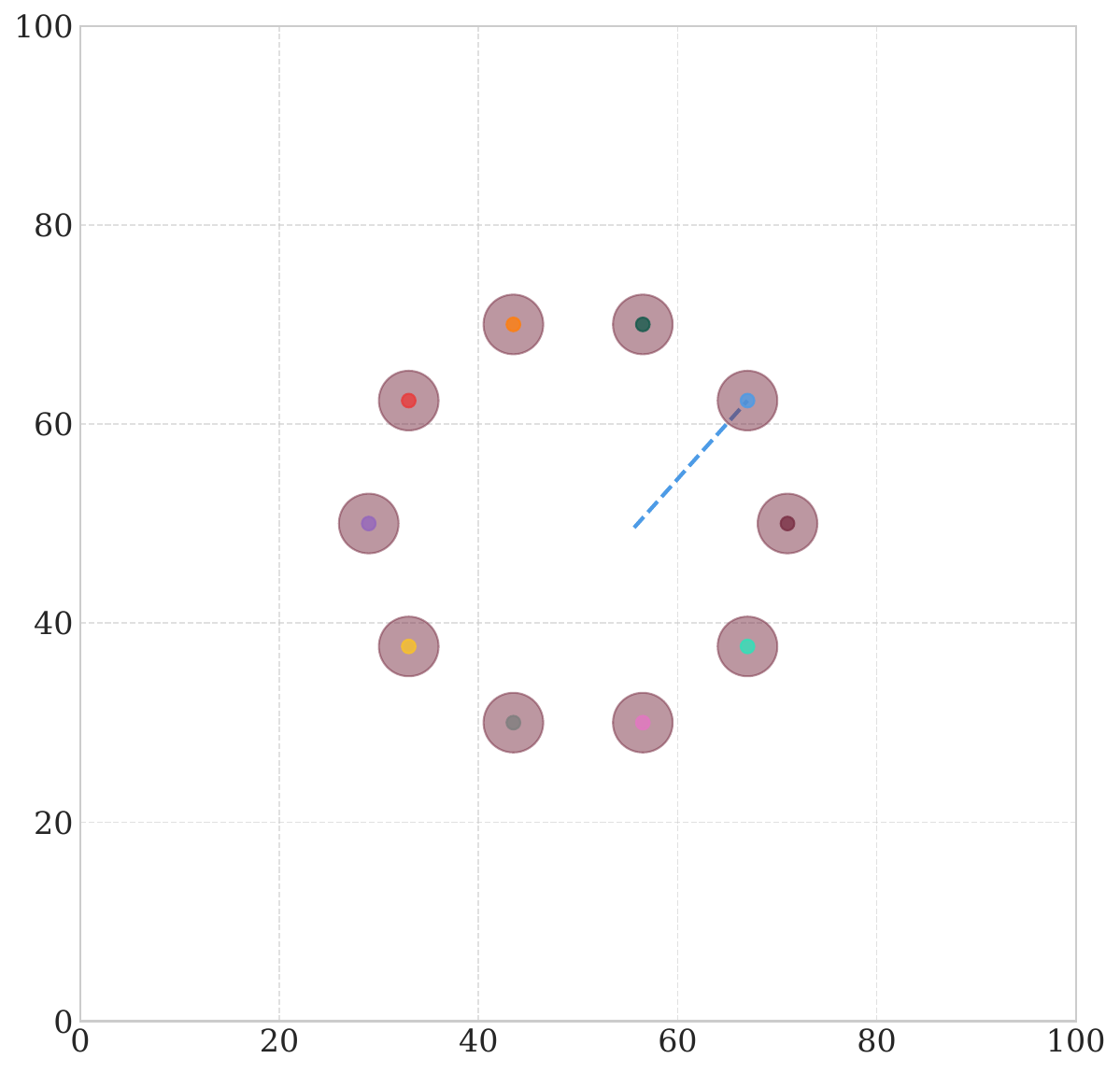}\label{fig-append:quali-circle-d}}
    
    \subfigure[Step 0]{\includegraphics[width=0.22\linewidth]{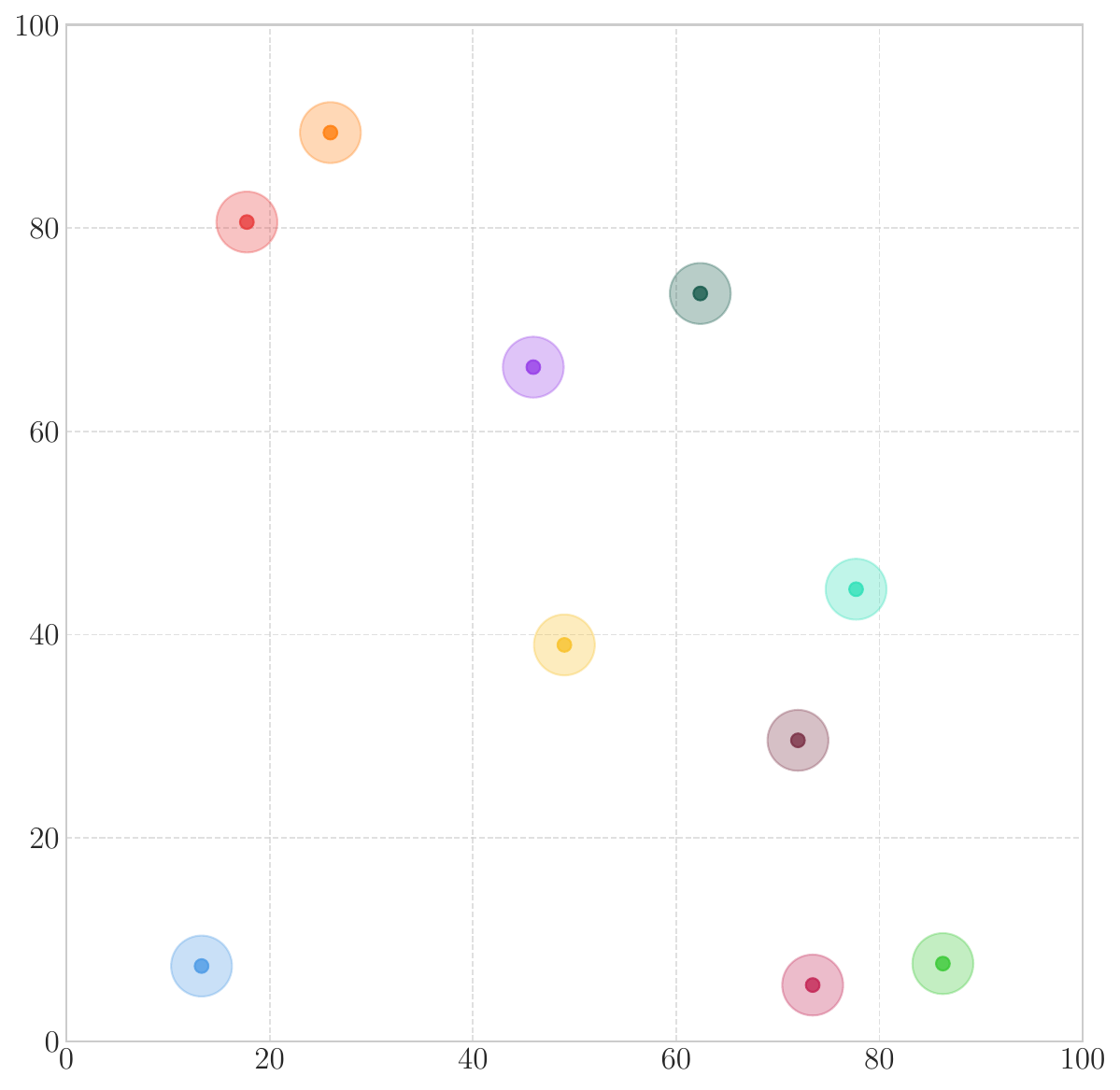}\label{fig-append:quali-circle-e}}
    \subfigure[Step 4]{\includegraphics[width=0.22\linewidth]{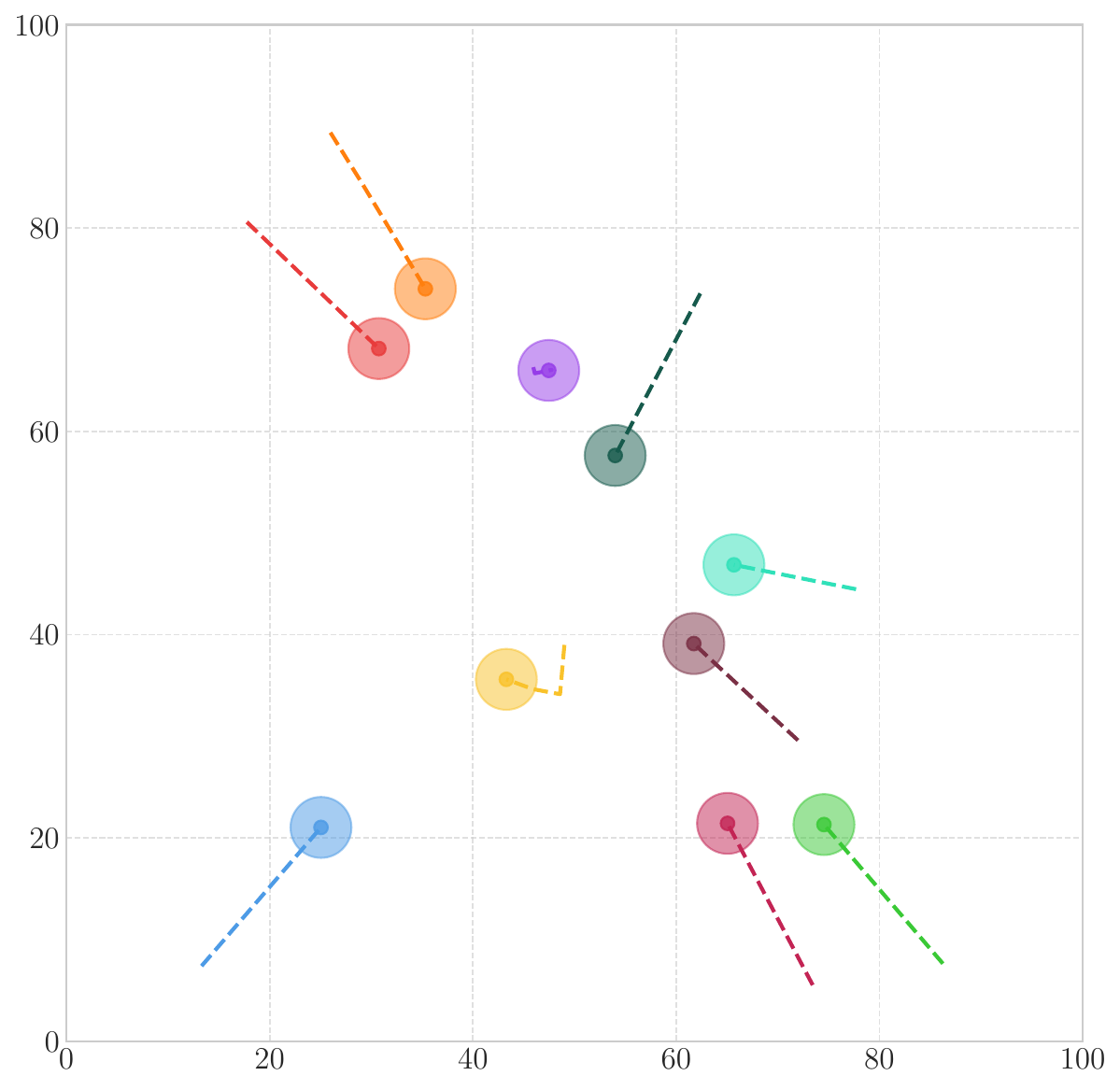}\label{fig-append:quali-circle-f}}
    \subfigure[Step 7]{\includegraphics[width=0.22\linewidth]{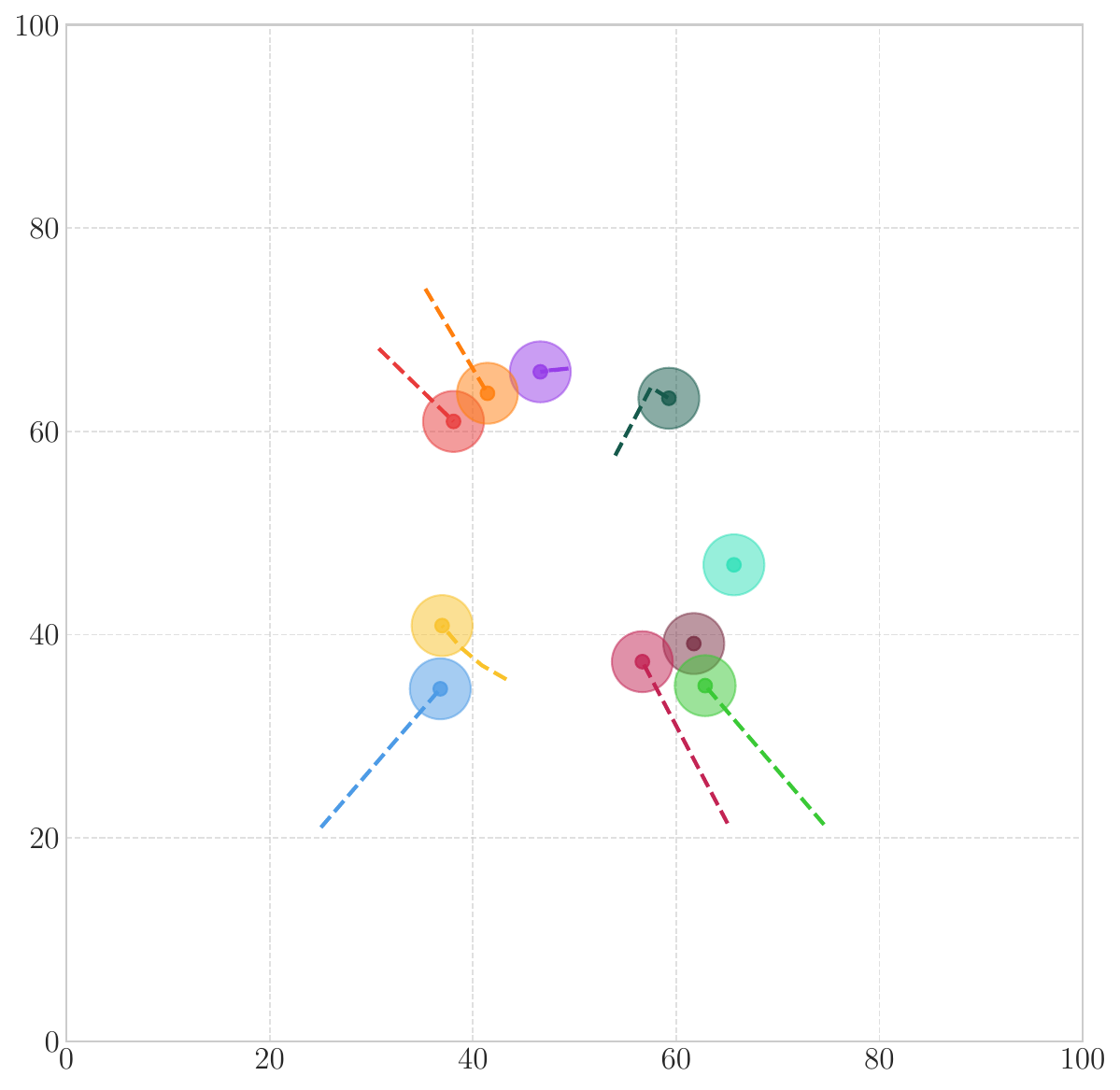}\label{fig-append:quali-circle-g}}
    \subfigure[Step 10]{\includegraphics[width=0.22\linewidth]{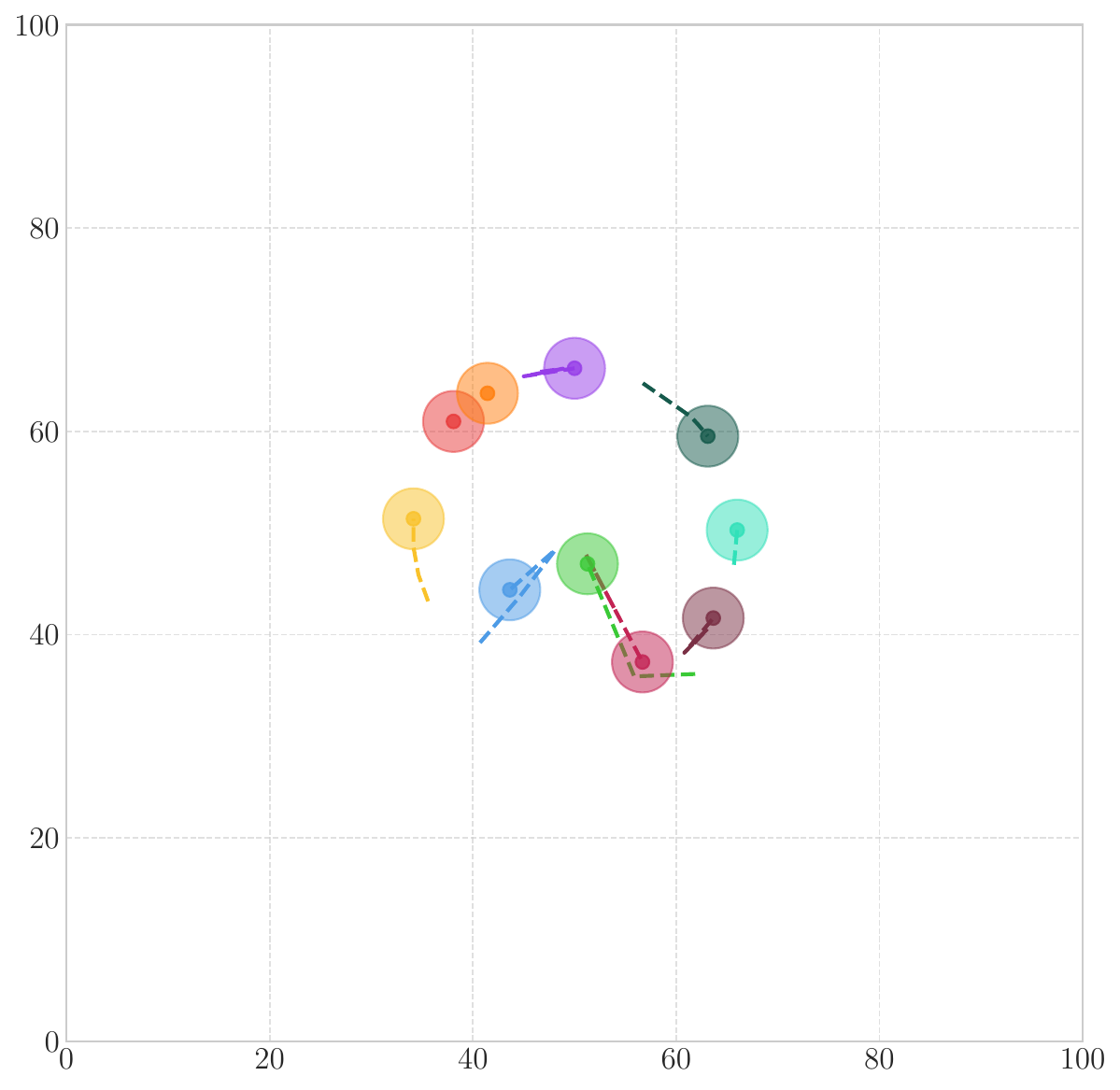}\label{fig-append:quali-circle-h}}
    \caption{Ten robots forming a circle. (a)–(d): Robots converge to evenly spaced circle using LLM-Flock. (e)–(h): Without consensus, inconsistent goal assignments lead to disorganized structures.}

    \label{fig-append:quali-circle}
    \vspace{-0.3cm}
\end{figure}
\begin{figure*}[!htbp]
    \centering
    \subfigure[Equilateral triangle with 3 robots]{\includegraphics[width=0.3\linewidth]{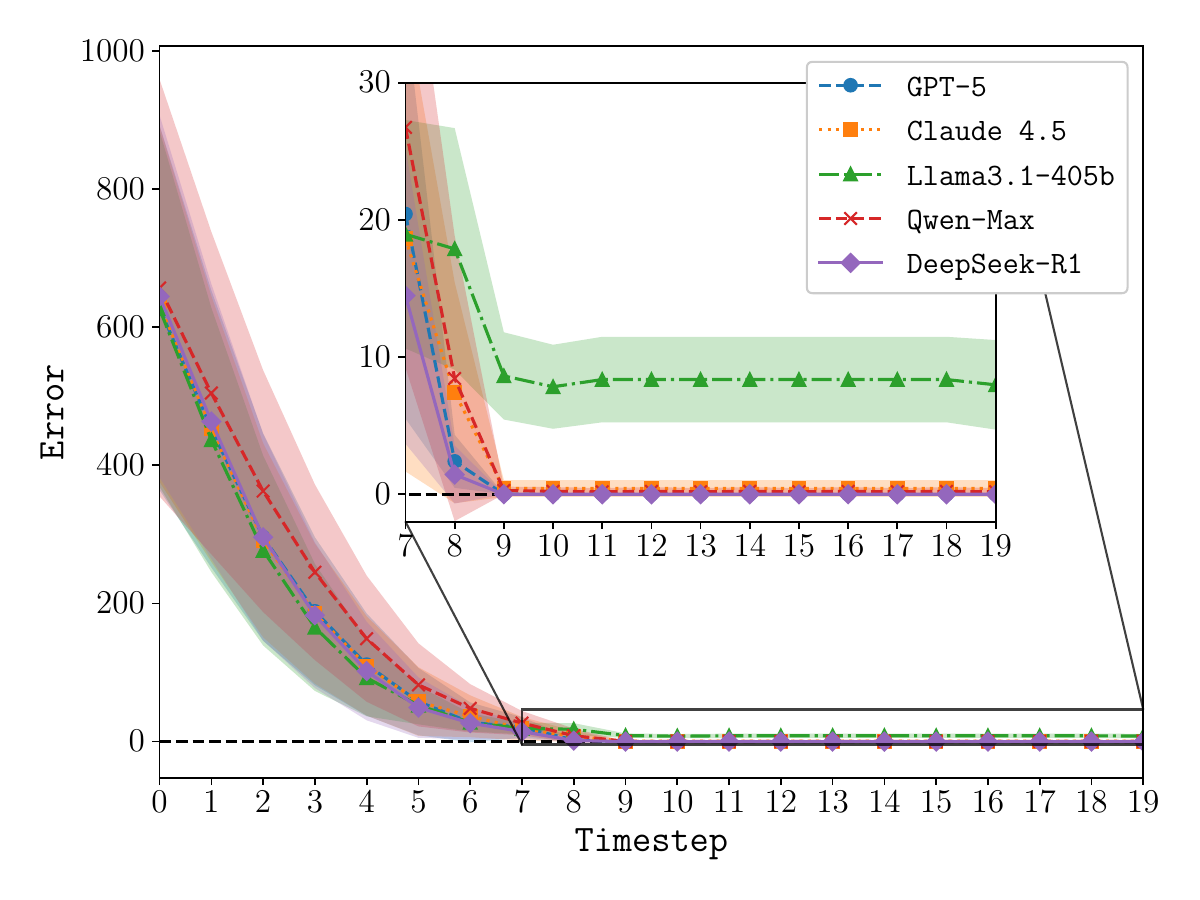}
    \label{fig-append:quant-plot-a}}
    \subfigure[Square with 8 robots]{\includegraphics[width=0.3\linewidth]{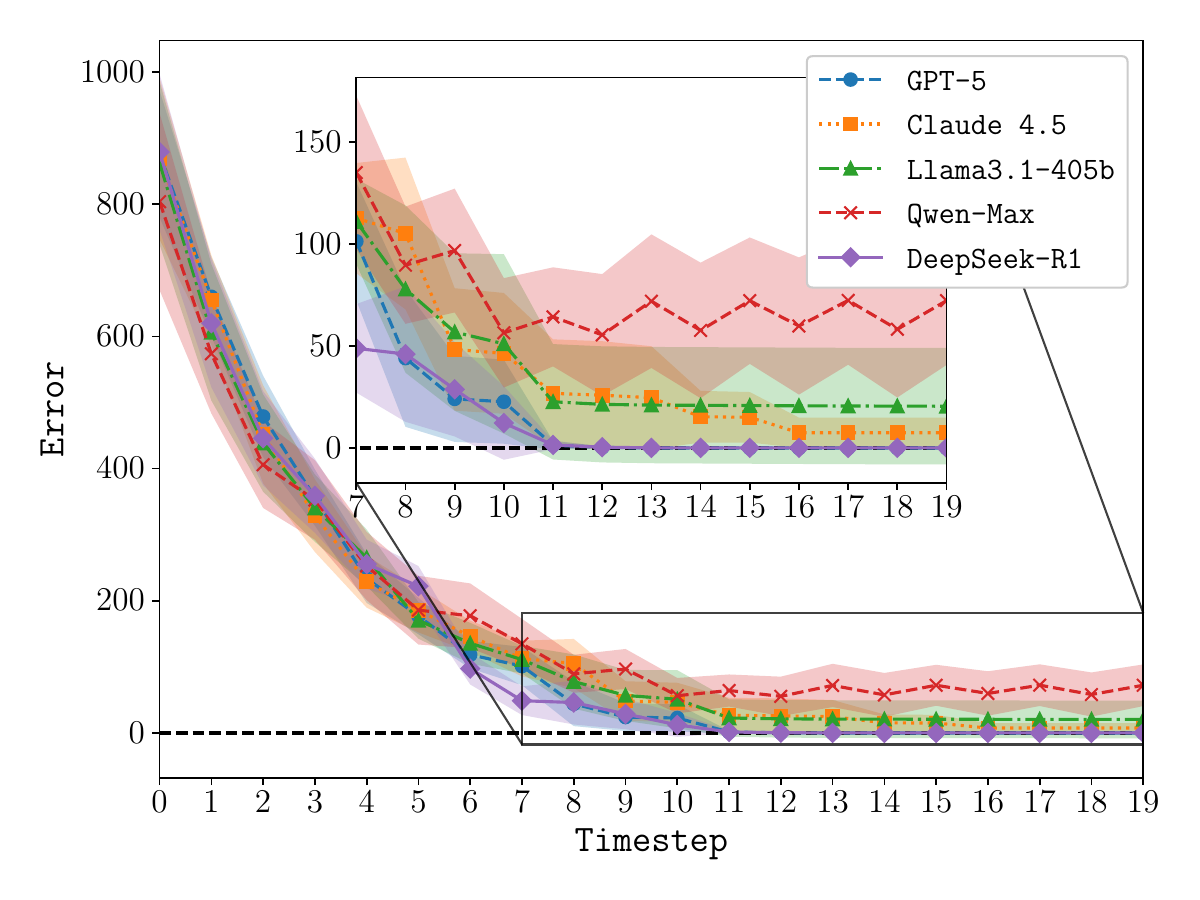}
    \label{fig-append:quant-plot-c}}
    \subfigure[Circle with 10 robots]{\includegraphics[width=0.3\linewidth]{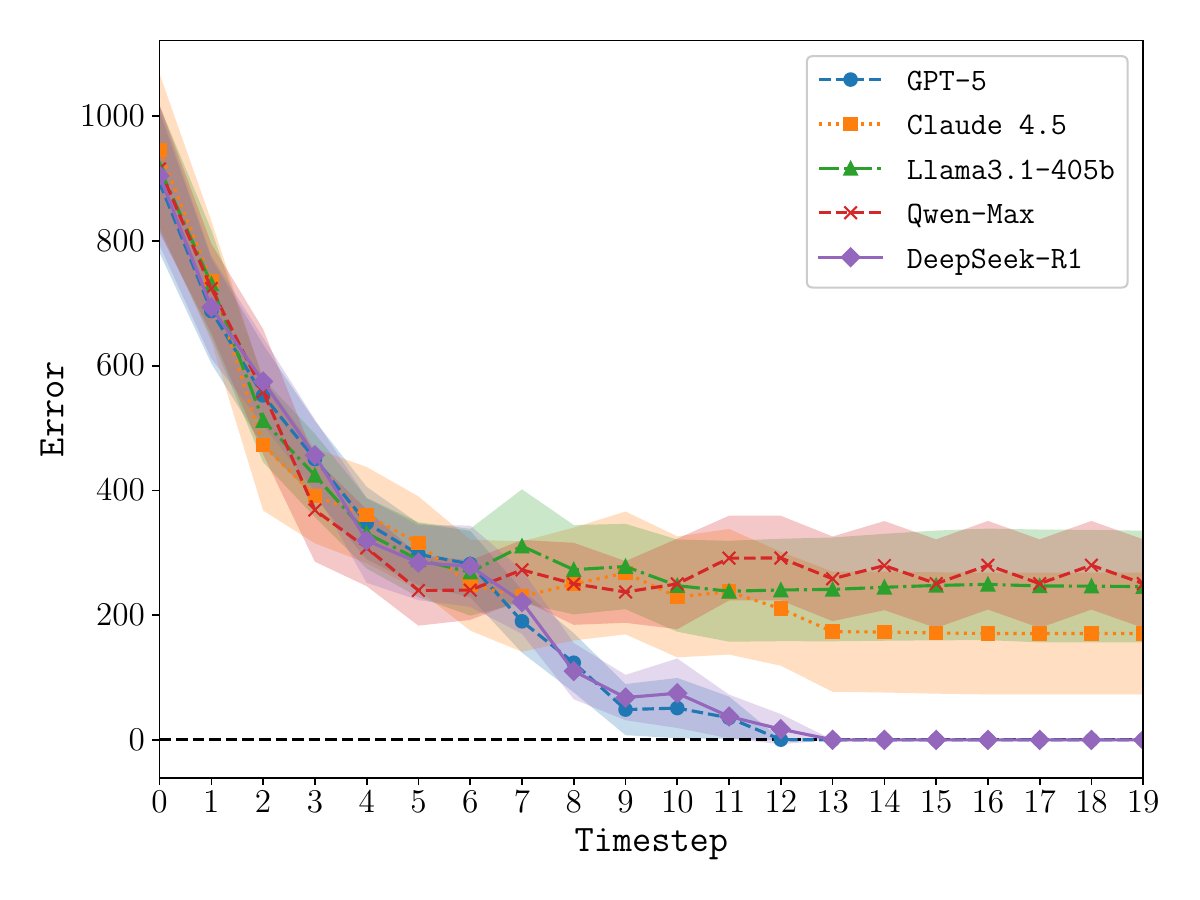}
    \label{fig-append:quant-plot-e}}
    \subfigure[Baseline comparison forming triangle with 3 robots]{\includegraphics[width=0.3\linewidth]{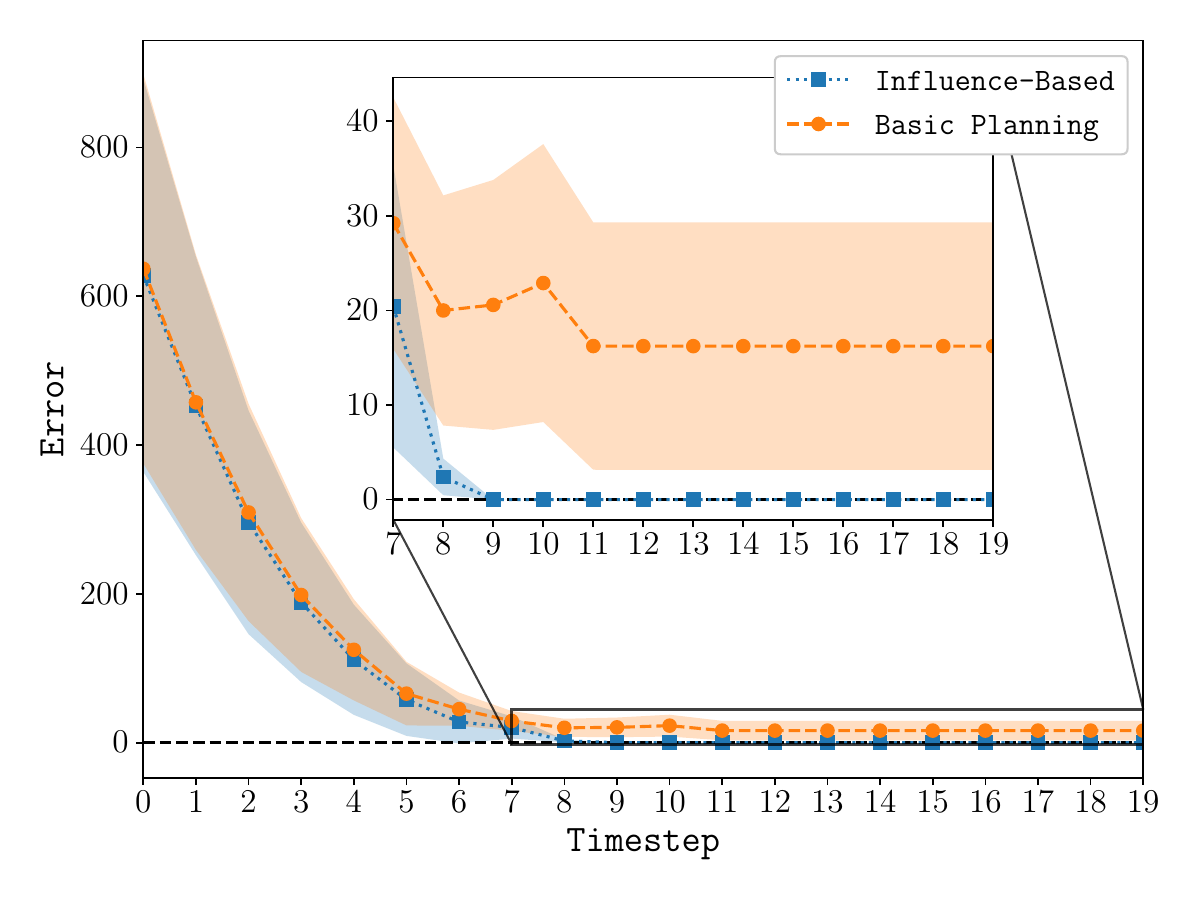}
    \label{fig-append:quant-plot-b}}
    \subfigure[Baseline comparison forming square with 8 robots]{\includegraphics[width=0.3\linewidth]{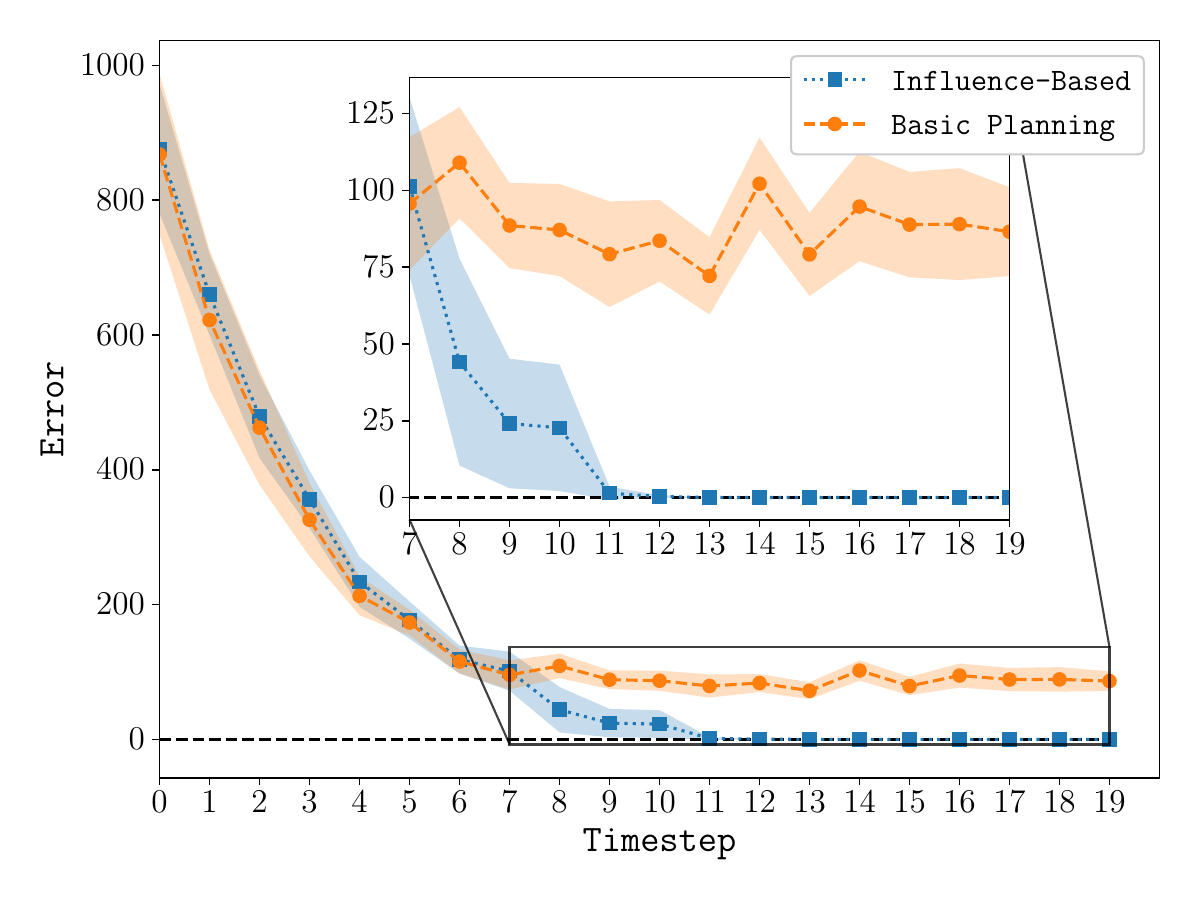}
    \label{fig-append:quant-plot-d}}
    \subfigure[Baseline comparison forming circle with 10 robots]{\includegraphics[width=0.3\linewidth]{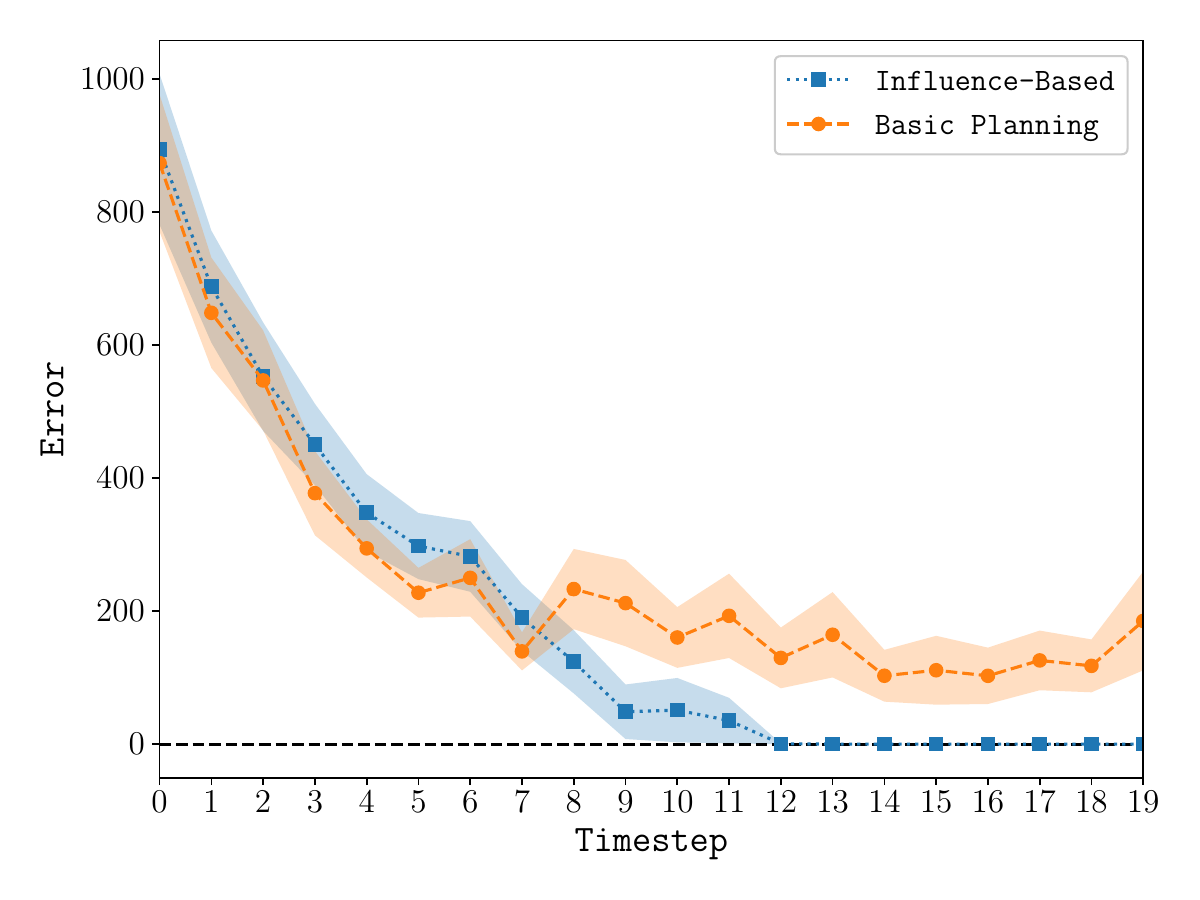}
    \label{fig-append:quant-plot-f}}

    \caption{Procrustes shape error over time across different formation tasks and LLM backends. Figures (a)-(c) show the average Procrustes error across 10 randomized trials using different LLMs under LLM-Flock. Figures (d)-(f) compare performance with and without consensus using GPT-5. Shaded regions represent 95\% confidence intervals. Results demonstrate that influence-based consensus significantly improves formation stability and consistency. Notably, LLM-Flock with GPT-5 can consistently produce a procrustes error of 0, which is on par with the traditional method, which asymptotically converges~\cite {olfati2006flocking}.}
    \label{fig-append:quant-plot}
    \vspace{-0.3cm}
\end{figure*}

Our evaluation covers diverse formation tasks, multiple LLM backends, and both simulated environments and real robot experiments. We first present qualitative results illustrating the emergence of flocking behaviors, including a comparison to a baseline without influence-based consensus. We then quantitatively analyze convergence performance across different LLMs and formation shapes. Finally, we validate LLM-Flock’s applicability through real-world experiments with a team of Crazyflie drones~\cite{crazyflie}.

All experiments, unless otherwise specified, use a desired inter-robot distance of 10 units, a maximum speed of 6 units per round, a minimum safe distance of 3 units, and a communication range of 15 units. All the robots are initialized randomly in a 2D environment with 100 by 100 units. Formations are centered at $[50,50]$, and robots are assumed to have perfect localization in simulation. 

\subsection{Qualitative Simulation Evaluation}

We qualitatively evaluate LLM-Flock by comparing its influence-based consensus approach against a baseline method \PL{in simulation}. Using GPT-5~\cite{singh2025openai} as the LLM backend, we test both approaches on triangle formations with three robots. 
In the baseline, each robot follows its initial plan independently, selecting its assigned goal without negotiating with neighbors. This setup highlights an approach that relies solely on unstructured LLM planning for decentralized formation control.

\PL{As shown in Fig.~\ref{fig:quali-baseline}, LLM-Flock consistently converges to symmetric triangle formations. Disk colors indicate each robot's adopted plan, and by the final frame (Fig.~\ref{fig:quali-tri3-d}), all robots share the same plan through influence-based adoption. In contrast, the baseline produces a collapsed or disorganized configuration as in Fig.~\ref{fig:quali-tri3-h}. We further test the square (8 robots, Fig.~\ref{fig-append:quali-square}) and circle (10 robots, Fig.~\ref{fig-append:quali-circle}) formations. In both cases, LLM-Flock yields evenly spaced formations that match the target geometry, whereas the baseline fails to form stable shapes.}
These results highlight LLM-Flock’s ability to coordinate larger teams and support diverse formation geometries through decentralized influence-based consensus. 
\subsection{Quantitative Simulation Evaluation}



We quantitatively evaluate the convergence performance of LLM-Flock across different language model backends and formation tasks \PL{in simulation}. As a convergence metric, we use the Procrustes shape error~\cite{gower2004procrustes, kendall1989survey}, which measures the geometric difference between the actual robot formation $X$ and the target formation $Y$ after optimal alignment through translation and rotation. Formally, the Procrustes error at timestep $t$ is computed as:
\begin{equation}
\text{Error}(t) = \frac{1}{N} \sum_{i=1}^N \left\| \mathbf{x}_i(t) - \mathbf{y}_i \right\|_2,
\end{equation}
where $\mathbf{x}_i(t)$ and $\mathbf{y}_i$ denote the positions of robot $i$ and actual target formation, respectively, and $N$ is the number of robots. 

For each configuration, we run 10 randomized trials and plot the mean Procrustes error over time, with shaded regions indicating the 95\% confidence interval. We use five popular LLMs: OpenAI GPT-5, Claude 4.5 Sonnet~\cite{anthropic2025claude45sonnet}, DeepSeek-R1~\cite{guo2025deepseek}, Llama3.1-405b~\cite{grattafiori2024llama}, and Qwen-Max~\cite{yang2024qwen2}. We evaluate each model on triangle formations with three and six robots, a square formation with eight robots, and a circle formation with ten robots. All experiments use the environmental parameters and formation center described earlier. For brevity, we only demonstrate results from square and circular formations.

Fig.~\ref{fig-append:quant-plot} shows that LLM-Flock enables stable convergence across all tested LLMs and team sizes. Models with stronger reasoning abilities, such as OpenAI GPT-5 and DeepSeek-R1, achieve faster convergence and consistently form the desired formations across all tasks. In contrast, general-purpose models such as Claude 4.5, Qwen-Max, and Llama 3.1 exhibit slower convergence and higher residual errors. 



\begin{figure}
    \centering
    \subfigure[Step 1]{\includegraphics[width=0.30\linewidth]{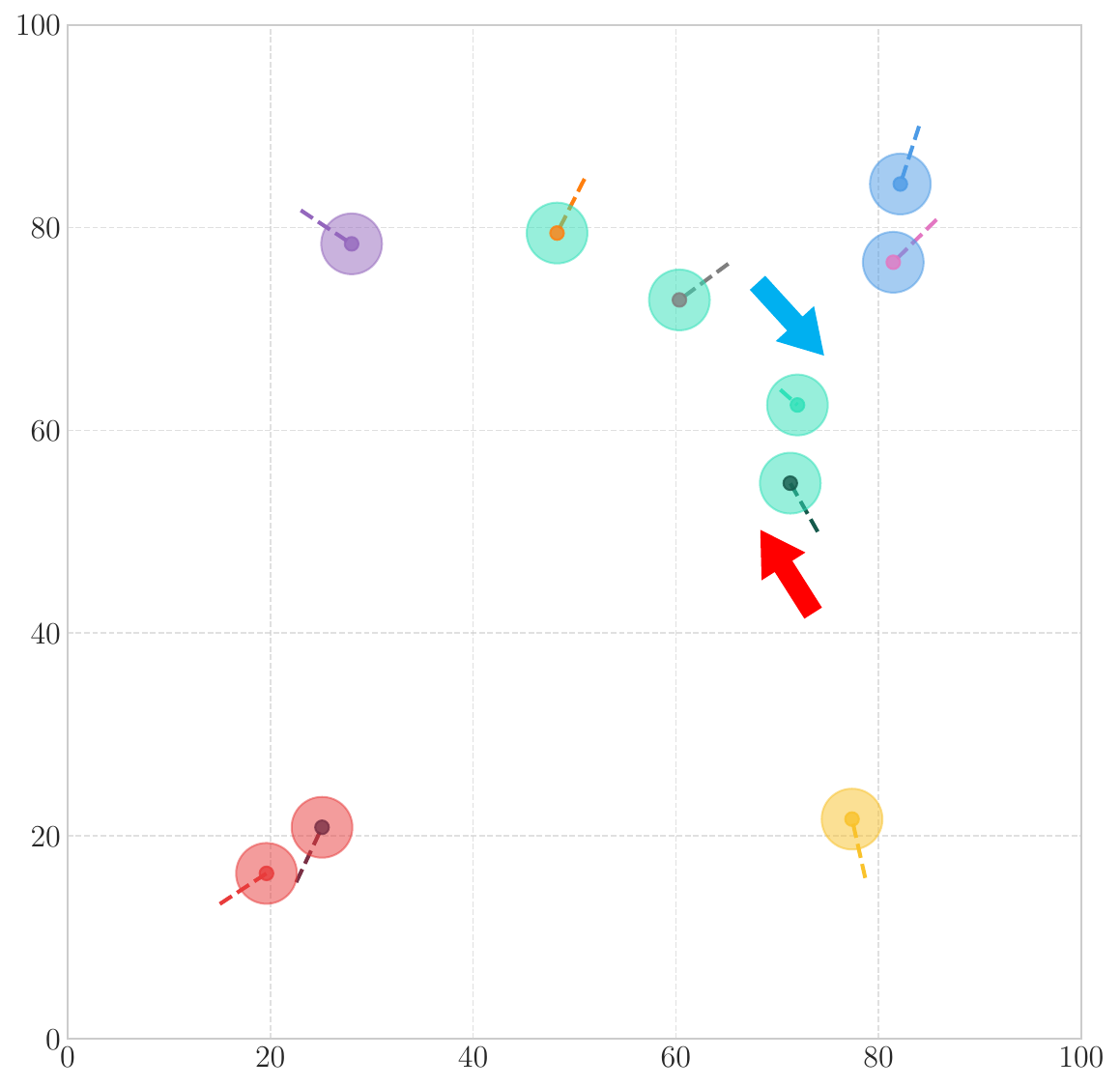}\label{fig:collide-1}}
    \subfigure[Step 2]{\includegraphics[width=0.30\linewidth]{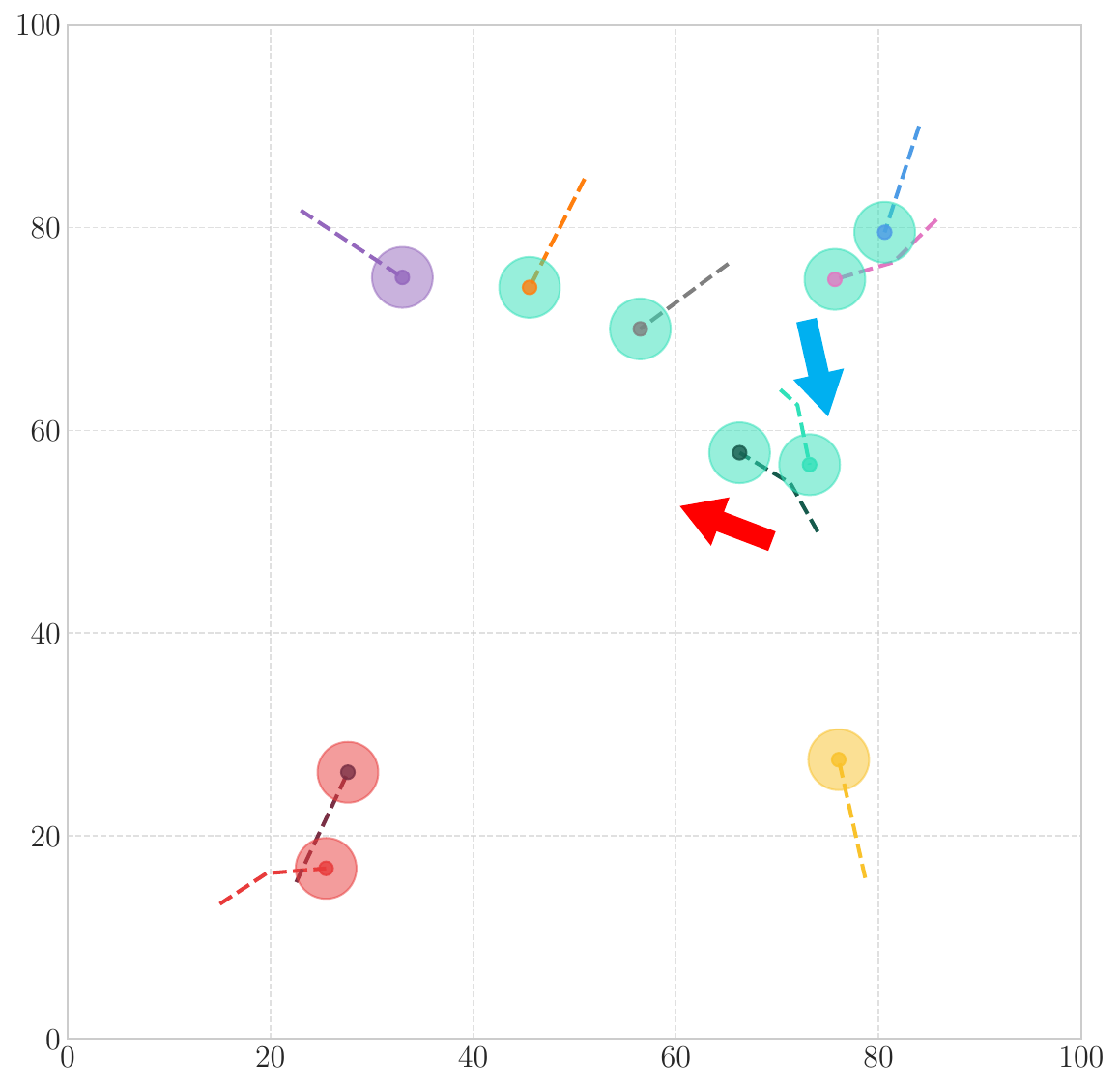}\label{fig:collide-2}}
    \subfigure[Step 3]{\includegraphics[width=0.30\linewidth]{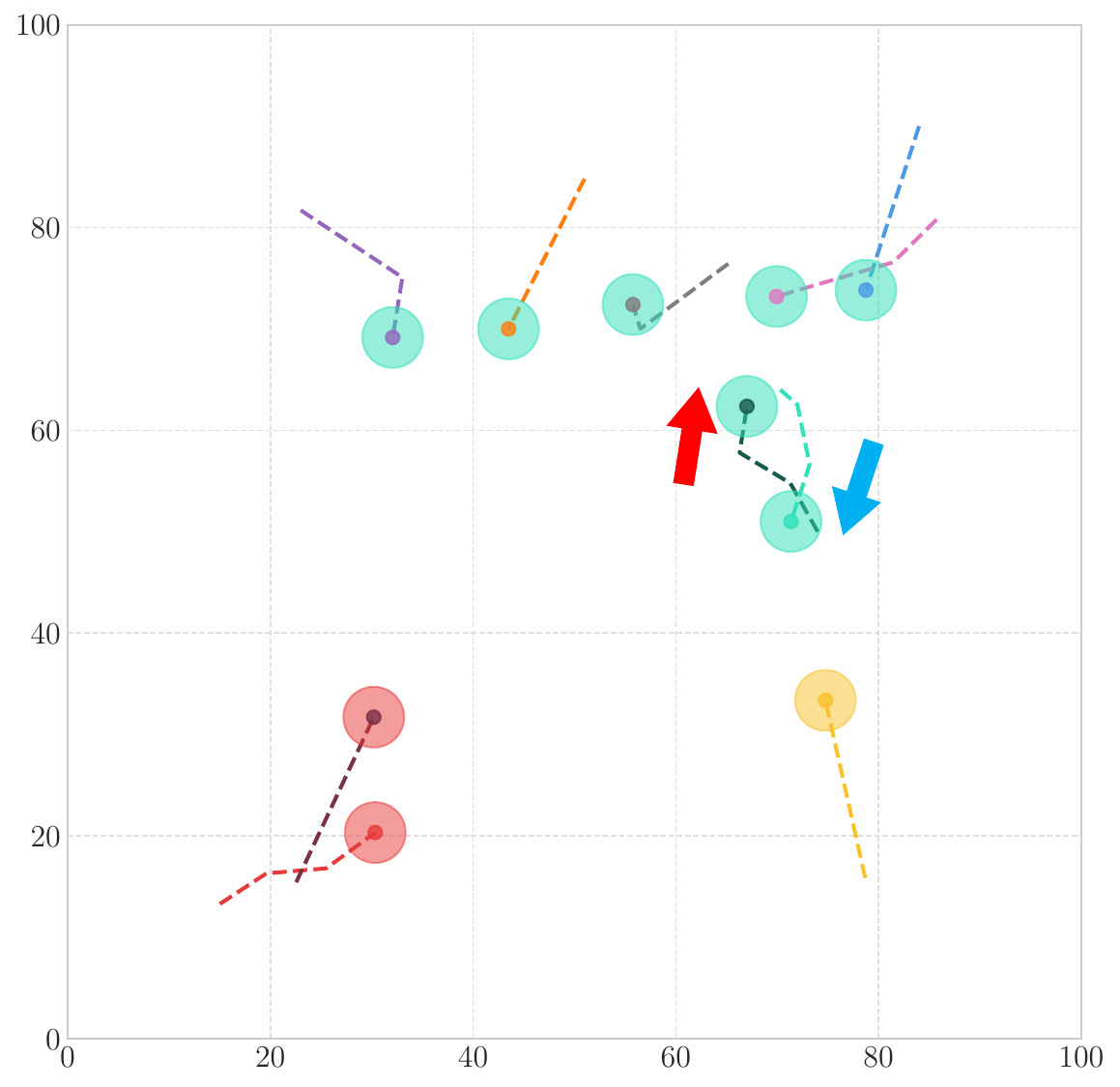}\label{fig:collide-3}}
    \caption{Collision avoidance demonstration with GPT-5 in a 10-robot circle formation. Two robots (cyan disks, upper right) approach head-on but, guided by the two-layer collision avoidance mechanism, choose different paths to avoid collision.}
    \label{fig:collide}
    \vspace{-0.3cm}
\end{figure}

\begin{figure}[ht!]
    \centering
    \subfigure[Step 0]{\includegraphics[width=0.22\linewidth]{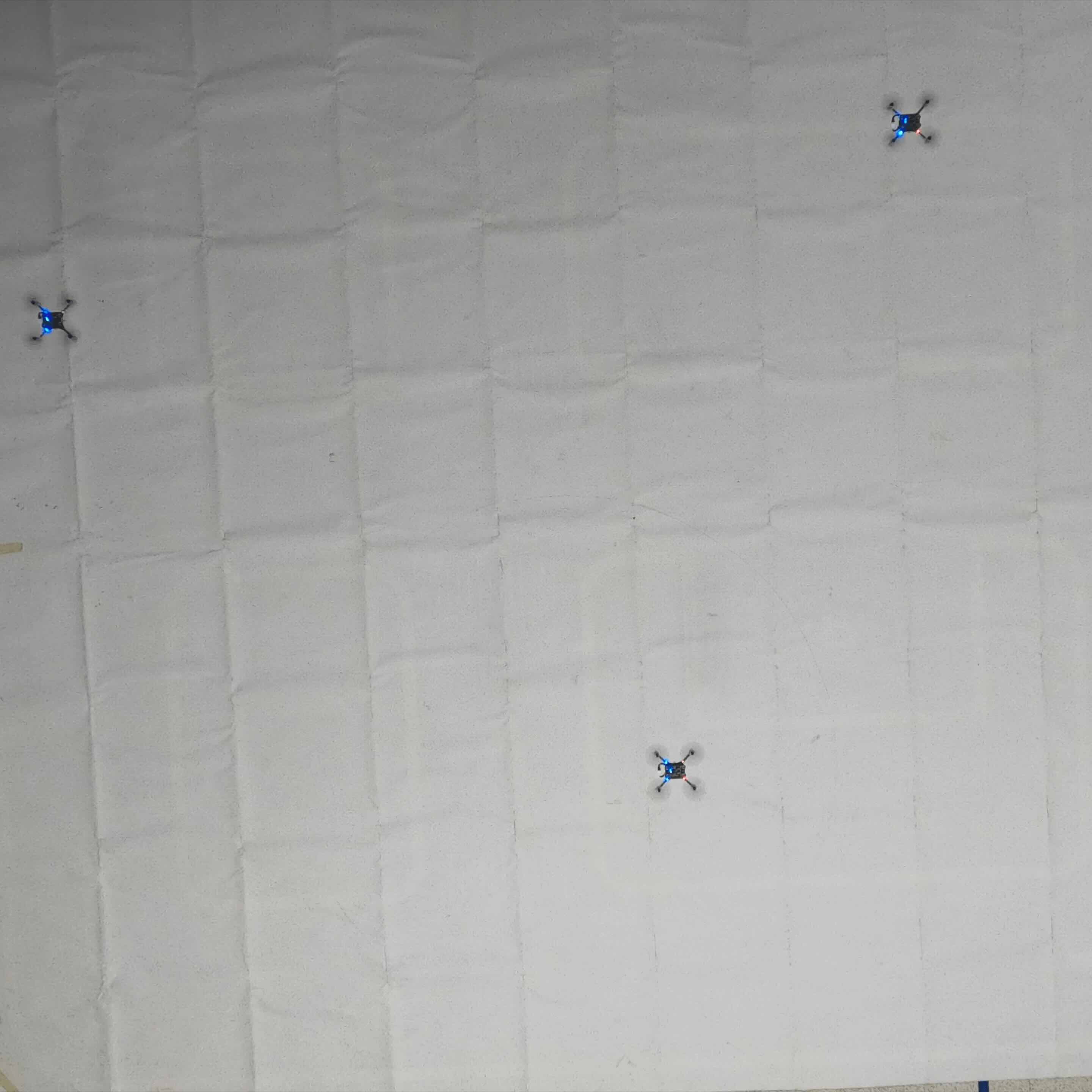}}
    \subfigure[Step 4]{\includegraphics[width=0.22\linewidth]{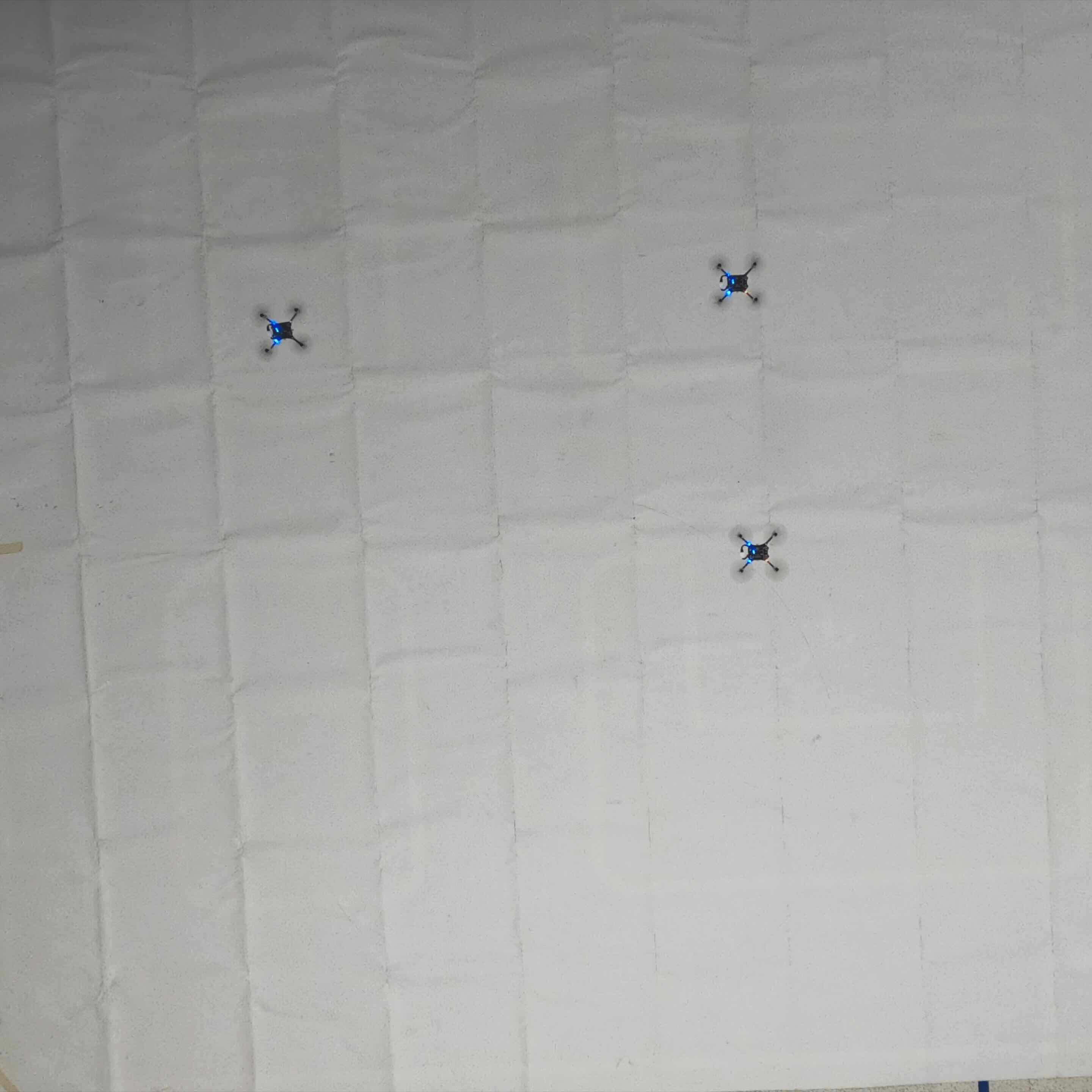}}
    \subfigure[Step 7]{\includegraphics[width=0.22\linewidth]{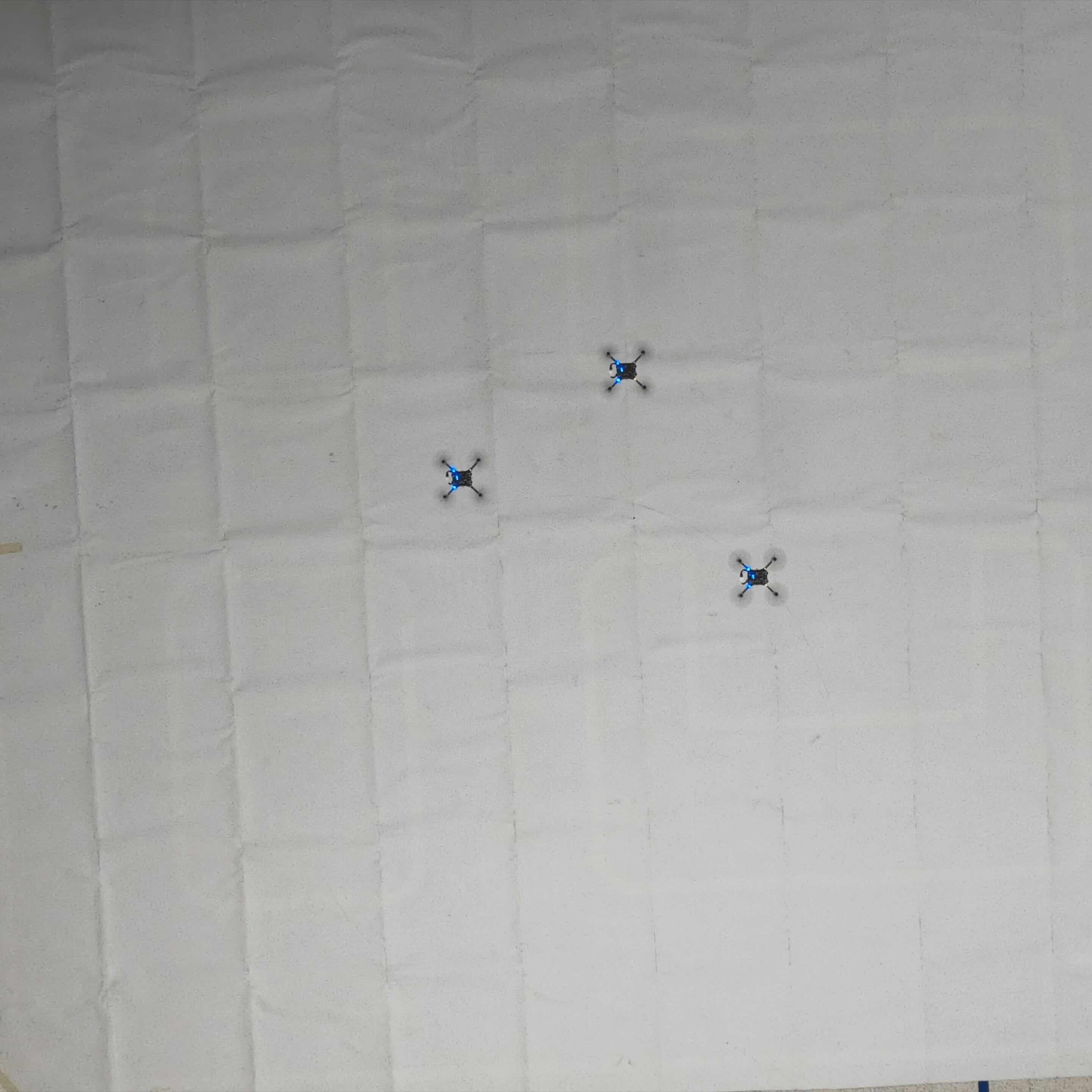}}
    \subfigure[Step 12]{\includegraphics[width=0.22\linewidth]{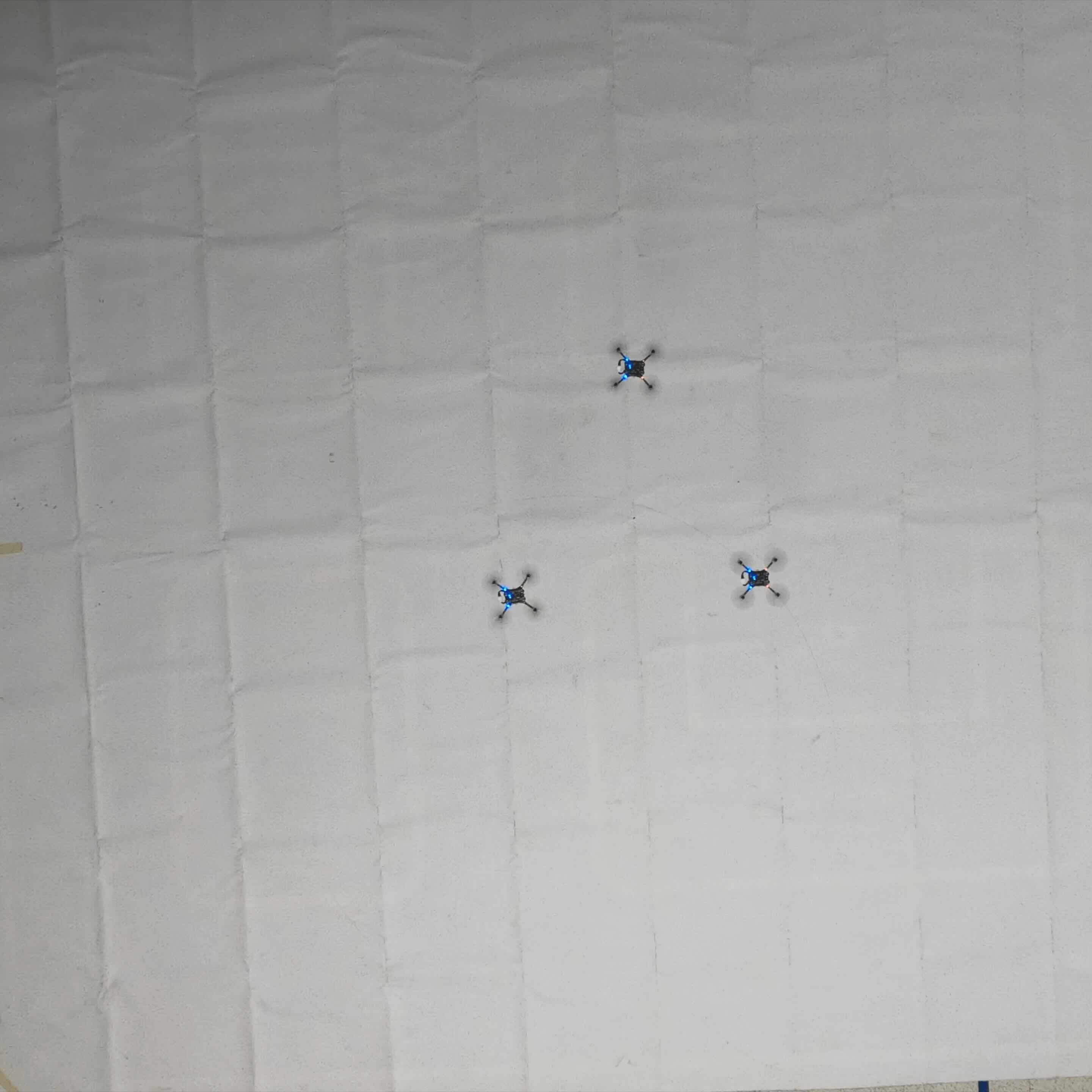}}

    \subfigure[Step 0]{\includegraphics[width=0.22\linewidth]{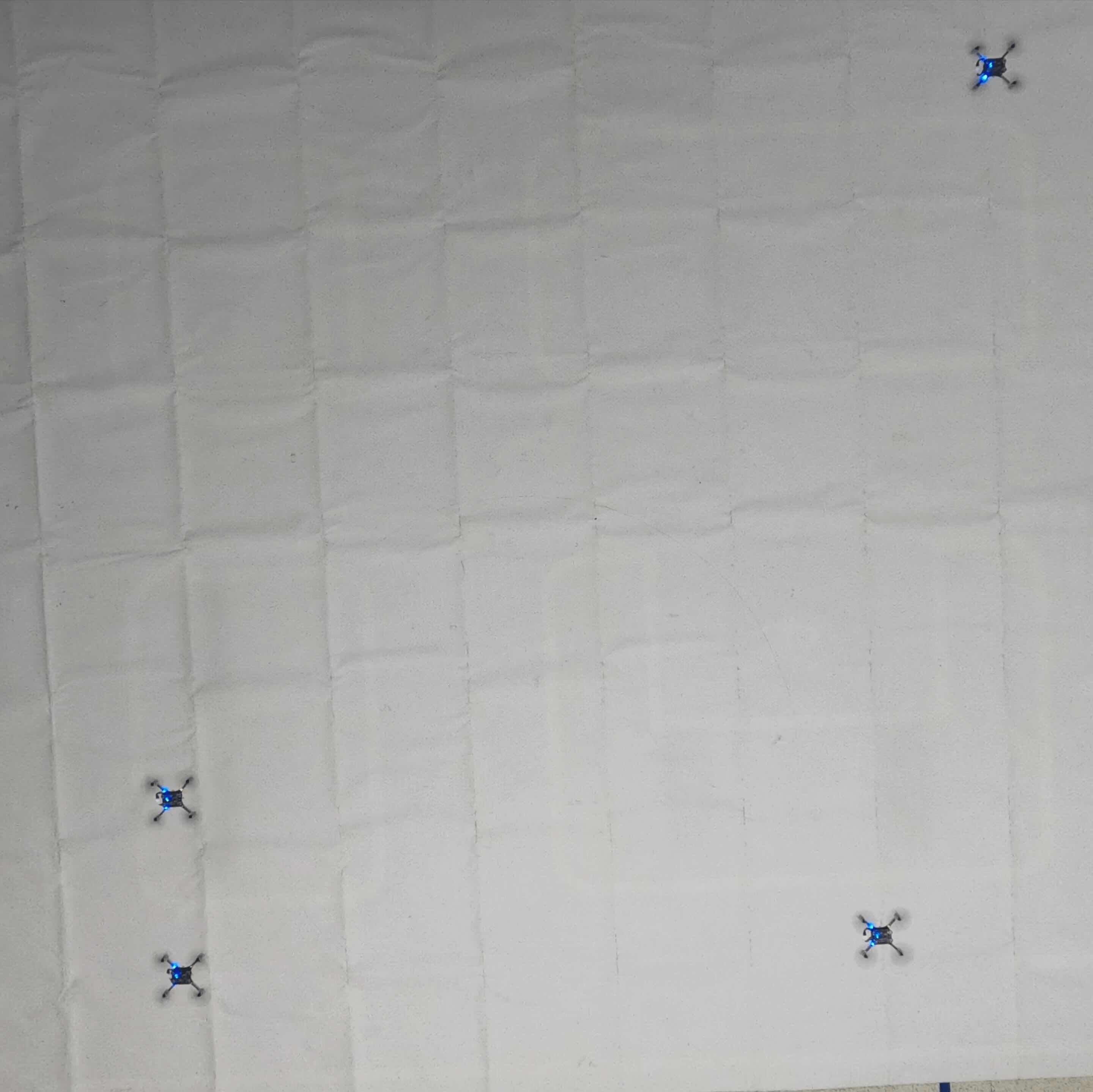}}
    \subfigure[Step 4]{\includegraphics[width=0.22\linewidth]{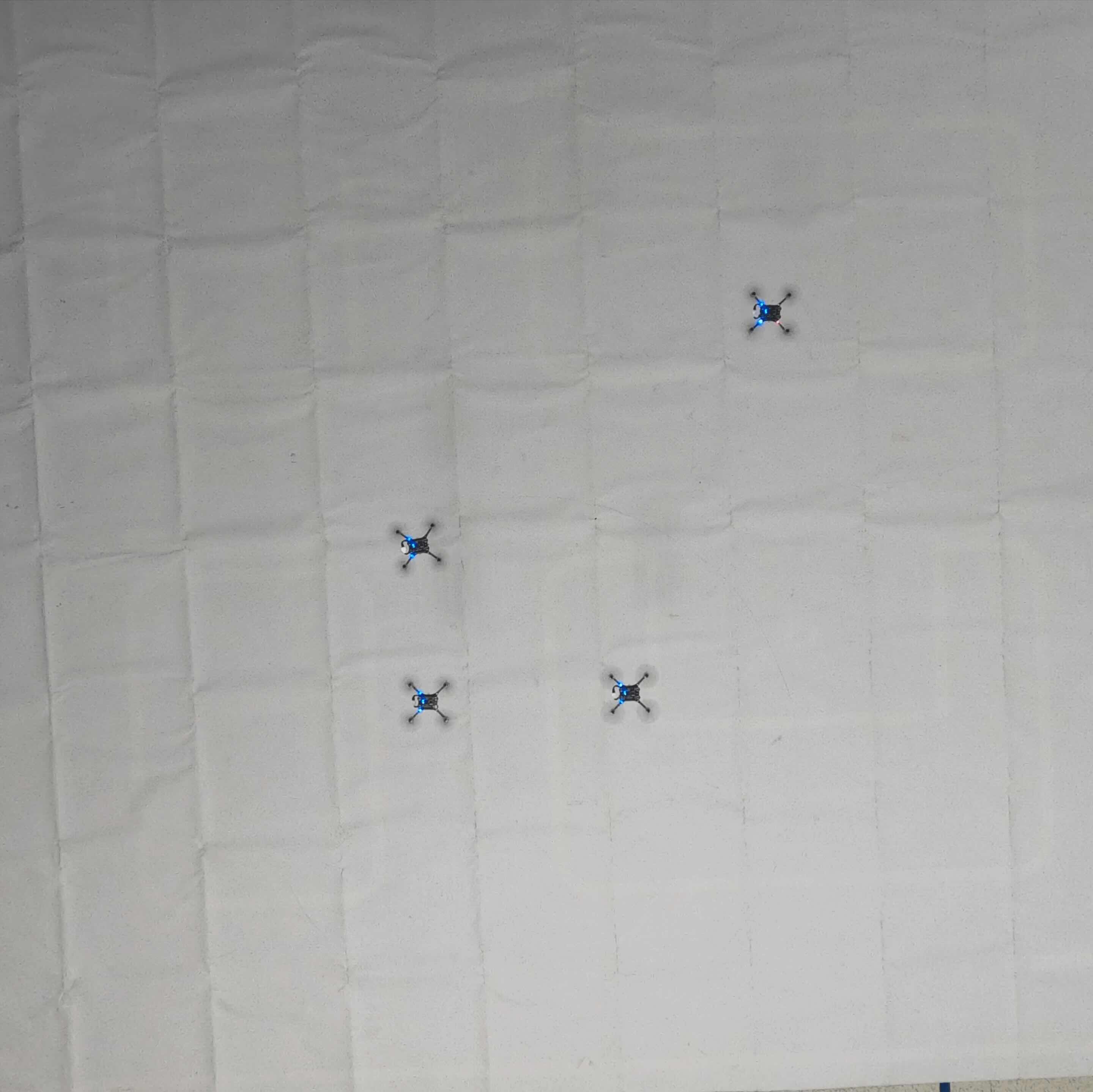}}
    \subfigure[Step 7]{\includegraphics[width=0.22\linewidth]{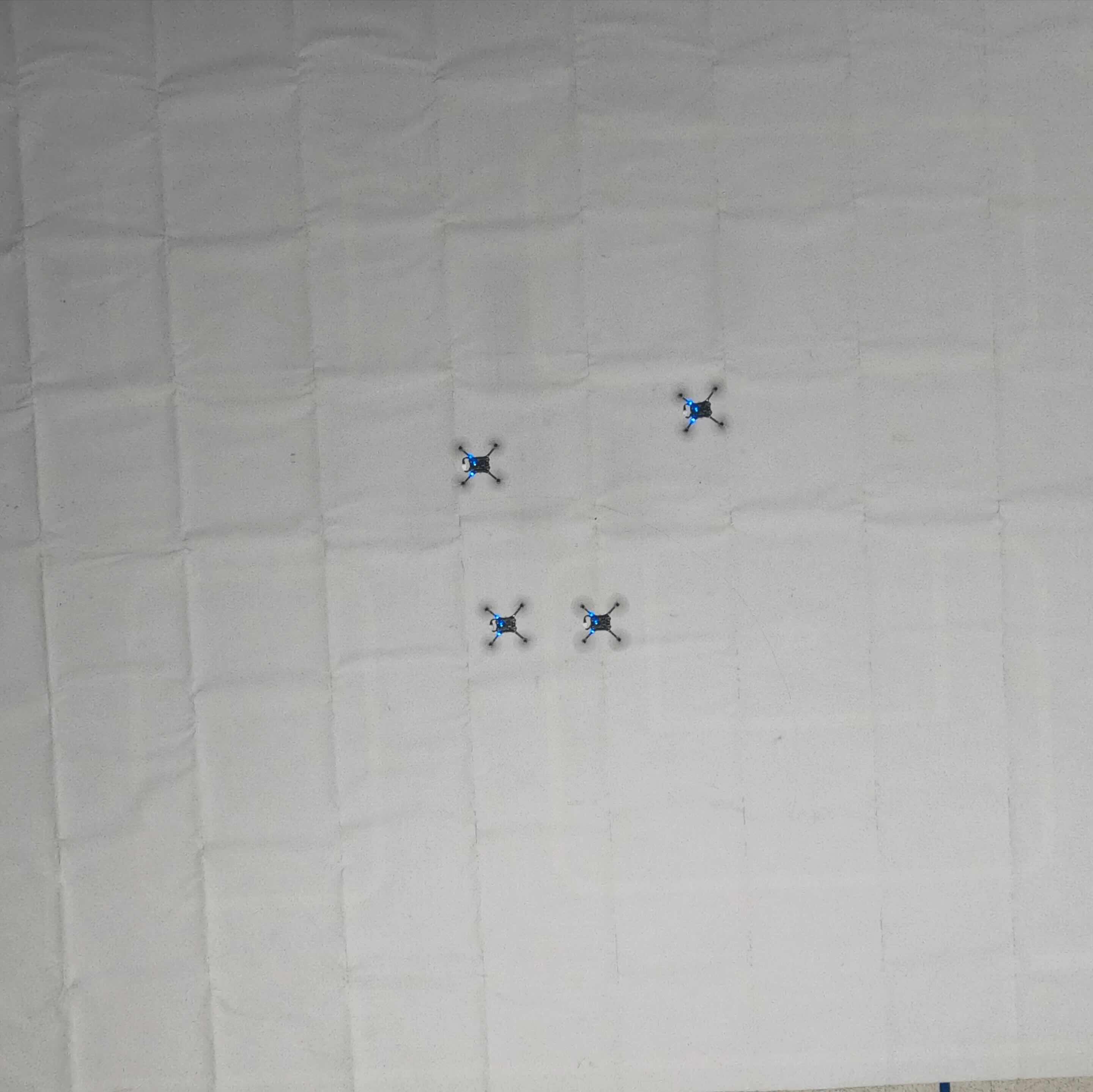}}
    \subfigure[Step 12]{\includegraphics[width=0.22\linewidth]{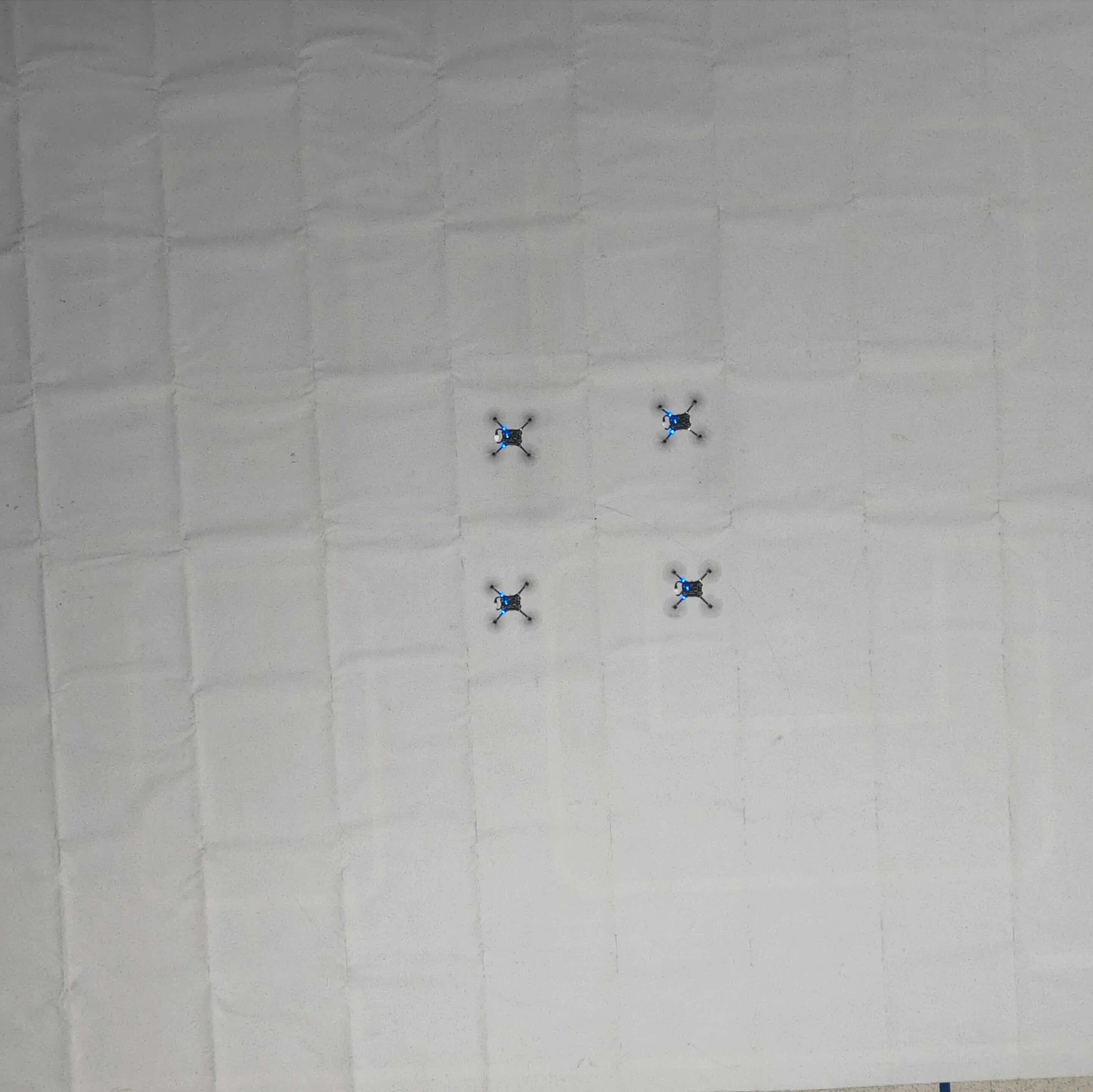}}

    \subfigure[Step 0]{\includegraphics[width=0.22\linewidth]{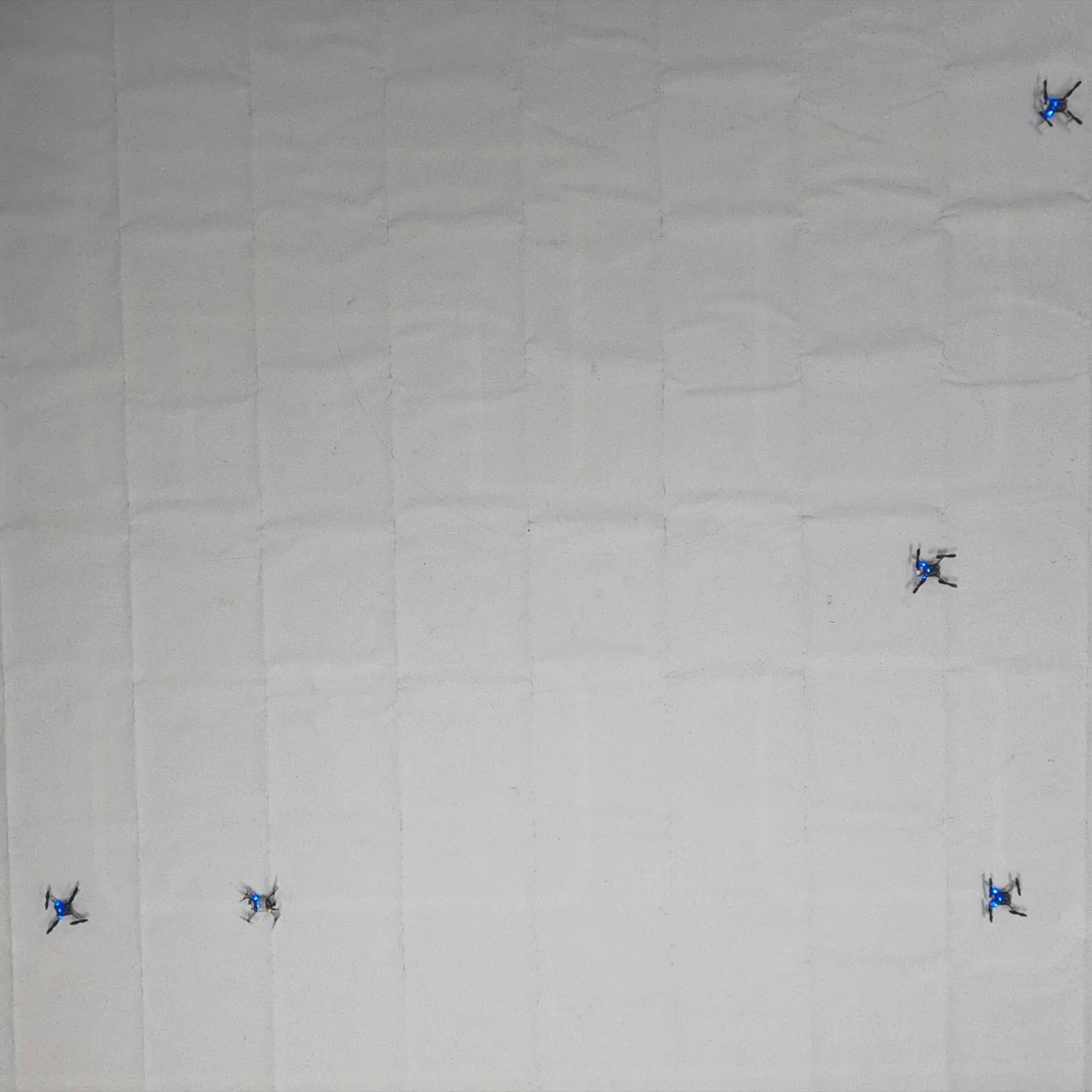}}
    \subfigure[Step 4]{\includegraphics[width=0.22\linewidth]{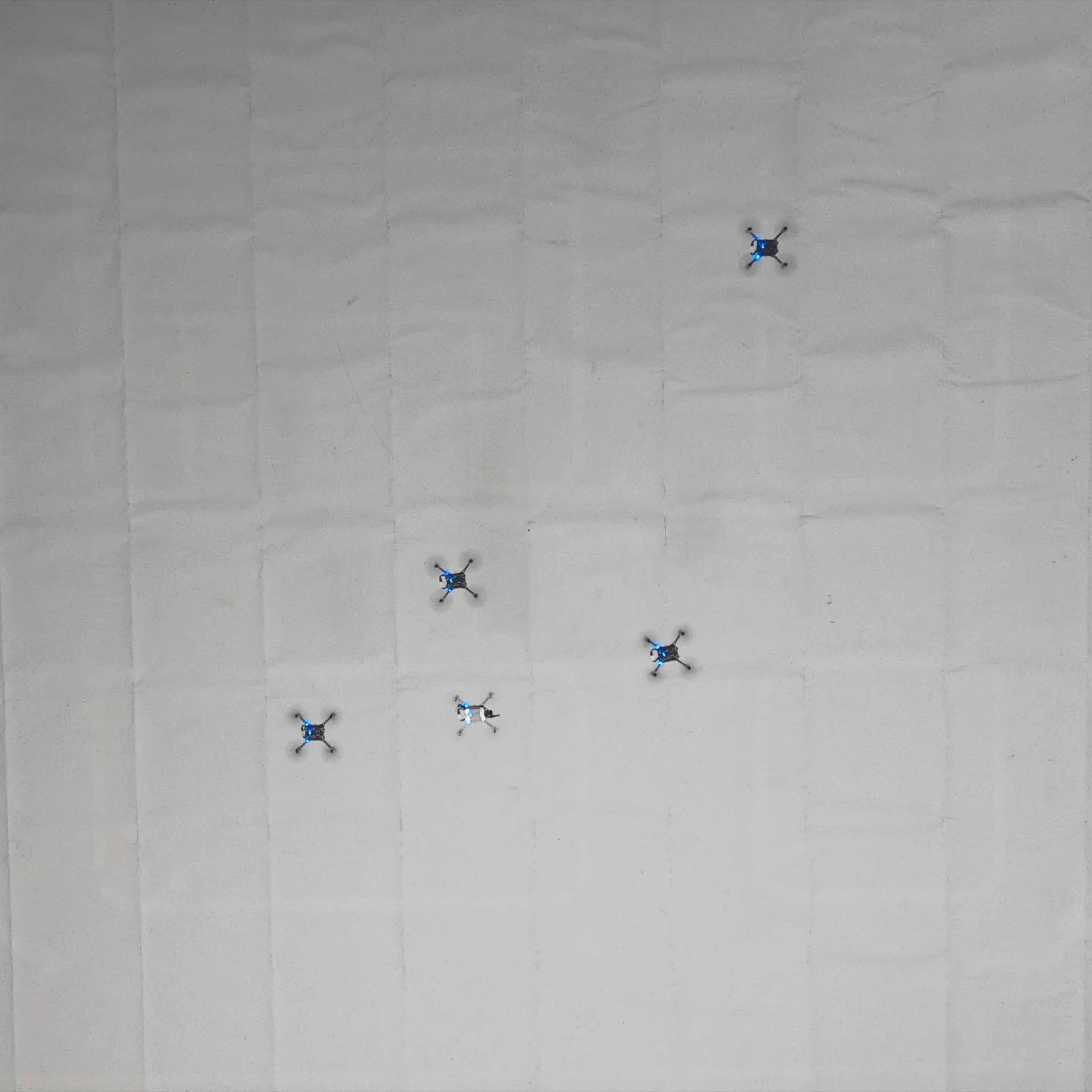}}
    \subfigure[Step 7]{\includegraphics[width=0.22\linewidth]{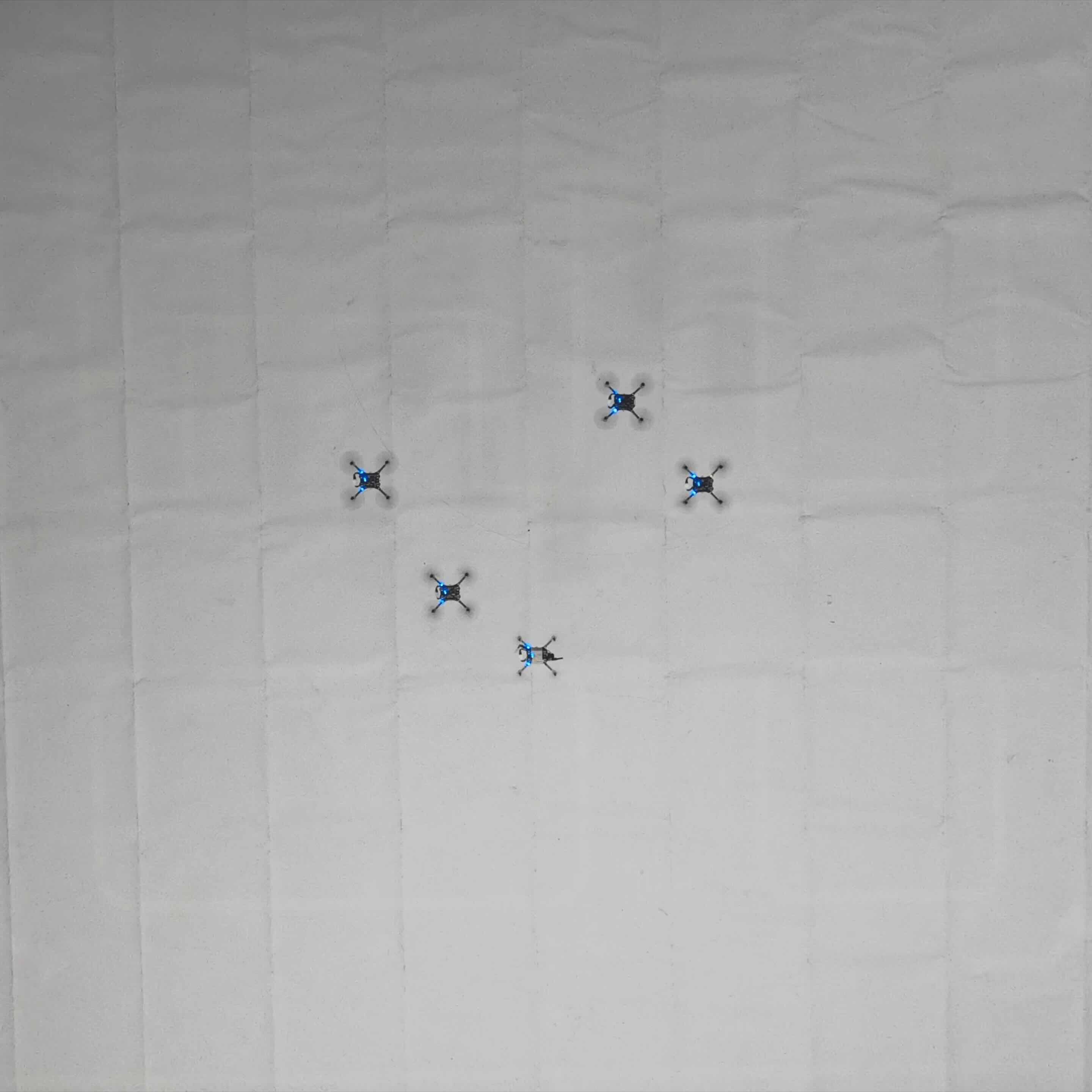}}
    \subfigure[Step 12]{\includegraphics[width=0.22\linewidth]{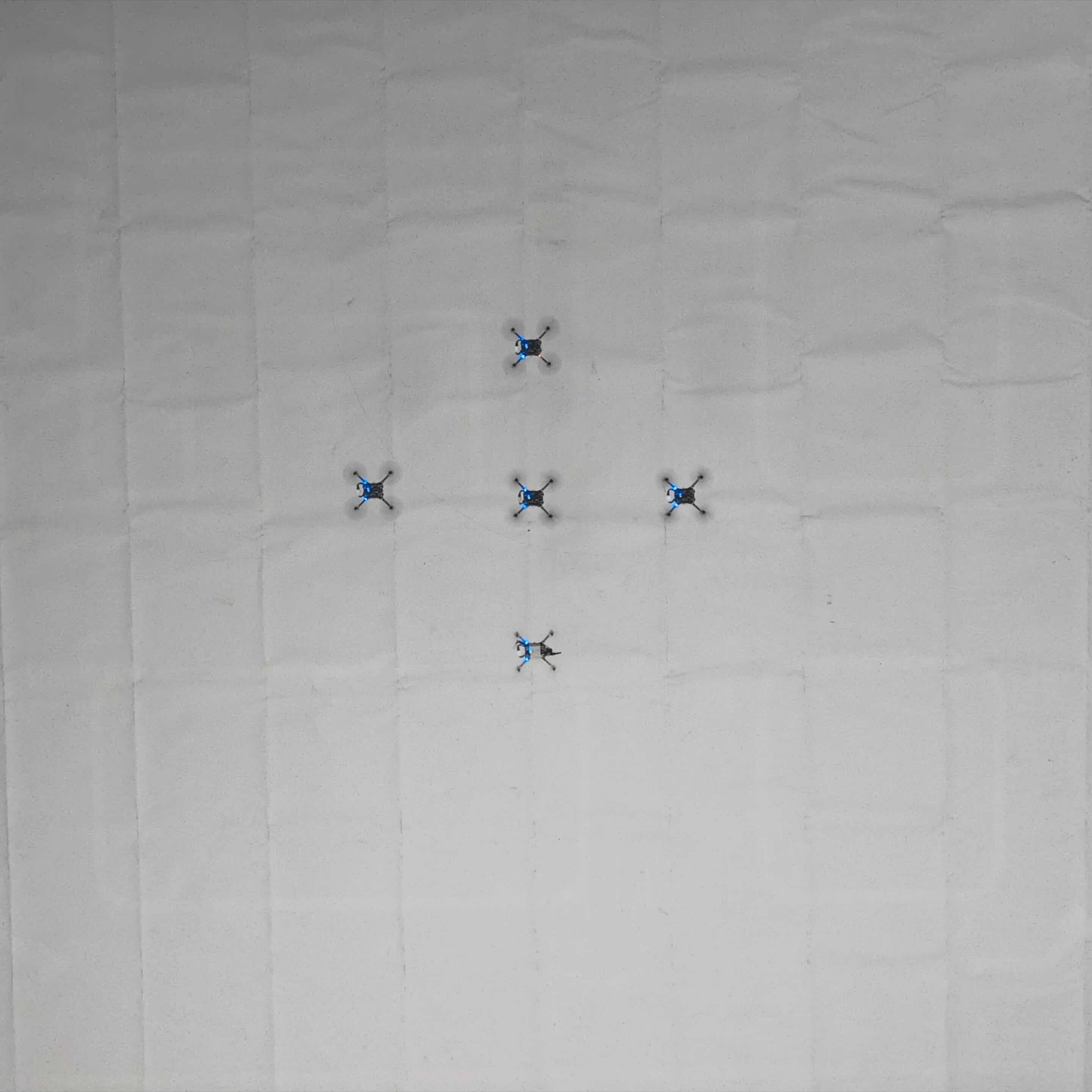}}

    \subfigure[Step 0]{\includegraphics[width=0.22\linewidth]{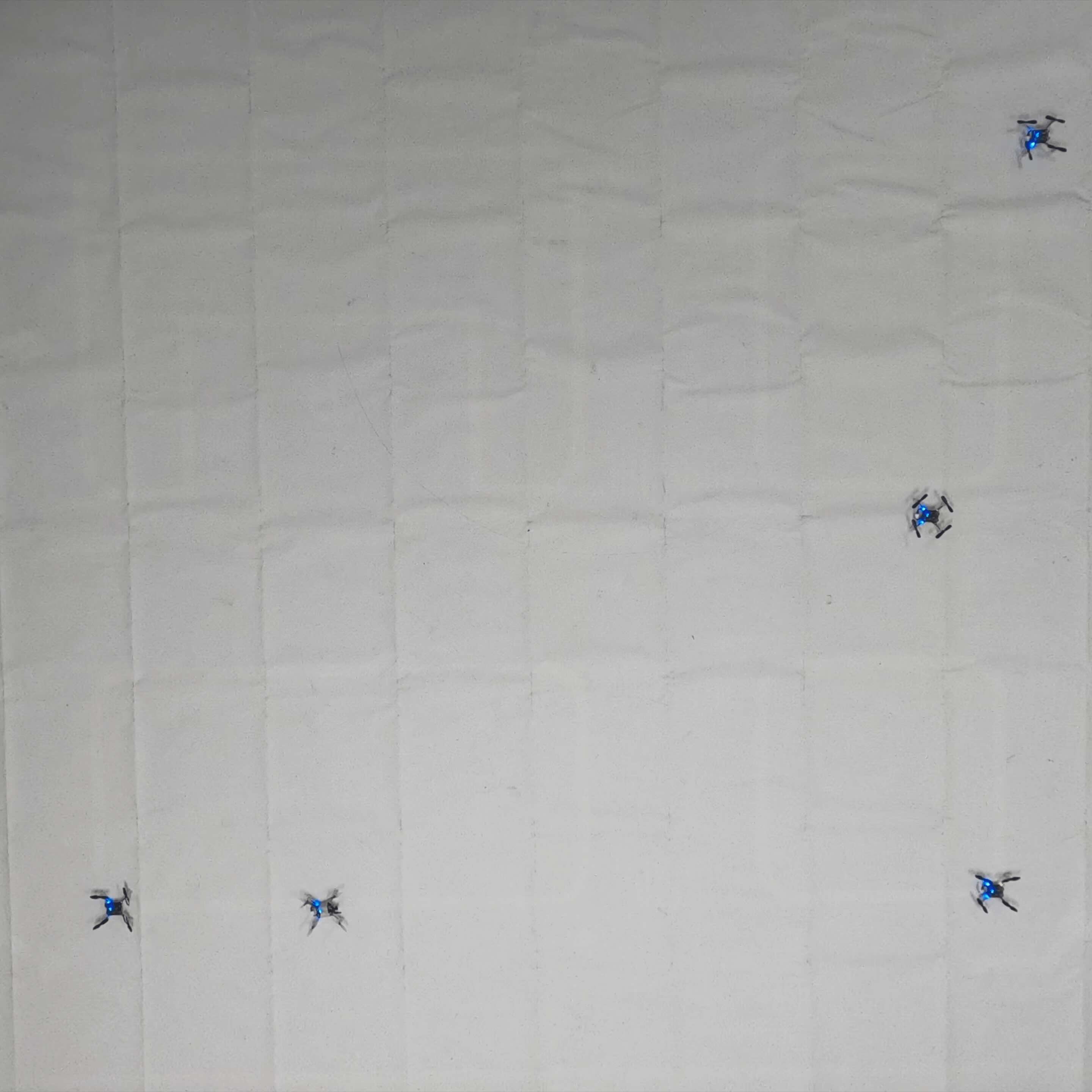}}
    \subfigure[Step 4]{\includegraphics[width=0.22\linewidth]{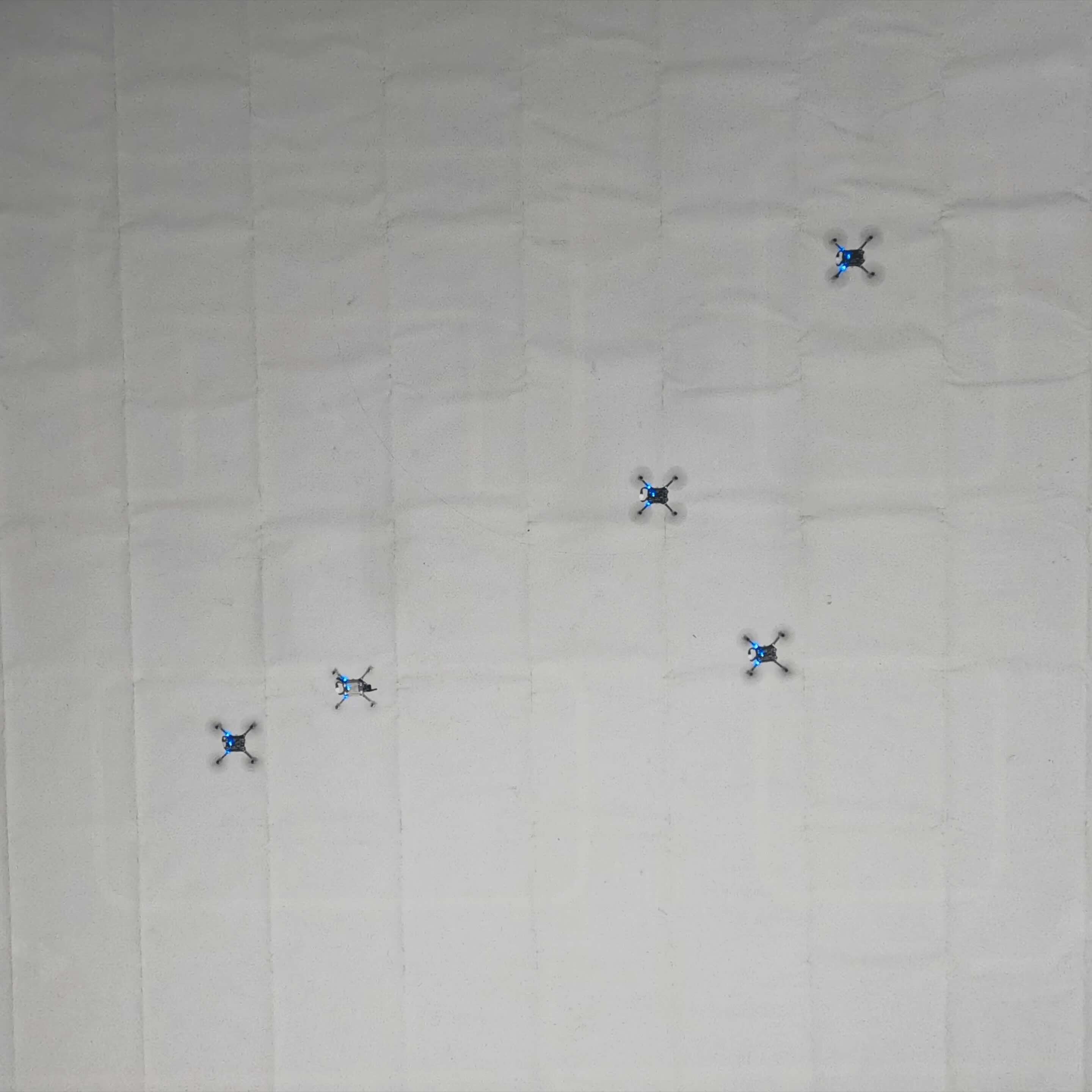}}
    \subfigure[Step 7]{\includegraphics[width=0.22\linewidth]{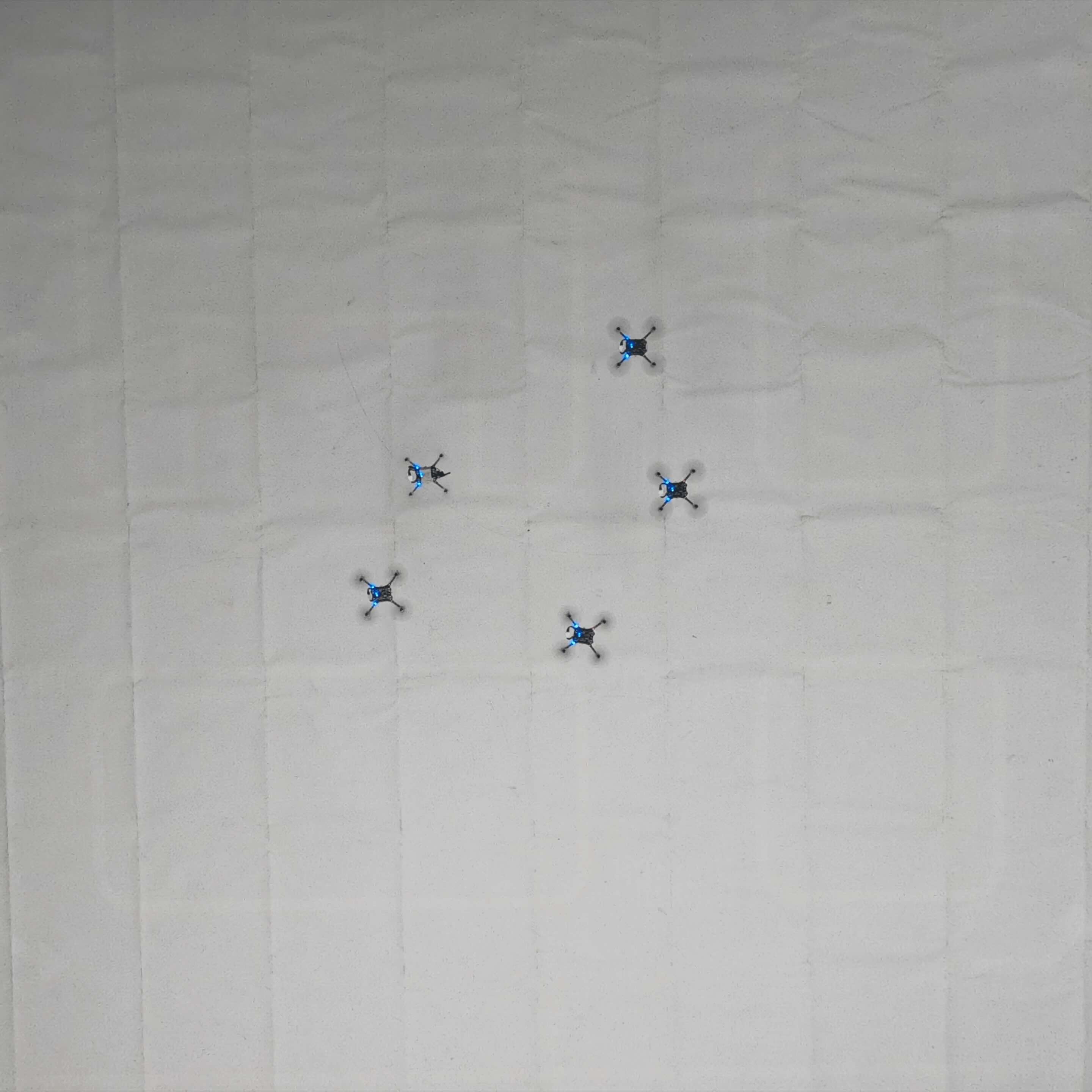}}
    \subfigure[Step 12]{\includegraphics[width=0.22\linewidth]{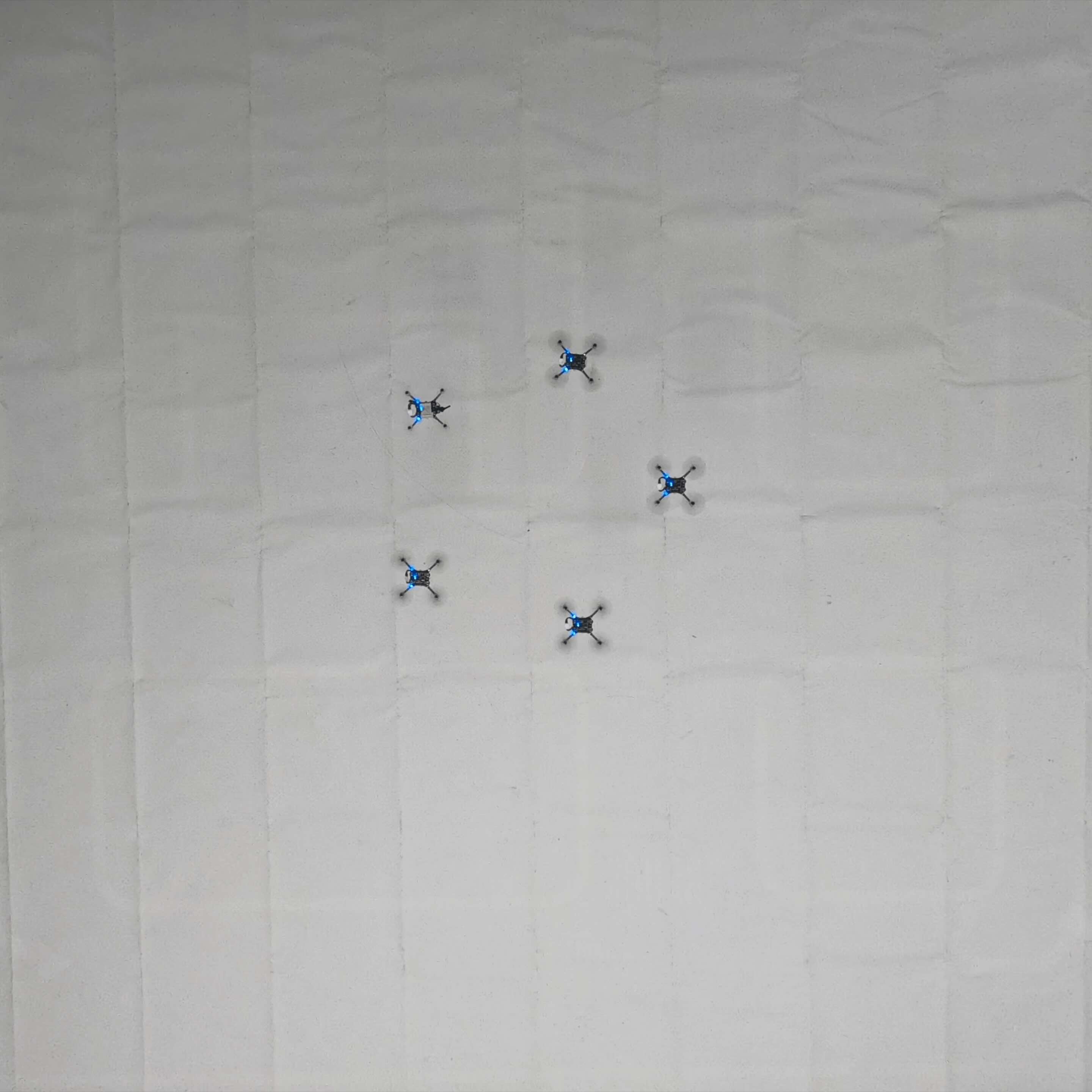}}
    \caption{Real drone experiments using proposed LLM-Flock framework with GPT-5 as LLM backend. (a)–(d): Forming a triangle with three drones. (e)–(h): Forming a square with four drones. (i)-(l): Forming a cross with five drones. (m)-(p): Forming a circle with five drones.}

    \label{fig:real-exp}
    \vspace{-0.3cm}
\end{figure}

\begin{table}[!hb]
    \centering
    \begin{tabular}{c|c|c|c}
         & GPT-5 & Claude 4.5 & DeepSeek-R1 \\
        \hline
        Prompt Only & 4 & 4 & 6 \\
        \hline
        Two-Layer (Re-prompt) & \textbf{0} & \textbf{0} & \textbf{0} \\
    \end{tabular}
    \caption{Collision trials in 10 trials of circle formation w/ 10 robots.}
    \label{tab:collision-stats}
    \vspace{-0.3cm}
\end{table}
\subsection{Collision Avoidance}
\PL{We evaluate the two-layer collision avoidance mechanism during circle formation with 10 robots using three LLMs: GPT-5, Claude 4.5, and DeepSeek-R1. Table~\ref{tab:collision-stats} compares collision occurrences with and without the collision checking and re-prompting layer across 10 trials per model.

With prompt-based urgency reasoning alone (Layer 1), GPT-5 and Claude 4.5 each produce 4 collisions, and DeepSeek-R1 produces 6 collisions, confirming that LLM spatial reasoning alone cannot guarantee collision-free behavior. Adding the collision checking and re-prompting layer (Layer 2) eliminates all collisions across all three models and all trials. Across all two-layer experiments, the minimum inter-agent distance never fell below the safe distance $d_s$.}

\PL{Fig.~\ref{fig:collide} illustrates a representative example using GPT-5. In Fig.~\ref{fig:collide-1}, two robots with cyan disks in the upper right corner are on converging paths. The collision checker detects the conflict and re-prompts the lower-urgency robot. As shown in Fig.~\ref{fig:collide-2} and \ref{fig:collide-3}, the two robots select divergent paths, achieving collision-free progress toward their goals.
}

\subsection{Real Robot Evaluation}


We validate the practical feasibility of LLM-Flock using Crazyflie drones in an indoor motion capture environment. Each robot executes the LLM-Flock protocol, performing independent plan generation and influence-based consensus with neighbors using the OpenAI GPT-5 model. 
Notably, the experiment is a proof-of-concept validation of our proposed method and does not capture the real-time GPT-5 response duration, which will be addressed in our future work.

We test a circle formation with five Crazyflie drones initialized from random positions. Snapshots in Fig.~\ref{fig:real-exp} show the team progressively converging to a stable circle without centralized control. These results show that LLM-Flock generalizes to real-world decentralized systems. 





\section{Conclusion}
\label{sec:conclusion}
We introduced LLM-Flock, a decentralized flocking framework that integrates LLM-based local planning with influence-based consensus \PL{and a two-layer collision avoidance mechanism}. Experiments across diverse formations, five LLMs, and real-world Crazyflie deployments demonstrate \PL{stable and scalable flocking without centralized coordination.} 

\PL{Several limitations motivate future work. First, each motion step requires a cloud-based API call, with response times exceeding 10 seconds, depending on the model, thereby limiting real-time applicability. Deploying locally hosted or distilled models could reduce this overhead. Alternatively, we can investigate asynchronous execution schemes in which robots can act on partial plans while awaiting updated LLM responses. Second, while the two-layer collision avoidance achieves collision-free behavior in our experiments, it does not provide formal mathematical safety guarantees. Integrating methods such as Control Barrier Functions as a low-level safety filter is a promising direction. Finally, future work can investigate the dynamic formation center to better capture real-world application scenarios.}




\bibliography{reference}  




\end{document}